\pdfoutput=1
\documentclass[sigconf]{acmart}

\usepackage{amsmath}
\usepackage{amssymb}
\usepackage{amsfonts}
\usepackage{chemarrow}
\usepackage{makecell}

\usepackage{graphicx}
\usepackage{subfigure}
\usepackage{color}
\usepackage[linesnumbered,ruled]{algorithm2e}
\newcommand{\BfPara}[1]{{\noindent\bf#1.}\xspace}
\usepackage{multirow}
\usepackage{float}

\usepackage[capitalise]{cleveref}

\usepackage{framed}
{\bfseries}{\itshape}
{\bfseries}{\itshape}
{\bfseries}{\itshape}
{\bfseries}{\itshape}
{\bfseries}{\itshape}
{\itshape}

\usepackage[inline]{enumitem}

\usepackage{xspace}
\usepackage{setspace}
\newcommand{\note}[1]{}

\newcommand{\etc}{{etc.}\xspace}
\newcommand{\eg}{{\em e.g.}\xspace}

\newcommand{\etal}{{\em et al.}\xspace}

\definecolor{linkcolour}{rgb}{0,0.2,0.6}
\definecolor{xgreen}{rgb}{0.2,0.6,0.0}
\definecolor{xred}{rgb}{0.7,0.1,0.0}
\hypersetup{colorlinks,breaklinks,citecolor=green!30!black, 
                                            urlcolor=red!20!black, 
                                            linkcolor=blue!50!black}
\def\equationautorefname~#1\null{(#1)\null}

\newcommand{\atan}{\text{atan}}

\colorlet{punct}{red!60!black}
\definecolor{background}{HTML}{ffffff }
\definecolor{delim}{RGB}{20,105,176}
\colorlet{numb}{magenta!60!black}

\definecolor{light-gray}{gray}{0.95}
\definecolor{darkgray}{rgb}{0.4, 0.4, 0.4}
\definecolor{editorGray}{rgb}{0.95, 0.95, 0.95}
\definecolor{editorOcher}{rgb}{1, 0.5, 0} 
\definecolor{editorGreen}{rgb}{0, 0.5, 0} 
\definecolor{orange}{rgb}{1,0.45,0.13}      
\definecolor{olive}{rgb}{0.17,0.59,0.20}
\definecolor{brown}{rgb}{0.69,0.31,0.31}
\definecolor{purple}{rgb}{0.38,0.18,0.81}
\definecolor{lightblue}{rgb}{0.1,0.57,0.7}
\definecolor{lightred}{rgb}{1,0.4,0.5}
\usepackage{upquote}
\usepackage{listings}

\newcommand{\ours}{{FAQ}\xspace}

\definecolor{pblue}{rgb}{0.13,0.13,1}
\definecolor{pgreen}{rgb}{0,0.5,0}
\definecolor{pred}{rgb}{0.9,0,0}
\definecolor{pgrey}{rgb}{0.46,0.45,0.48}

\renewcommand\footnotetextcopyrightpermission[1]{} 
\begin{document}

\title{Generating Adversarial Examples with an Optimized Quality}


\author{Aminollah Khormali}
      \affiliation{\institution{University of Central Florida}}
      \email{aminkhormali@Knights.ucf.edu}

\author{DaeHun Nyang}
      \affiliation{\institution{INHA University}}
      \email{nyang@inha.ac.kr}
       
\author{David Mohaisen}
      \affiliation{\institution{University of Central Florida}}
      \email{mohaisen@ucf.edu}

\begin{abstract}
Deep learning models are widely used in a range of application areas, such as computer vision, computer security, etc. However, deep learning models are vulnerable to Adversarial Examples (AEs), carefully crafted samples to deceive those models. Recent studies have introduced new adversarial attack methods, but, to the best of our knowledge, none provided {\em guaranteed quality} for the crafted examples as part of their creation, beyond simple quality measures such as Misclassification Rate (MR). In this paper, we incorporate Image Quality Assessment (IQA) metrics into the design and generation process of AEs. We propose an evolutionary-based single- and multi-objective optimization approaches that generate AEs with high misclassification rate and explicitly improve the quality, thus indistinguishability, of the samples, while perturbing only a limited number of pixels. In particular, several IQA metrics, including edge analysis, Fourier analysis, and feature descriptors, are leveraged into the process of generating AEs. Unique characteristics of the evolutionary-based algorithm enable us to simultaneously optimize the misclassification rate and the IQA metrics of the AEs. In order to evaluate the performance of the proposed method, we conduct intensive experiments on different well-known benchmark datasets (MNIST, CIFAR, GTSRB, and Open Image Dataset V5), while considering various objective optimization configurations. The results obtained from our experiments, when compared with the existing attack methods, validate our initial hypothesis that the use of IQA metrics within generation process of AEs can substantially improve their quality, while maintaining high misclassification rate. Finally, transferability and human perception studies are provided, demonstrating acceptable performance.

\end{abstract}

\maketitle

\section{Introduction}\label{sec:introduction}

The increasing use of deep learning networks incentivizes adversaries to manipulate those networks such that the model outputs their desired output, \eg, misclassification. In particular, recent works at the intersection of machine learning and security have shown that adversaries can force deep learning models to produce adversary-selected outputs through Adversarial Examples (AEs)~\cite{PapernotMJFCS16,Moosavi_Dezfooli16}. AEs threaten the security of critical ML and DL applications, since miscreants could utilize them for malicious purposes, such as misleading autonomous driving vehicles~\cite{kurakin2017adversarial, eykholt2018robust, Xie2017Adversarial},  hijacking voice controlled intelligent agents~\cite{carlini2016hidden, zhang2017dolphinattack}, evading deep learning-based malware detection systems~\cite{grosse2017adversarial, al2018adversarial, kolosnjaji2018adversarial}, \etc

AEs are created by adding a small perturbation to the original input of the ML algorithm in order to produce the adversary's desired outputs~\cite{PapernotM0JS16}. Since the crafted AEs are generated by applying limited changes to the original inputs, they are very similar to the original ones, and are not necessarily outside of the training data manifold. Ideally, algorithms crafting AEs need to minimize the perturbation, thus making AEs hard to distinguish from legitimate samples. As these attacks occur after the training is completed, there is no need to tamper with the training procedures. Recently, several adversarial generation algorithms have been presented, including the fast gradient sign method~\cite{Goodfellow2015Explaining}, the Jacobian-based saliency map approach~\cite{PapernotMJFCS16}, the virtual adversarial method~\cite{miyato2015distributional}, the Carlini and Wanger (C\&W) method~\cite{Carlini017}, the NewtonFool~\cite{Jang0J17}, and the Projected Gradient Descent (PGD)~\cite{madry2018towards}, among others.

To evaluate the performance of the AEs' methods, the research community explored various assessment metrics. The misclassification rate, as a result of fooling the learning model, is commonly used~\cite{Goodfellow2015Explaining}.  Another widely used metric is the similarity between the  generated AE and the original sample. For example, the average distortion introduced by the algorithm is used as a metric for similarity. Moreover, the human perceptibility was considered as an assessment metric~\cite{PapernotMJFCS16}. Others~\cite{Jang0J17} investigated the quality of the generated samples using various generator models. However, they had no control over the quality metrics of the generated examples within the process of crafting the AEs. Moreover, although these metrics are used to assess the AEs,  incorporating Image Quality Assessment (IQA) metrics into AEs' generation to produce high-quality AEs is, surprisingly, missing from the literature. It is noted that such IQA metrics can be utilized as a protection mechanism against adversarial attacks, specifically in image related DL model such as face spoofing~\cite{galbally2014face}, whereby AEs with lower quality are detected and discarded. Therefore, for successful attacks, the generated AEs should be of high quality, in terms of IQA metrics, in order to remain indistinguishable using those IQA metrics. 

\BfPara{Why IQA Metrics} Image visual perception (approximated by L2-norm~\cite{Carlini017}) is different from image quality assessment metrics. In automated image recognition in a  self-driving cars' control systems, for example, the human perception-based visual quality plays very little role in determining actions, and decisions are based on quality metrics: poor quality images will be rejected as a possible input. Moreover, generating AEs with high IQA metrics, such as edges, HOG descriptors, \etc, for images that include those used in automated AI-based setting (e.g., traffic signs) have significant implications, and where other approaches that provide lower IQA metrics may fail, an approach that ensures misclassification while maintaining a high IQA metrics may succeed in fooling those automated detection tools, challenging the robustness of such systems.

\BfPara{Contributions} 
To address the aforementioned gap, we propose \ours (\underline{F}alse \underline{a}nd \underline{Q}uality), an approach to produce high-quality AEs in terms of IQA metrics with the following concrete contributions. 
\begin{enumerate}[label={\arabic*)},start=1]
    \item We propose \ours which leverages image quality metrics, such as edge analysis, FFT analysis and HOG feature descriptors, into the AE generation process. Unlike other existing approaches, \ours enables control over multiple characteristics of the generated AEs, such as brightness, global shape, \etc, leading to AEs that are very similar to the original images.
    \item To address the scalability of AE generation, we devise an approach in which we perturb only a limited number of pixels in the candidate image while achieving high misclassification rate. For the selected pixels we search for an optimal value of the perturbations through an evolutionary-based multi-objective optimization algorithm, which is sufficient to achieve high misclassification rate while producing high-quality AEs in terms of IQA metrics.
    \item We evaluate the performance of \ours through intensive experiments on four different benchmarks, namely MNIST hand-written digits, Fashion MNIST, CIFAR-10, and GTSRB (and a high resolution dataset). A comparison of the obtained results from different configurations of \ours with that of popular AE methods confirm the effectiveness of \ours in generating high quality AEs. 
\end{enumerate}
\BfPara{Organization} \cref{sec:Prelim} provides a brief background on quality metrics. \cref{sec:Methedology} describes our evolutionary-based approach for generating high-quality AEs and configurations. \cref{sec:Eval_Discussion} describes the  performance evaluation through various multi-objective quality metric configurations, as well as the results and discussion. \cref{sec:related} describes the related work, followed by conclusions in~\cref{sec:conclusion}. 

\section{Background and Preliminaries}\label{sec:Prelim}
To ensure AE's quality by design, we incorporate IQA metrics and evolutionary-based optimization algorithms in the process for generating AEs. To set out, we first describe the background necessary for understanding those techniques. We need saliency maps to analyze the sensitivity of model output to its input features, described in~\cref{sec:saliency}. Three popular quality metrics used in our analysis are described in the three subsequent sections: Canny edge analysis in~\cref{sec:CED}, fast Fourier transform analysis in~\cref{sec:FFT}, and histogram of oriented gradient in~\cref{sec:HOG}. The particle swarm optimization algorithm is described in~\cref{sec:PSO}. 

\subsection{Threat Model}\label{sec:Prelem_Threatmodel}
In adversarial learning, the main goal of the adversary is to produce an input sample $x'$ such that it is misclassified by the model $f$. Attacks on deep learning networks can be categorized from multiple perspectives, including the adversary's goal and capabilities. Based on the orientation of the attacker the attack can be either targeted or untargeted, and based on his knowledge about the model the attack can be either black- or white-box attacks~\cite{papernot2017practical,wang2018great}. In this study, we assume that the adversary has full knowledge of the topology of the model, the link weights, \etc In addition, it is assumed that the adversary is trying to conduct untargeted misclassification attacks. A brief categorization of the adversarial attacks on deep learning networks is provided in~\autoref{app:1}.

\subsection{Sensitivity Analysis}\label{sec:saliency}
The Jacobian-based Saliency Map Approach (JSMA), due to Papernot \etal~\cite{PapernotMJFCS16}, produce AEs and force the model to classify the input AEs as a specific class. The intuition of the adversarial saliency map is to find features that, once modified, would have the most impact on the output of the ML/DL-based classifier. Thus, such pixels are used by adversaries to perform source-targeted attacks. One advantage of this map is achieving misclassification while distorting a small number of pixels. For example, perturbing 32 pixels leads to $\approx97$\% misclassification rate for targeted-misclassification on the MNIST hand-written digits dataset~\cite{PapernotMJFCS16}. 


For a DL model $F$, the saliency map is defined based on the forward derivative of the model. In \autoref{eq:saliency}, we compute the saliency score for each pixel of a given image where perturbing highly scored pixels is more likely to yield a desired output. Note that the adversary desires to misclassify a given input sample to a target class other than the original one; i.e., increasing the probability of a target class $t$ while decreasing the probability of all others. This task can be performed by increasing the value of the salient pixels, computed using the following saliency map $S(I, t)$:
\begin{equation}\label{eq:saliency}
    S(I,t)[i]=\left\{ \begin{matrix}
    0 \ if \ J_{it}< 0 \ or \ \sum_{j\neq t}{}J_{it}\left ( I \right ) > 0\\ 
    J_{it}\left ( I \right )\left | \sum_{j\neq t}{}J_{it}\left ( I \right ) \right | \ otherwise
    \end{matrix}\right. .
\end{equation}
In~\autoref{eq:saliency}, $i$ denotes the pixels of a given image $I$, while  $J_{it}\left ( I \right )$ shows $J_F\left [ i , j \right ]\left ( I \right )= \frac{\partial F_{j}\left ( I \right )}{\partial X_{i}}$. In summary, pixels with high value of $S \left (I, t \right) \left [i \right ]$ correspond to pixels that increasing their value would result in increasing the probability of target class, or reducing the probability of the other classes, or both cases. We refer the interested readers to~\cite{PapernotMJFCS16} for more information.

\begin{figure*}[ht]
\centering
		\subfigure[MNIST-digits\label{fig:Digits}] {\includegraphics[width=0.18\textwidth]{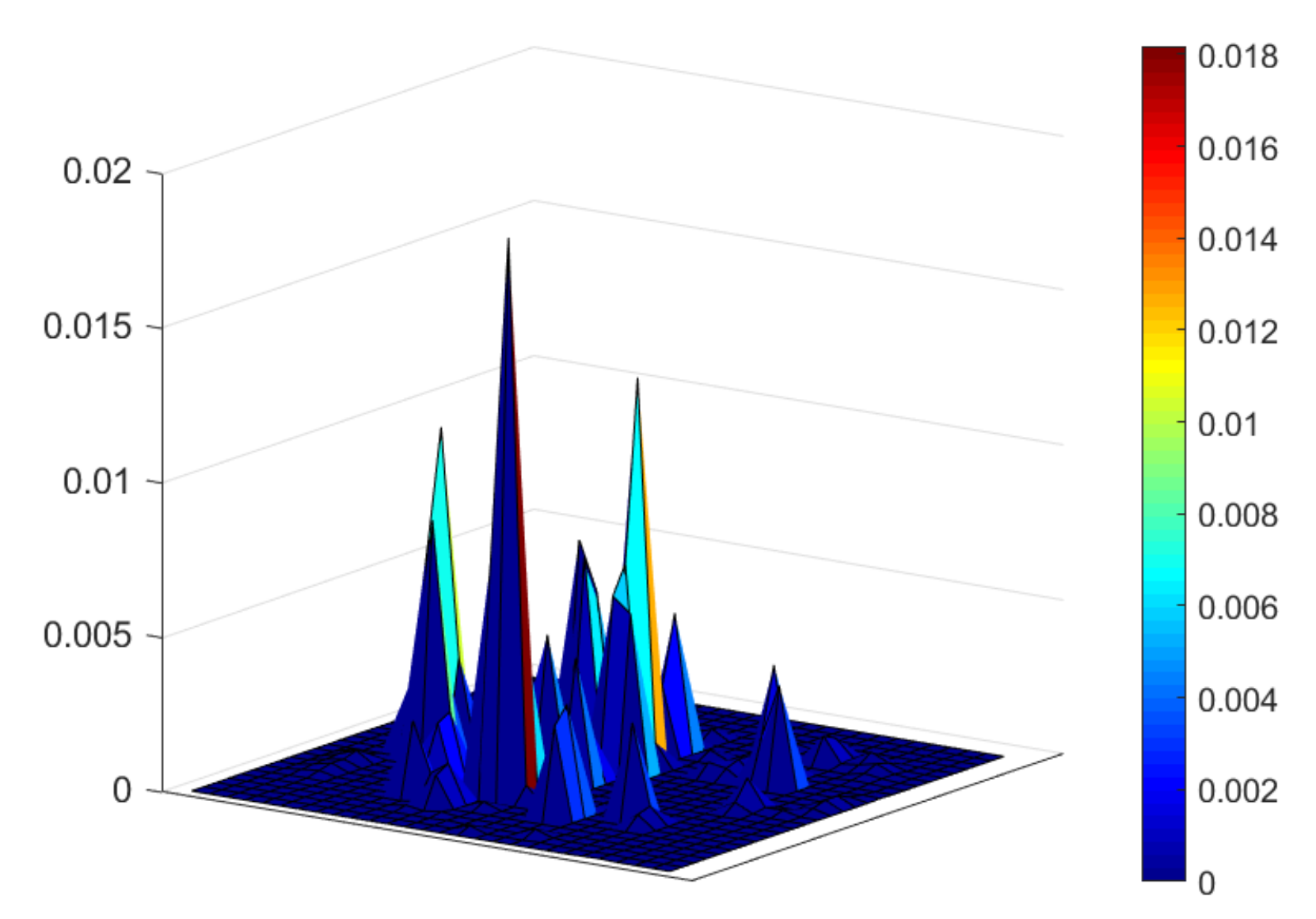}}
		\subfigure[Fashion MNIST \label{fig:Fashion}] {\includegraphics[width=0.18\textwidth]{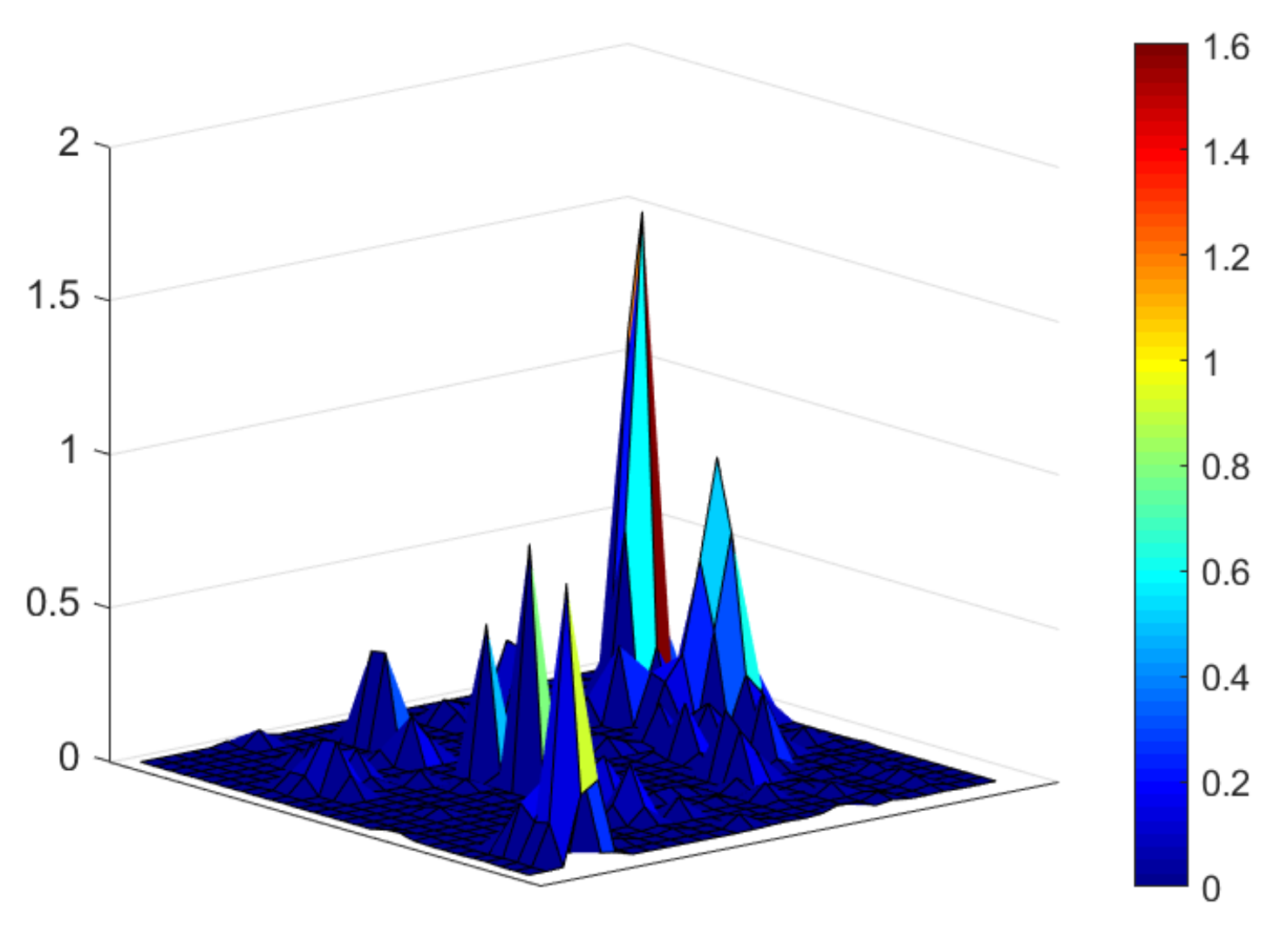}}
		\subfigure[CIFAR-10 \label{fig:CIFAR}] {\includegraphics[width=0.18\textwidth]{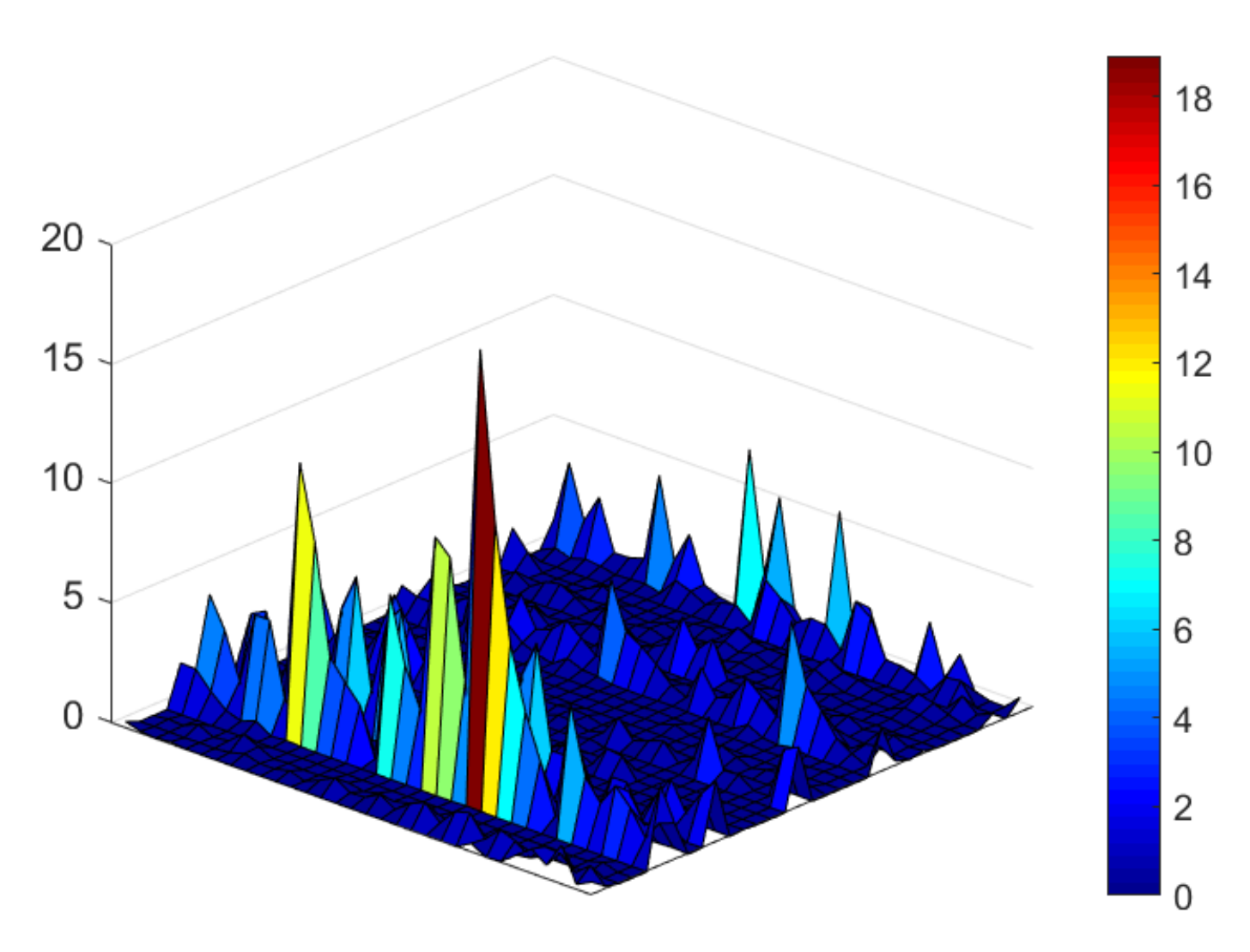}}
		\subfigure[GTSRB \label{fig:GTSRB}] {\includegraphics[width=0.18\textwidth]{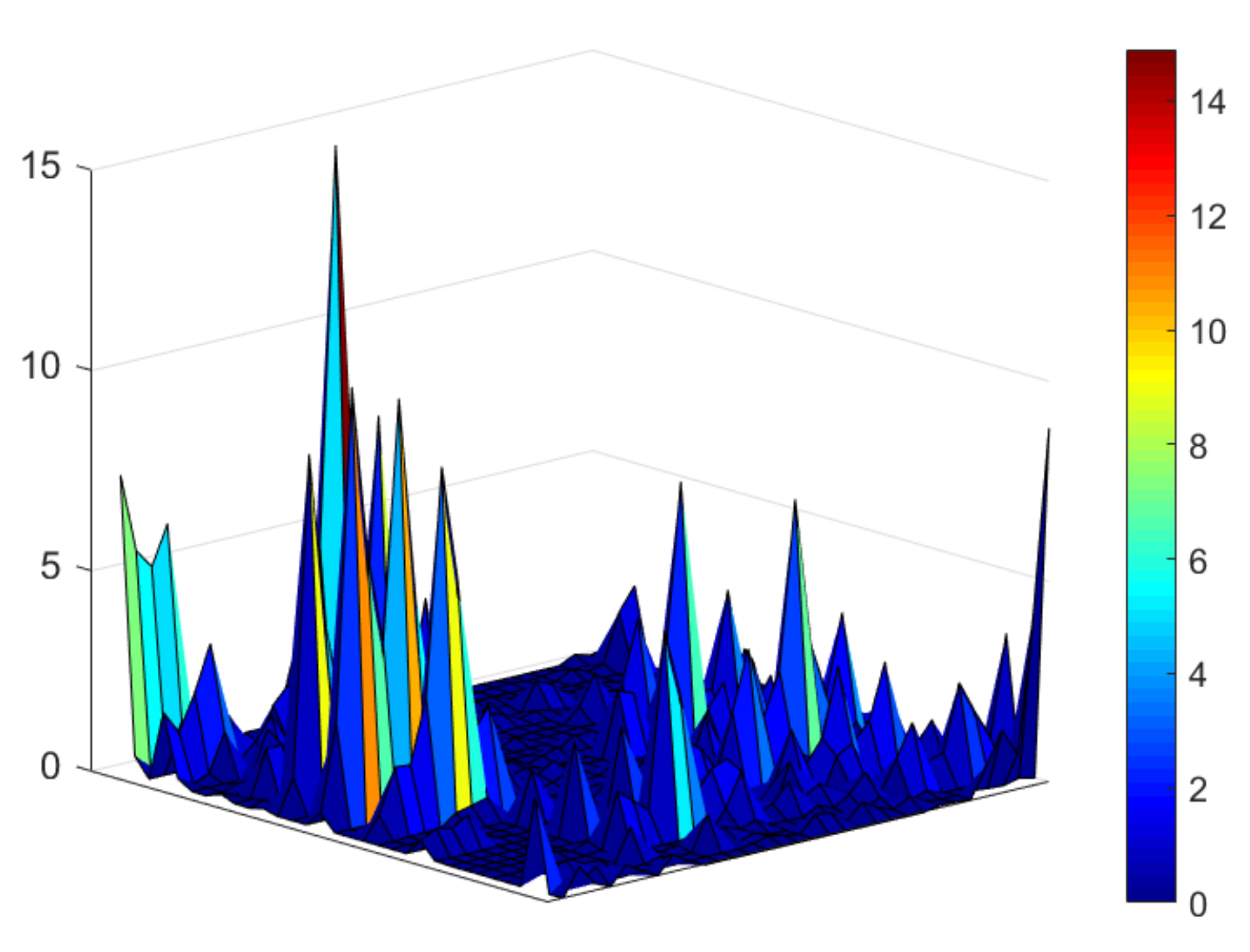}}
		\subfigure[Open Image \label{fig:OpenImage}] {\includegraphics[width=0.18\textwidth]{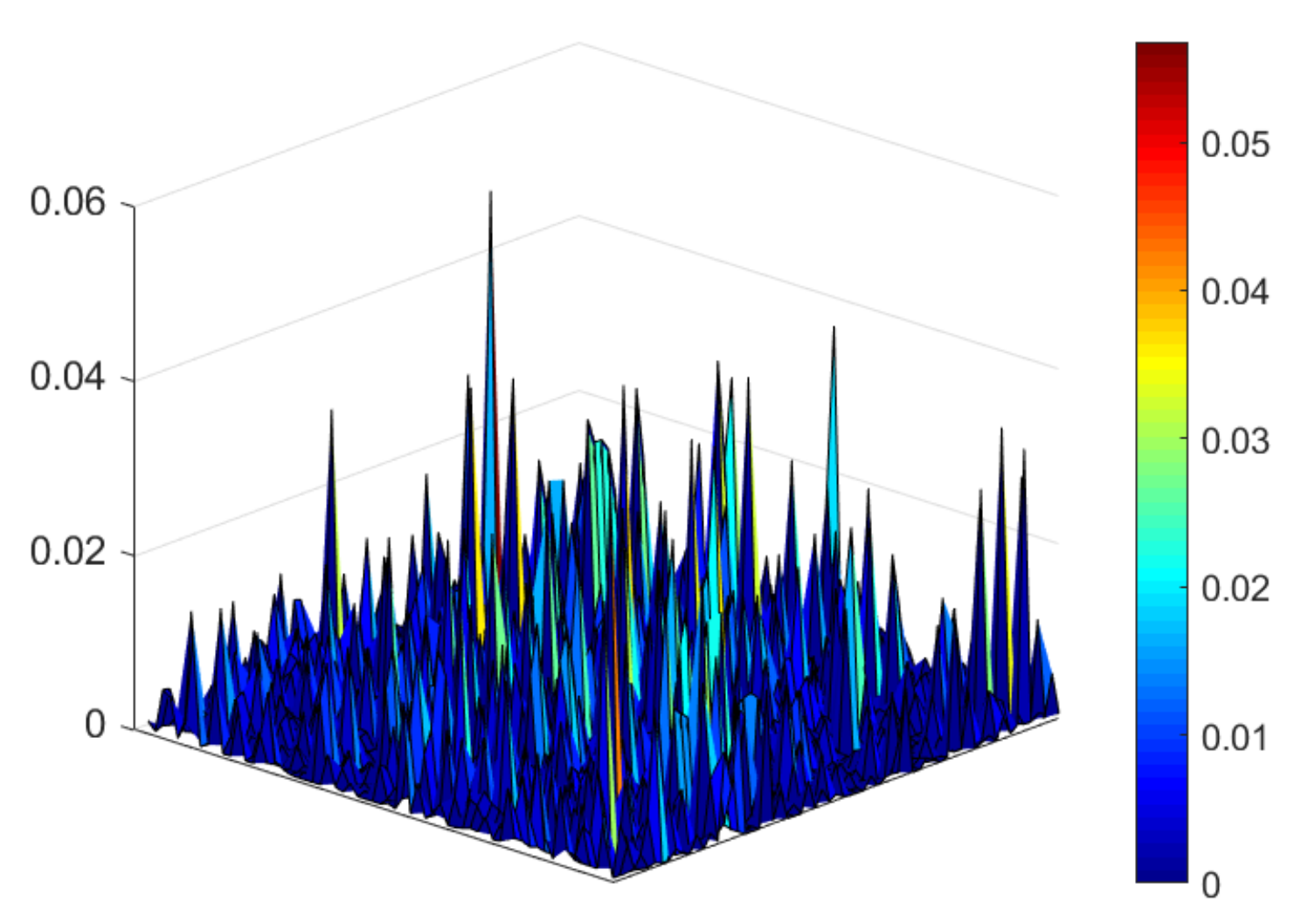}}\vspace{-2mm}
\caption{A sample saliency map for each benchmark dataset using associated trained CNN model: \cref{fig:Digits} MNIST hand-written digits, \cref{fig:Fashion} Fashion MNIST, \cref{fig:CIFAR} CIFAR-10, \cref{fig:GTSRB} GTSRB,  and  \cref{fig:OpenImage} Open Image datasets. Pixels with large saliency score have significant impact on the model output, thus are potential pixels to be perturbed.}\label{fig:SalielncyScore}\vspace{-3mm}
\end{figure*}

\subsection{Canny Edge Detection}\label{sec:CED}
In the image processing and computer-vision communities, an edge is defined as a set of curved line segments, composed of pixels, at which brightness of the image changes drastically. Thus, edges contain valuable structural information about properties of the given image, and have been used in many computer-vision applications, specifically in the areas of local feature detection. Edges in an image may correspond to the following changes: 1) depth discontinuity, 2) surface orientation discontinuity, 3) changes in material properties, and 4) changes in the illumination of the scene. 

There are several edge detection methods, such as the \textit{Canny Edge Detector (CED)}~\cite{Canny86a} and Deriche Edge Detector (DED)~\cite{Deriche87}. In this work, we used CED as a metric to evaluate and compare the total number of edges in the original images versus crafted AEs. CED, a well-known edge detector, fulfills the following requirements: 1) closeness of real and detected edges, 2) no duplicated detection of a given edge, and 3) low false positive rate. A high-level architecture of the CED algorithm is shown in~\cref{fig:Canny_flowchart}. The steps that CED takes to detect edges, including noise reduction, computing the intensity gradient, non-maximum suppression, and hysteresis thresholding are described in detail in~\autoref{app:1}.

\subsection{Fast Fourier Transform}\label{sec:FFT}
In signal processing, analyzing signals in the frequency domain offers more information than analyzing them in temporal or spectral domains. Spectral analysis is a well-known statistical method to transform signals into frequency domain. The intuition behind spectral analysis is that each waveform is composed of the sum of sine waves with different phase relationships, amplitudes, and frequencies~\cite{brigham1988fast,alleyne1991two}. There are various mathematical transforms converting signals into spectra, and Fourier transform is a prominent method for transforming signals into frequency domain.

In our analysis we used 2D FFT to analyze the quality of generated AEs. Spectra on two dimensional spatial frequency domain can be achieved through Fourier transform on two dimensional spatial domain. Representative information of features of an image would be provided, as these spectra represent the periodic structures across positions in space. Particularly, rough shape structure of a given image matches the low spatial frequency, while detailed features such as edges and illuminations can be represented by spectra on high spatial frequency parts. The 2D Fourier transform is briefly described in~\autoref{app:1}.

\subsection{Histogram of Oriented Gradient}\label{sec:HOG}
In the computer-vision community, feature descriptors are widely used to identify important regions of a given image. The Histogram of Oriented Gradient (HOG) is a popular feature descriptor used to detect objects in an image, primarily for pedestrian detection in static images~\cite{DalalT05}. The intuition behind HOG is to describe appearance and shape of local object in a given image using intensity gradient distribution or direction of the edges, even without accurate information regarding the location of the edges. HOG is a popular method to detect objects, and offers the following properties: 1) invariant to geometric transformations, 2) fine orientation sampling, 4) coarse spatial sampling, and 4) local normalization. ~\autoref{app:1} provides more details on the main steps of the HOG descriptor, consisting of gradient computation, orientation binning, block normalization, and object orientation.

\subsection{Particle Swarm Optimization}\label{sec:PSO}

The Particle Swarm Optimization (PSO) algorithm is an evolutionary-based optimization method proposed by Eberhart and Kennedy~\cite{eberhart1995new}, and is inspired by the behavior of a group of particles, such as birds, fish, or people. Similar to other population-based optimization algorithms, the PSO algorithm comprises a set of individuals, known as swarm particles, using their information about potential answers for an optimization task in a given multidimensional search space. 

Swarm particles' position determines a potential solution for the optimization problem at hand. The PSO algorithm searches for better positions that provide better fitness evaluation in each iteration. Each particles will have a specific fitness value with regard to its position. Thus, the particles will move to positions in the search space that give better fitness values. This movement takes place with specific velocity for each particle and in repeated iterations, and continues until the optimum solution with the highest fitness value is obtained. The value of velocity is updated in each iteration based on the particles' previous position and that of neighbors~\cite{jiao2008elite}. 

The velocity of each individual in the search space is determined by both its own and its companions'~\cite{shi2001particle}. In the $N$-dimensional search space each individual is considered as a volume-less particle. The $i$th particle at time step $t$ can be represented as $x_i\left ( t \right )=\left ( x_{i1}\left ( t \right ), x_{i2}\left ( t \right ), \dots, x_{iN}\left ( t \right )\right )$. The $i$th particle's best previous position is recorded and represented as $p_i=\left ( p_{i1}, p_{i2}, \dots, p_{iN}\right )$.  The best global particle achieving the best objective values among all particles in the population is denoted by $g{best}$. The movement velocity of a particle at a time step is represented as $v_i\left ( t \right )=\left ( v_{i1}\left ( t \right ), v_{i2}\left ( t \right ), ..., v_{iN}\left ( t \right )\right )$. 
The velocity and position of particles are updated in each search iteration based on~\cite{eberhart1995new,shi2001particle}:
\begin{equation}\label{eq:pso_velocity}
\resizebox{.42 \textwidth}{!}{
    $\nu _{id}^{\left ( t \right )}=w_{i}\nu _{in}^{\left ( t-1 \right )}+ c_{1}r_{1}\left ( p _{in}^{\left ( t-1 \right )}-x _{in}^{\left ( t-1 \right )} \right )+ c_{2}r_{2}\left ( p _{gn}^{\left ( t-1 \right )}-x _{in}^{\left ( t-1 \right )} \right ), $
    }
\end{equation}

\begin{equation}\label{eq:pso_position}
x _{in}^{\left ( t \right )}=x _{in}^{\left ( t-1 \right )}+\nu _{in}^{\left ( t \right )}
\end{equation}

where $n$ is the dimension; $1\leq n\leq N$, $c_{1}$  and $c_{2}$ are positive constants, $r_{1}$  and $r_{2}$ are two random functions in the range $\left [ 0 \ \ 1 \right ]$, and  $w$ is the inertia weight. For the neighborhood $lbest$ model, the only change is to substitute $p_{ln}$ for $p_{gn}$ in the equation for velocity~\autoref{eq:pso_velocity}. This equation in the global model is used to compute a particle's new velocity with regard to not only its previous velocity and the distance of its own best experience $p{best}$ from the current position but also the group's best experience $g{best}$~\cite{eberhart1995new}. 

One of the key advantages of particle swarm optimization algorithm is its simplicity, since it requires only a few parameters to be adjusted. The general structure of PSO algorithm is shown in~\cref{fig:General_Flowchart}. Further information about PSO algorithm can be found in~\cite{eberhart1995new,shi2001particle}.

\section{Generating Examples with High IQA}\label{sec:Methedology}

Applying \textit{small changes} to the input data is a common step among several methods that generate AEs. However, different definitions have been proposed for \textit{small changes}, including the average distortion introduced by the algorithm and the number of distorted pixels. Thus, it would be desirable if an algorithm can produce AEs by not only applying perturbations with a small size but also by changing a small number of pixels, instead of a whole feature space. Note that not all regions from the input domain contribute to the AEs~\cite{PapernotMJFCS16}. To this end, in this study we sort the pixels based on their saliency score and restrict the number of pixels to be perturbed to less than $\approx10\%$. Moreover, we leverage the IQA metrics  within the process of crafting AEs to control their quality.

First, features with a significant impact on the decision boundaries of the DL network are extracted and sorted in a decreasing order. Then, we attempt to generate AEs that force the DL-based model into misclassification, by applying an optimal value of perturbation into only the top $n$ features, while maintaining similar IQA metrics to the original sample. To set out, two different configurations, with and without quality metrics, are designed: the Single-Objective Optimization (SOO) and the Multi-Objective Optimization (MOO) configuration. The goal of SOO is to generate AEs by applying an optimal size of perturbation into the top $n$ salient pixels, where none of the IQA metrics are considered in the AE generation process. MOO configuration takes into account the IQA metrics within the AE creation process.

Unlike~\cite{PapernotMJFCS16}, we do not set the value of the pixel to be perturbed into one; rather, we utilize a evolutionary-based optimization method to find the optimal value of the perturbation to achieve misclassification. \ours benefits from an evolutionary-based search method, PSO algorithm, to find an optimal value of the perturbations. Note that evolutionary-based search algorithms are less likely to be trapped in a local minima, unlike gradient-based algorithms.

\begin{figure}[t]
\centering
       {\includegraphics[width=0.45\textwidth]{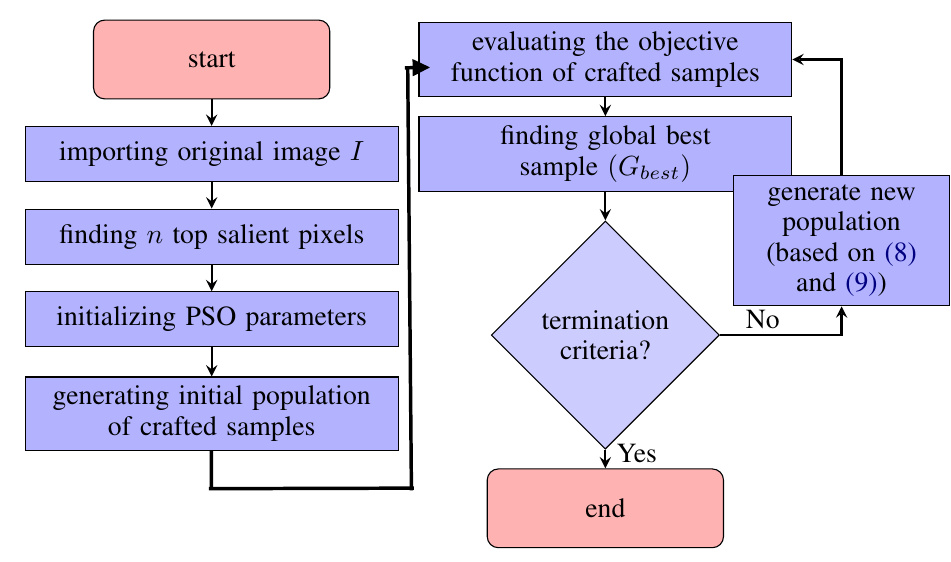}}
\caption{High-level structure of \ours. }
\label{fig:General_Flowchart}
\end{figure}

A flowchart of \ours is in~\cref{fig:General_Flowchart}. For every input image $I_{org}$, we analyze the sensitivity of the CNN model's output to its input features using the saliency map. These features are then sorted in a decreasing order using their saliency score. Pixels with a large saliency score have more impact on the output of the model, thus their modification is more likely to lead to misclassification. In the rest of this section, more detailed information about the two configurations of \ours is provided. First, we present the SOO configuration in~section~\ref{sec:SOO_methodology} and then the MOO configuration is presented in~section~\ref{sec:CC_methodology}. 

\subsection{Single-Objective Optimization (SOO)}\label{sec:SOO_methodology}

In this section, we show how to generate AEs by applying optimal perturbations into the original image $I$ such that the model output $f(x_{adv})$ is labeled with a different class than the original output $f(x)$. In general, the SOO configuration aims to answer the following three key questions:
\begin{enumerate*}
\item Can we generate AEs through distorting only a small number of pixels, instead of the whole feature space?
\item What is the average size of the distortion introduced by different adversarial algorithms?
\item How does the misclassification rate relate to the number of perturbed pixels?
\end{enumerate*}

To evaluate the SOO configuration, we used the misclassification rate as the number of AEs that successfully force the learning model to misclassification (targeted-misclassification in the case of JSMA) divided by the total number of attacked samples. The taken steps to craft AEs using the SOO configuration are shown in~\cref{al:MOO_Adv_Three}; for SOO, we use the weights $w_{i=2:4}=0$ (no IQA metrics are used). 

The SOO configuration is similar to other adversarial attack algorithms, where the main goal is to achieve misclassification. However, in SOO we try to achieve this goal by modifying only a small number of pixels, instead of distorting the whole feature space. {\em Note that we do not calculate forward derivatives in each iteration as it is done in~\cite{PapernotMJFCS16}, which in turn increases the computation time. Rather, we select the top $n$ pixels once at the initial step and stick with those pixels during the rest of our SOO run, and conduct our PSO search for the best perturbation values of those pixels to fulfills our goal}. Note that our approach is also different from~\cite{PapernotMJFCS16}, where in each iteration two top salient pixels' values are set to one until misclassification is achieved. Moreover, we focus on a limited number of pixels, which were selected initially based on the saliency map, and solve an evolutionary-based optimization problem, consisting of $n$ variables, to find the minimal values of perturbations that lead to misclassification.

\subsection{Multi-Objective Optimization (MOO)} \label{sec:CC_methodology} 
The goal is to produce AE $I_{adv}$ that is very similar to the original input $I_{org}$. Although several works have considered the number of distorted pixels or the difference in the L2-norm between $I_{org}$ and $I_{adv}$ as metrics for indistinguishability of the produced AEs~\cite{Jang0J17}, there is no work that incorporates IQA metrics into the AE generation process. It should be noted that there is a difference between assessing these quality metrics of the produced AEs, as it is performed in~\cite{Jang0J17}, and taking into account these IQA metrics as part of the objective function within the generation process: the former only measures the IQA metrics of the crafted AE without any control over them, while the latter controls the quality of the generated samples in terms of IQA metrics during the production process. The MOO configuration is defined by adding an IQA metric into the objective function of the SOO approach. Using MOO, we attempt to generate AEs by distorting only a small number of pixels while improving the quality of the generated AE in term of the IQA metrics. Particularly, we try to answer the following questions:
\begin{enumerate*}
\item How can we incorporate the aforementioned IQA metrics into the process of generating AEs? 
\item How do these metrics impact the misclassification rate and the quality of the crafted samples? 
\end{enumerate*}

To answer both questions, we incorporate the edge analysis, FFT analysis, and HOG feature descriptors, into the objective to be minimized simultaneously. Based on the MOO configuration, AEs that offer lower IQA metrics yield larger objective values, thus likely to be discarded within the AE creation process. Optimization of these metrics at the same time will result in AEs that are close to the original image in multiple aspects, such as brightness, global shape, feature descriptors, \etc These objectives are described below.

\BfPara{Edge Analysis}\label{sec:edge_TOC} 
Edge analysis is important to detect points in a given image at which the brightness of the image changes drastically. Edge analysis has been used in many computer-vision and image processing applications, specifically for feature extraction~\cite{UmbaughSF11}. Similarly, edge analysis can play a critical role in the context of adversarial machine learning, as crafted AEs should offer similar brightness properties to that of the original image. Therefore, the difference between the  number of detected edges in the adversarial and in the original image is considered as a metric to be minimized.

\BfPara{FFT Analysis}\label{sec:FFT_TOC} 
We consider FFT analysis of the applied changes to the original image as a quality metric. The FFT analysis provides valuable information about the characteristics of an image, e.g., global shape information and sharp changes. Thus, the FFT has been widely applied to decompose an image into its sine and cosine components, and as a feature in image recognition tasks. The FFT analysis can play an important role in the context of adversarial learning, since it reveals valuable perturbation information. In this work, we considered the L2-norm difference of the perturbation's two dimensional FFT, $O_{3}=\left \| FFT2\left( I_{org}-I_{adv} \right) \right \|$, as a metric to be minimized. A smaller value of this metric corresponds to less feature distortion due to the introduced perturbation.

\BfPara{HOG Analysis}\label{sec:HOG_TOC} 
Using HOG analysis we aim to produce AEs with high MR and similar feature descriptors to that of the original image. In this work, we considered the L2-norm difference of HOG feature descriptors of the adversarial sample ${H_{adv}}$ and the original image ${H_{adv}}$, as a metric to be minimized; $O_{4}=\left \| H_{org} - H_{adv} \right \|$. 

\cref{al:MOO_Adv_Three} shows how to produce AEs while incorporating the three IQA metrics, by taking $I_{org}$, $N_{p}$, $w_1$, $w_2$, $w_3$, $w_4$, $maxIter$, and $F$ as inputs, calculating the saliency of the input features, generating the initial population of the crafted samples, calculating the value of each quality metric---including edges $O_2$, FFT analysis $O_3$, and HOG feature descriptor $O_4$---and evaluating the final objective function $O_{f}$, which is a weighted linear combination of these objective $O_{f}=\sum_{i=1}^4 w_{i}O_i$, 
where $O_1$ is the objective corresponding to misclassification rate, while $O_2$, $O_3$, and $O_4$ are objectives corresponding to the aforementioned IQA metrics. Similarly, $w_1$, $w_2$, $w_3$, and $w_4$ are associated weights with the related objective functions. The algorithm iterates until it meets the termination.

\begin{algorithm}[t]
\small
    \SetKwInOut{Input}{Input}
    \SetKwInOut{Output}{Output}

    \underline{Function CC} $(\texttt{$I_{org}$, $N_{p}$, $w_{i=1:4}$, $maxIter$, $F$})$\;
    \Input{$I_{org}$, $N_{p}$, $w_{1}$, $w_{2}$, $maxIter$, $F$}
    \Output{$I_{adv}$}
    
    ${L_{org}}\gets$ argmax$F(I_{org})$
    
    ${E_{org}}\gets$ CED($I_{org}$)
    
    
    ${H_{org}}\gets$ HOG($I_{org}$)
    
    calculate saliency score of $I_{org}$ (based on~\autoref{eq:saliency})
    
    select pixels associated with $n$ top scores
    
    initialize PSO parameters
    
    generate initial population of crafted samples ($I_{adv}$) by adding random perturbation to selected $n$ pixels 
    
    ${L_{adv}}\gets$ argmax$F(I_{adv})$
    
    ${E_{adv}}\gets$ CED($I_{adv}$)
    
    
    ${H_{adv}}\gets$ HOG($I_{adv}$)
    
    \eIf{${L_{org}} \neq {L_{adv}} $}
      {
      $O_{1}=\left \| I_{org} - I_{adv} \right \|$
      
      $O_{2}=\left | E_{org} - E_{adv} \right |$
      
      
      $O_{3}=\left \| FFT2(I_{org}-I_{adv}) \right \|$
      
      $O_{4}=\left \| H_{org} - H_{adv} \right \|$
      
      $O_{f}=\sum_{i=1}^4 w_{i}O_i$
      
      evaluate the objective $O_f$ of all crafted samples
      
      $G_{best} \gets$ global best sample
    
        \eIf{\texttt{Termination criteria is satisfied}}
        {
        return $G_{best}$
        }
        {
        generate new population of crafted samples (based on~\autoref{eq:pso_velocity} and~\autoref{eq:pso_position})
        
        go to 7
        }      
        }
        {
        generate new population of crafted samples
      
        go to 7
        }
    
    \caption{A sample algorithm for generating AEs based on \ours. Here we consider all of the three IQA metrics within the process to optimize the quality of the generated AEs.}
    \label{al:MOO_Adv_Three}
\end{algorithm}

\section{Experiments and Evaluation}\label{sec:Eval_Discussion}


To demonstrate our method, we conducted an extensive set of experiments using benchmarks outlined in \cref{sec:dataset}. Moreover, we compared the results obtained from the proposed approach with that of various adversarial attack methods, listed in~\cref{sec:adv_attacks}. In \cref{sec:exp_setup}, we describe the implementation of the different aspects of our approach, including the deep model architecture, particle swarm optimization algorithm, attack methods, and the quality metrics. The main results and a discussion are in~\cref{sec:results}. A human perception study is outlined in \cref{sec:human}, followed by transferability analysis in section~\ref{sec:transfer}, and detection in~\ref{sec:detection}

\subsection{Main Datasets}\label{sec:dataset}
We used multiple popular image classification benchmarks, including MNIST hand-written digits~\cite{lecunBBH98}, Fashion MNIST~\cite{xiao2017online}, CIFAR-10~\cite{KrizhevskyH09}, GTSRB~\cite{stallkamp2011german}, and Open Image V5~\cite{OpenImage}, by Google AI, to evaluate the performance of our approach. A brief description of these datasets are provided in below. Note that in all of our experiments we consider only the samples that were classified correctly by the DL model for the AE generation. 

\BfPara{MNIST-Digits} The Mixed National Institute of Standards and Technology (MNIST) hand-written digits benchmark is a set of labeled gray-scale images consisting of 50K training samples and test set of 10K samples. The images are normalized, centered, and have a fixed size of 28x28 pixels. Each image is labeled as $\{0, ...,9\}$. MNIST-Digits is a popular database for image recognition tasks~\cite{lecunBBH98}. 

\BfPara{Fashion MNIST} The Fashion MNIST benchmark is a set of labeled gray-scale images consisting of 60K training samples and a test set of 10K samples. The images are normalized, centered and have a fixed size of 28x28 pixels. Each image is labeled as $\{0, ...,9\}$. Fashion MNIST is popular for image processing and classification tasks~\cite{xiao2017online}. 

\BfPara{CIFAR-10} The Canadian Institute For Advanced Research (CIFAR-10) dataset is a subset of 80M tiny images. It consists of 60K low-resolution 32x32 images in 10 different classes, including cars, birds, cats, dogs, frogs, horses, and trucks. CIFAR-10 dataset is widely used in machine learning and computer vision classification tasks~\cite{KrizhevskyH09}.

\BfPara{GTSRB-10} The German Traffic Sign Recognition Benchmark (GTSRB)~\cite{stallkamp2011german} is a popular multi-class  classification dataset. GTSRB has more than 39K training images of various sizes, and more than 12K test images. The images are very similar to real-life data, with 43 different traffic signs. Accurate traffic sign recognition is essential for autonomous vehicles, specifically for self-driving cars.

\BfPara{Open Image}
In order to understand the impact of the image resolution on the performance of \ours and other methods, we conducted several experiments using the aforementioned attack methods on a CNN model trained over the Open Image Dataset V5~\cite{OpenImage}; a large dataset by Google AI containing more than 8.9M high resolution images covering almost 20K class labels. To set out our experiments and considering multiple factors such as {\em time constraints} and comparability with other utilized benchmarks in this study, we conducted our experiments on a set of 10 class labels, selected at random.



\subsection{AE Generation Methods}\label{sec:adv_attacks}

For a better perspective on the performance of \ours, we compared our results with state-of-the-art adversarial attack methods. We chose  five generic algorithms already published and well-known in the community, including the Fast Gradient Sign Method (FGSM), due to Goodfellow \etal~\cite{Goodfellow2015Explaining}, Jacobian-based Saliency Map Approach (JSMA), due to Papernot \etal~\cite{PapernotMJFCS16} C\&W, due to Carlini and Wagner~\cite{Carlini017}, Projected Gradient Descent (PGD), due to Madry \etal~\cite{MadryMSTV17}, and Momentum Iterative Method (MIM), due to Dong \etal~\cite{dong2018boosting}. We briefly describe these methods in~\autoref{app:1}, and refer the interested reader to original presenting sources.

\subsection{Main Experimental Setup}\label{sec:exp_setup}

We built deep learning models based on the CNN architecture trained over the benchmarks in \cref{sec:dataset}. The trained models is considered as as baseline when generating AEs. Next, we incorporate the PSO algorithm along with the IQA metrics to produce AEs as outlined in~\cref{sec:Methedology}. In this section, we outline the implementation of our proposed approach in detail.

\BfPara{Evaluation system} Experiments are conducted using Python 3.6.0 run over Ubuntu 16.04 using a system with an I5-8500 CPU @ 3.00 GHz, with 32 GB DDR4 of RAM, 512 GB SSD for storage, as well as NVIDIA Titan RTX, RTX 2080Ti, and GTX 980Ti (GPUs).

\BfPara{Model Architecture}
The CNN we used in our experiments has multiple consecutive convolutional layers with ReLu activation, followed by a fully connected layer outputting softmax values for each class. We tested different values for the number of convolutional layers, batch size, and epochs to improve accuracy rate of the trained networks over the aforementioned datasets. We achieved 99.12\% and 98.94\% accuracy rate for the MNIST hand-written digits and Fashion MNIST test datasets, respectively, after 50-epochs training and batch size of 150 with three convolutional layers. Moreover, we trained two more CNN models on CIFAR-10 and GTSRB datasets with six convolutional layers, which achieved a detection accuracy rate of 86.61\% and 92.24\% on the test samples, respectively. Further details about CNNs can be found in~\cite{krizhevsky2012imagenet}.

\BfPara{Canny Edge Detector}
For edge analysis, we used CED implementation from~\textit{scikit-image}~\cite{scikit_canny}. The algorithm uses a 2D grayscale image $I$, the standard deviation value $\sigma$ of a Gaussian filter, and values of lower and higher thresholds as inputs, and provides the binary edge map of the given image as an output. The performance of CED is highly dependent on three main parameters: $\sigma$, $t_{low}$, and $t_{high}$. In our experiments, we considered $\sigma = 2$, $t_{low} = 0.6$ and $t_{high} = 0.8$. Note that proper values of $t_{low}$ and $t_{high}$ should be determined empirically.

\BfPara{Fast Fourier Transform}
The 2D discrete Fourier transform allows mapping images onto their spatial frequencies domain, consequently analyzing the perturbations with regard to their spatial frequencies, where a high frequency corresponds to details of the features and sharp changes in the values while lower frequencies provide information about the global shape. Because well-designed AEs are expected not to change the global shape, we include lower spatial frequencies in our analysis. We argue that if the global shape of the image remains unaffected they should result in similar lower spatial frequencies. However, if the perturbation affects the general shape, then discarding this part of the frequencies would lead to loss of critical information. Therefore, in this study we consider the L2-norm difference of the perturbation's FFT as a metric to evaluate the quality of the AEs. The smaller the L2-norm difference, the higher the quality of the AEs.

\BfPara{Histogram of Oriented Gradient}
To compute the HOG feature descriptor vector of a given image we used the HOG function provided by \textit{scikit-image}~\cite{scikit_HOG}. The algorithm takes as an input the image $I$, the number of orientations, number of pixels per cell, number of cells per block, and the normalization method, and calculates the HOG feature descriptor vector. In our experiments we considered 9 channels in each histogram, cell size of $8 \times 8$ pixels, block size of $2 \times 2$ cells, and L2-norm normalization method to calculate the HOG feature descriptor of a given image.

\BfPara{Attacks} 
We used the \textit{Cleverhans}~\cite{papernot2018cleverhans} library for the implementation of FGSM, C\&W, MIM, PGD, and JSMA. Cleverhans is a Python library for conducting adversarial attacks and  building defenses on machine/deep learning systems.

\BfPara{Particle Swarm Optimization \& Configuration}
There are multiple convergence topologies in the literature, \eg ring, star, \etc In this study, we used the ring topology where particles are connected with their neighbors. Star topology performs better in the local best scenario of the PSO. Note that both local best and global best scenarios of the PSO are similar in the sense that the social component of the velocity updates causes both to move towards the global best particle. However, as the local best scenario is less susceptible to being trapped in local minima and performs better, we used it in this study. In order to deal with the velocity explosion problem, we used the clamping technique. The performance of the PSO algorithm depends on suitable selection of the parameters' values and their adjustment along the search process. In the PSO algorithm, the number of particles $N_{p}$ and the maximum total number of iterations $maxIter$ play a key role in the optimization process and are generally problem-dependent. Proper values of PSO parameters can lead to better convergence, fewer iterations, shorter running time, and better balance between local and global search. In our experiments, we set the number of swarms to be $N_{p}=300$ and the maximum number of iterations to be $maxIter=200$. While the values of  $c_1$ and $c_2$ were set empirically, $c_1 + c_2 \leqslant 4$, $r1$ and $r2$ are random numbers drawn from a uniformly distributed in the range of $[0 \ 1]$, to maintain the diversity of the population. 

\subsection{Main Results and Discussion}\label{sec:results}
We considered two different configurations in our experiments, namely SOO and MOO configurations. We considered several constraints in our experiments, such as perturbing a very small number of the pixels (at most 30---out of 784 pixels---in MNIST datasets, 70 pixels in CIFAR-10, 90 pixels in GTSRB, and 400 pixels in Open Image dataset), high misclassification rate of generated AEs, and improving the quality of the generated AEs in terms of IQA metrics.
To overcome these constraints, we first use the saliency map algorithm introduced in~\cite{PapernotMJFCS16} to rank pixels based on their saliency score. The saliency map of a given input image for different datasets are shown in~\cref{fig:SalielncyScore}. Larger values of saliency score correspond to pixels with high impact on the model output. Thus, we argue that perturbing those pixels would likely lead to misclassification.  

\begin{figure*}[!h]
\hfill
\centering
\begin{minipage}{0.99\textwidth}
\begin{subfigure}[Original]{\includegraphics[width=2.3cm]{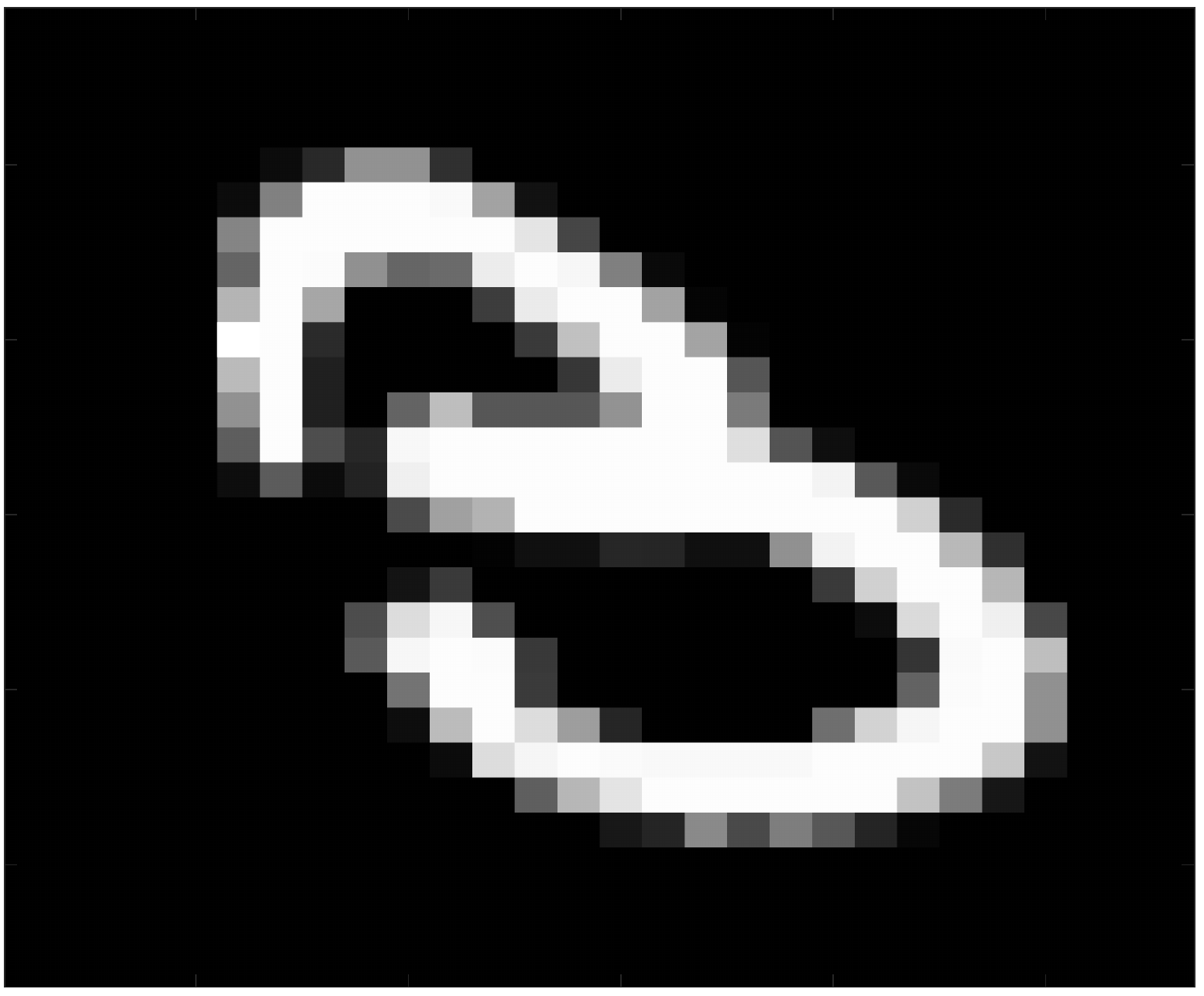}} 
\hfill
\end{subfigure}
\begin{subfigure}[FGSM]{\includegraphics[width=2.3cm]{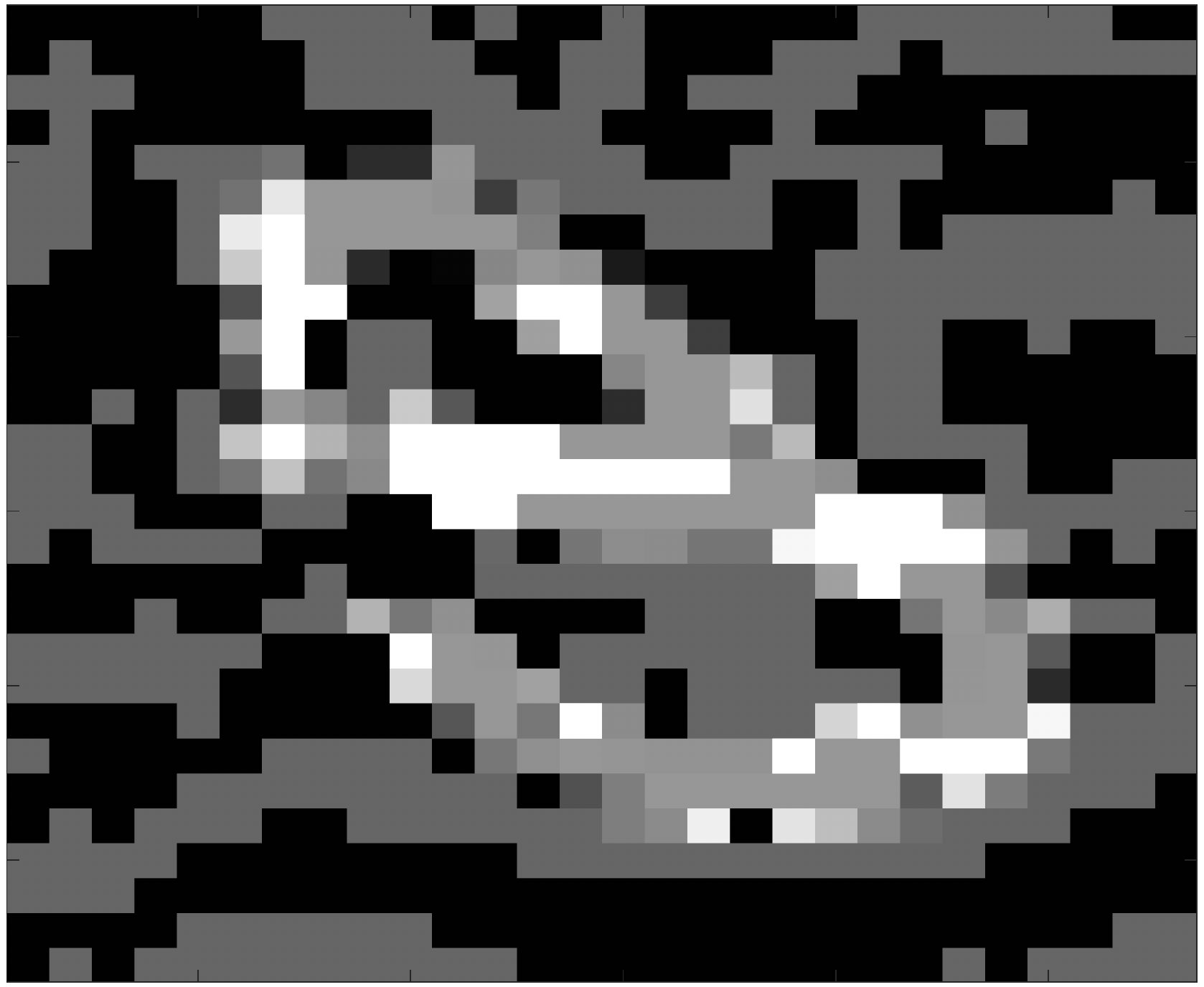}} 
\hfill
\end{subfigure}
\begin{subfigure}[C\&W]{\includegraphics[width=2.3cm]{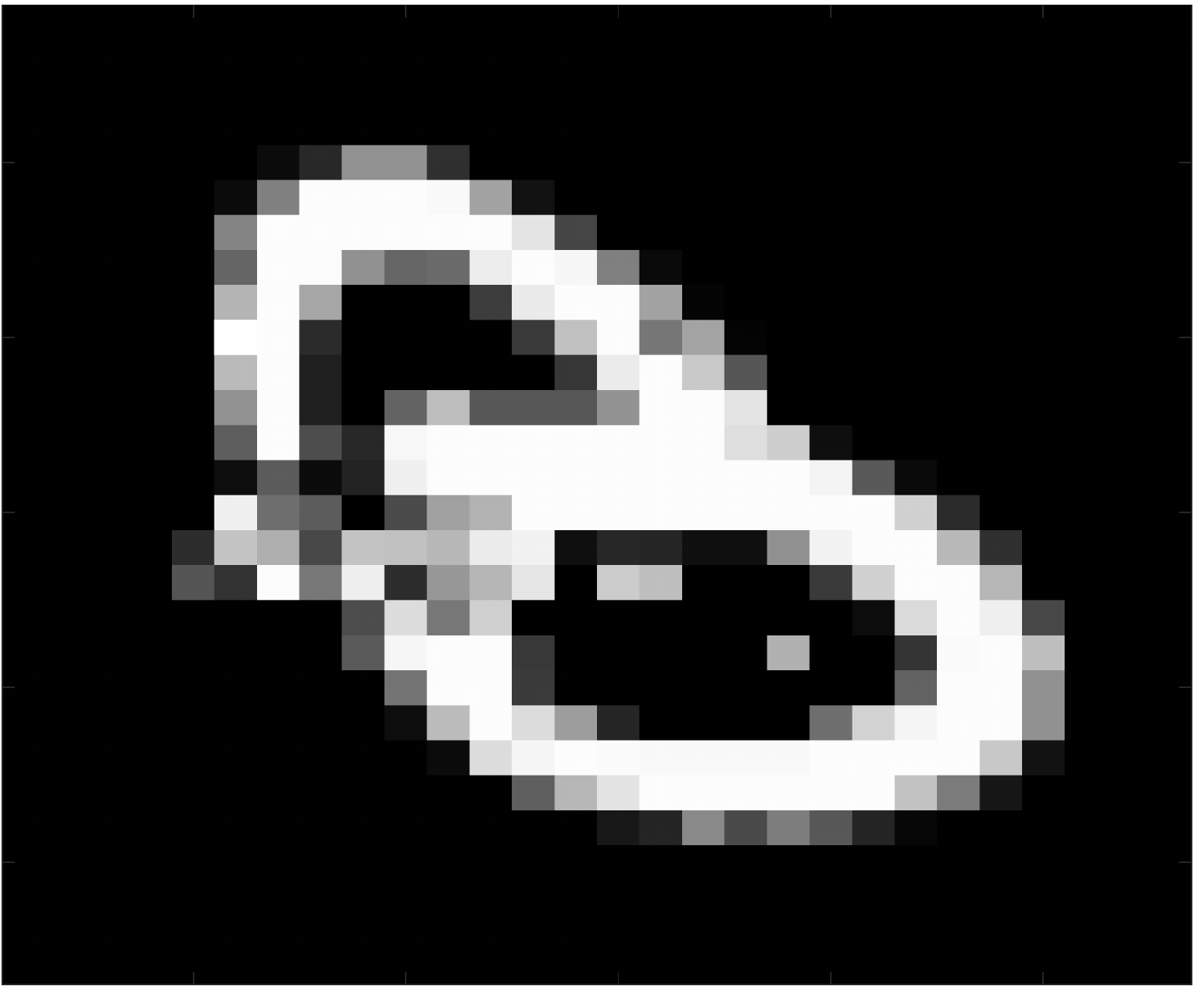}} 
\hfill
\end{subfigure}
\begin{subfigure}[PGD]{\includegraphics[width=2.3cm]{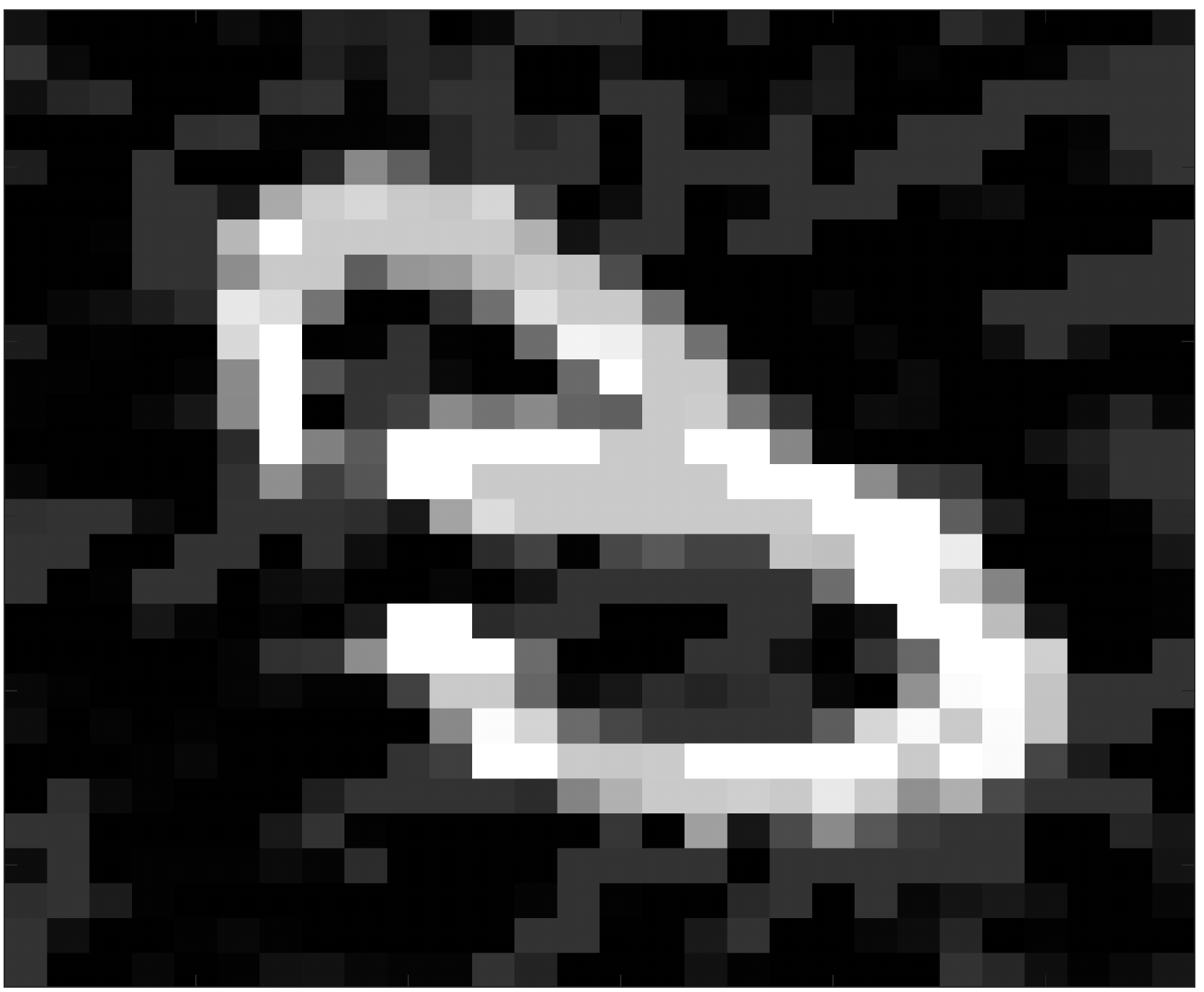}}
\hfill
\end{subfigure}
\begin{subfigure}[MIM]{\includegraphics[width=2.3cm]{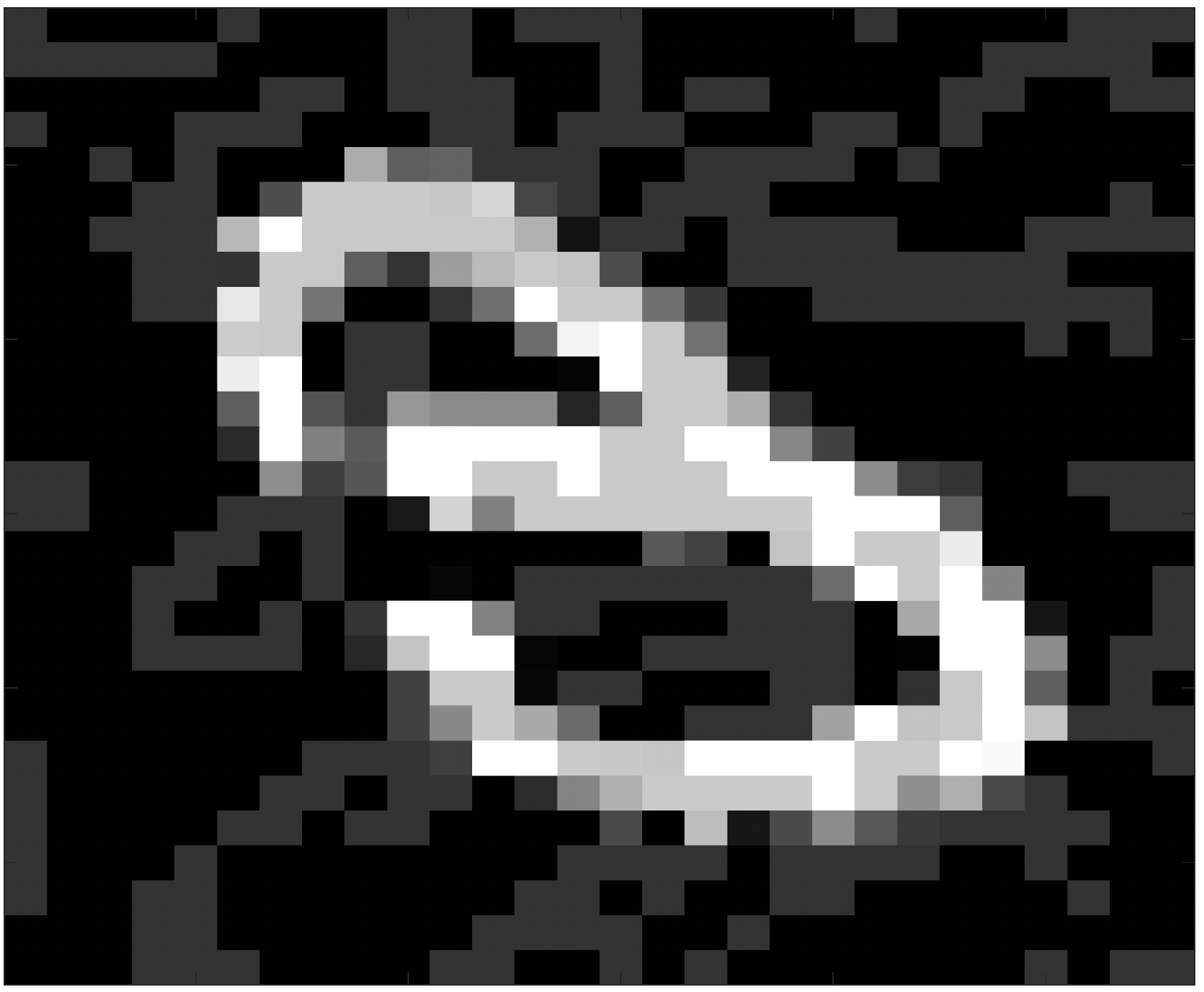}} 
\hfill
\end{subfigure}
\begin{subfigure}[JSMA]{\includegraphics[width=2.3cm]{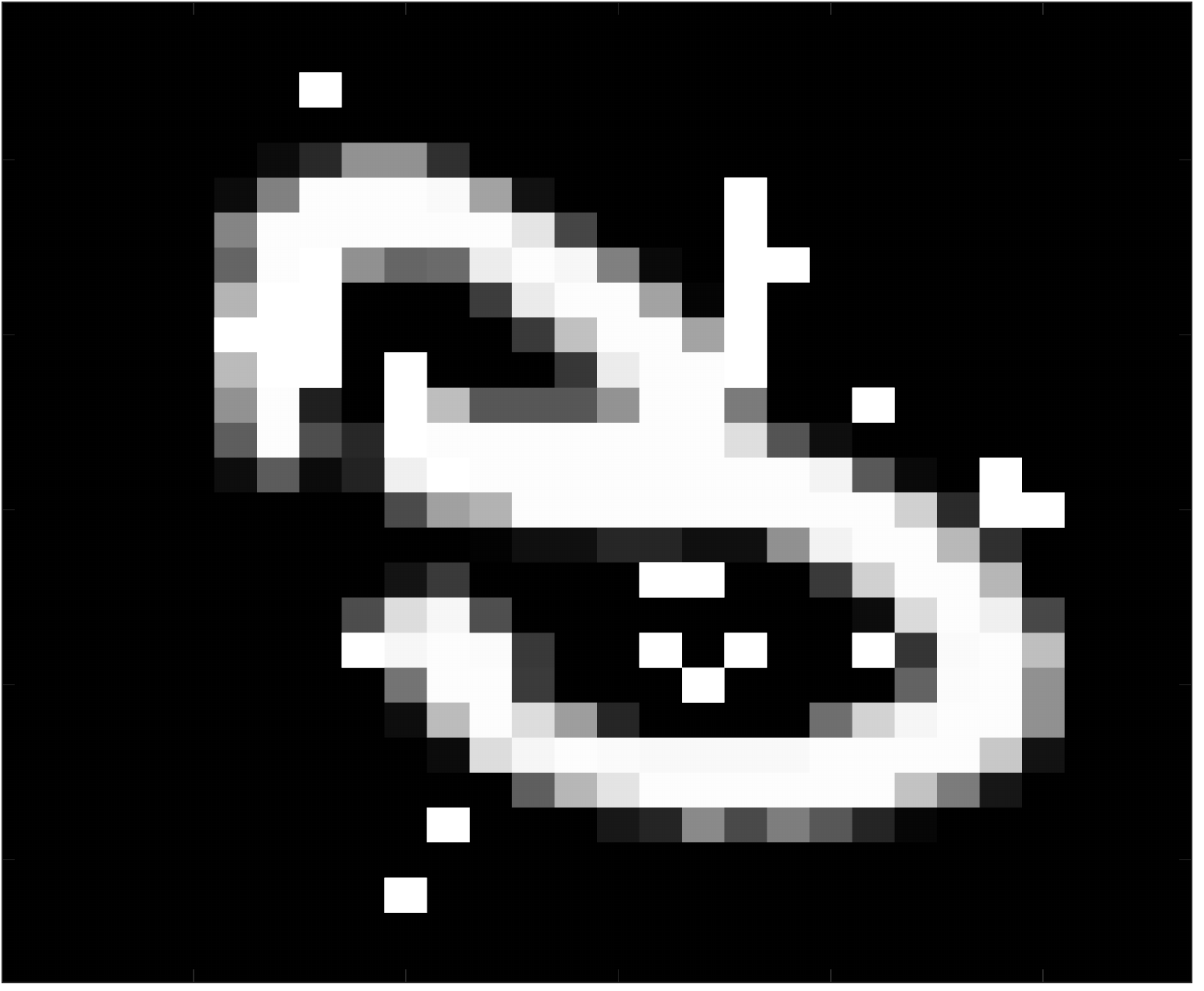}} 
\hfill
\end{subfigure}
\begin{subfigure}[\ours]{\includegraphics[width=2.3cm]{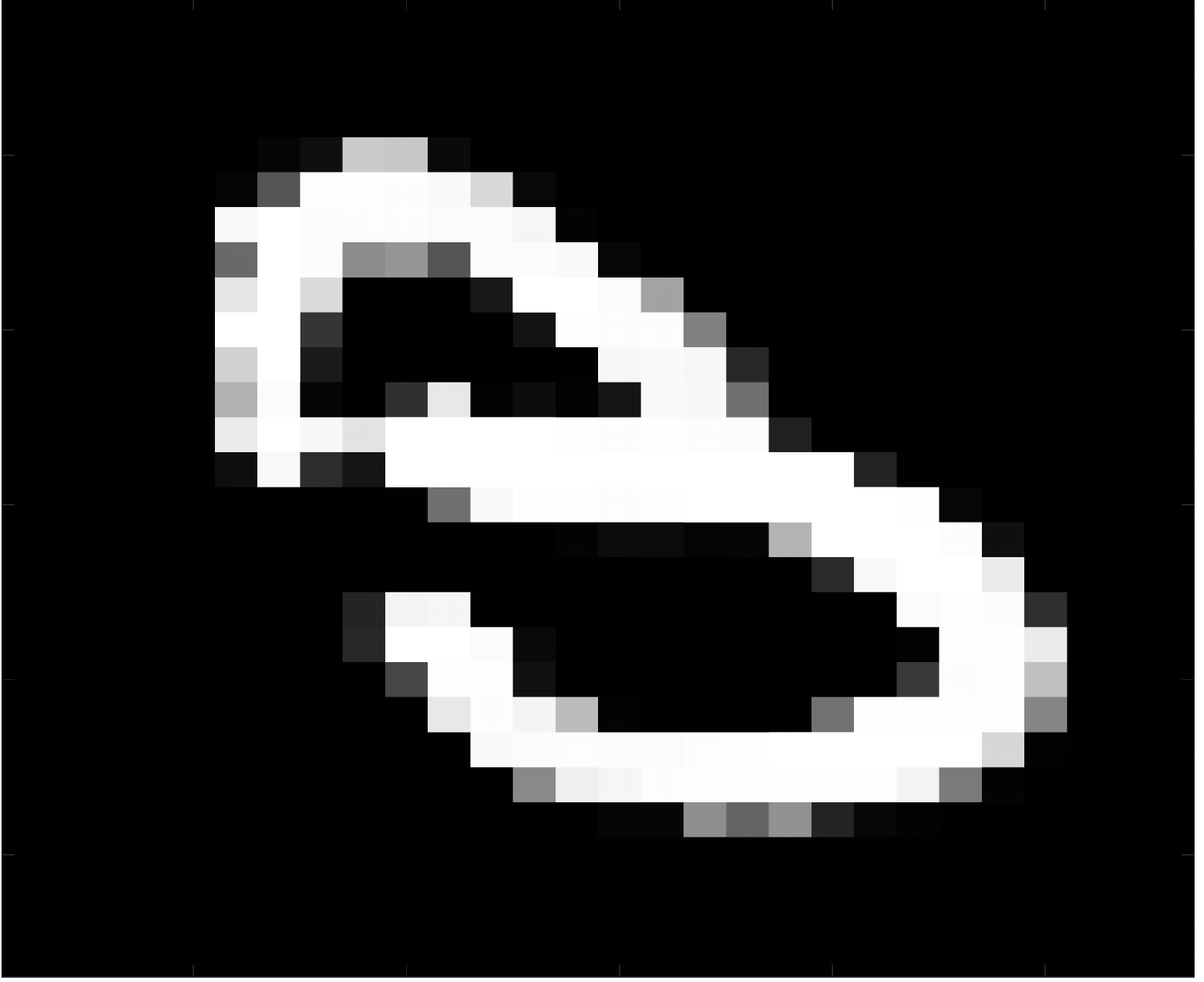}} 
\hfill
\end{subfigure}\vspace{-3mm}
\caption{Sample AEs generated for MNIST hand-written digits (untargeted attack). Our approach generates AEs by perturbing only 30 out of 784 pixels in a given image, while maintaining close IQA metrics to the original input.}
\label{fig:Org_Adv_samples_Digits}

\end{minipage}

\begin{minipage}{0.99\textwidth}
\begin{subfigure}[Original]{\includegraphics[width=2.3cm]{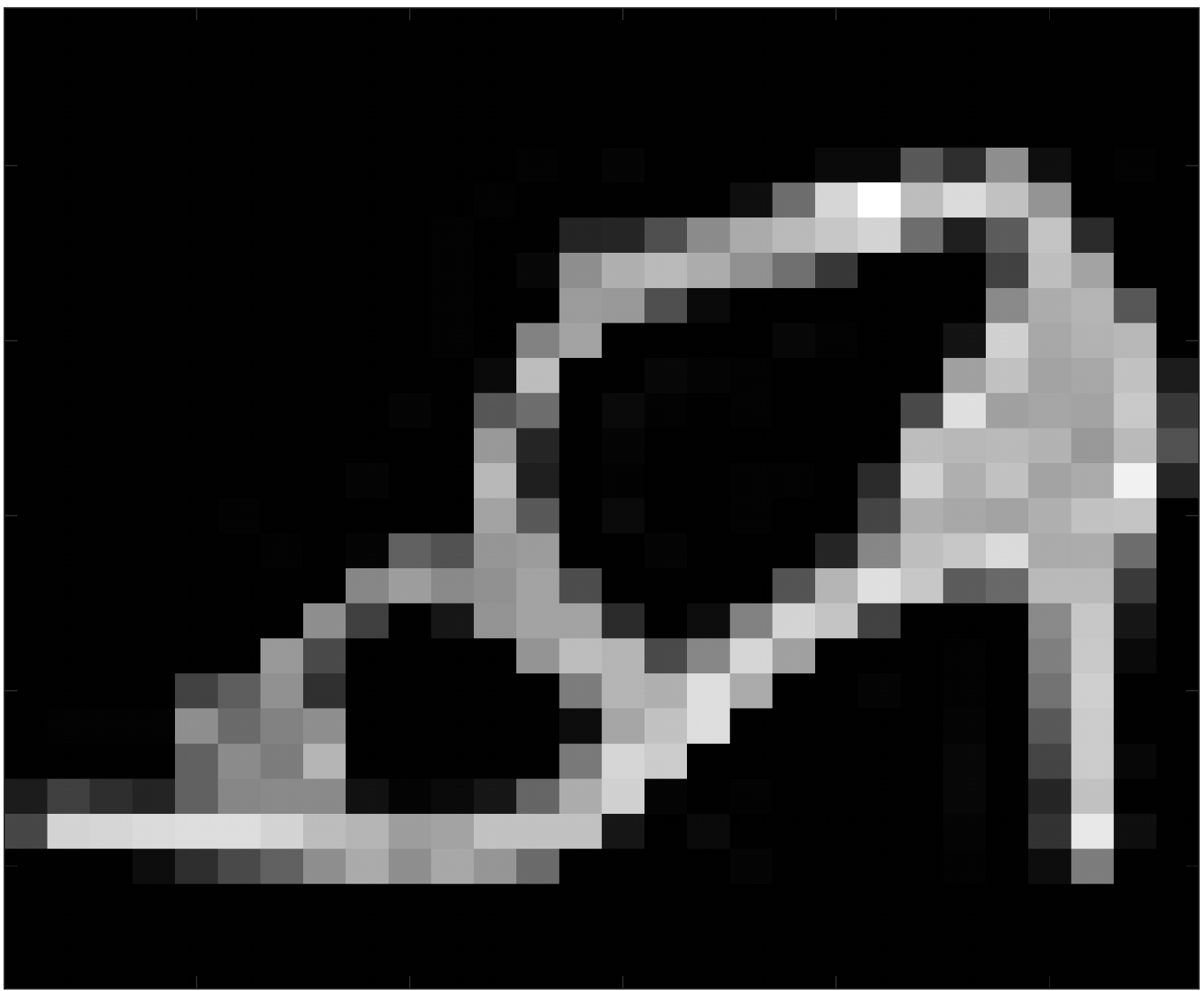}} 
\hfill
\end{subfigure}
\begin{subfigure}[FGSM]{\includegraphics[width=2.3cm]{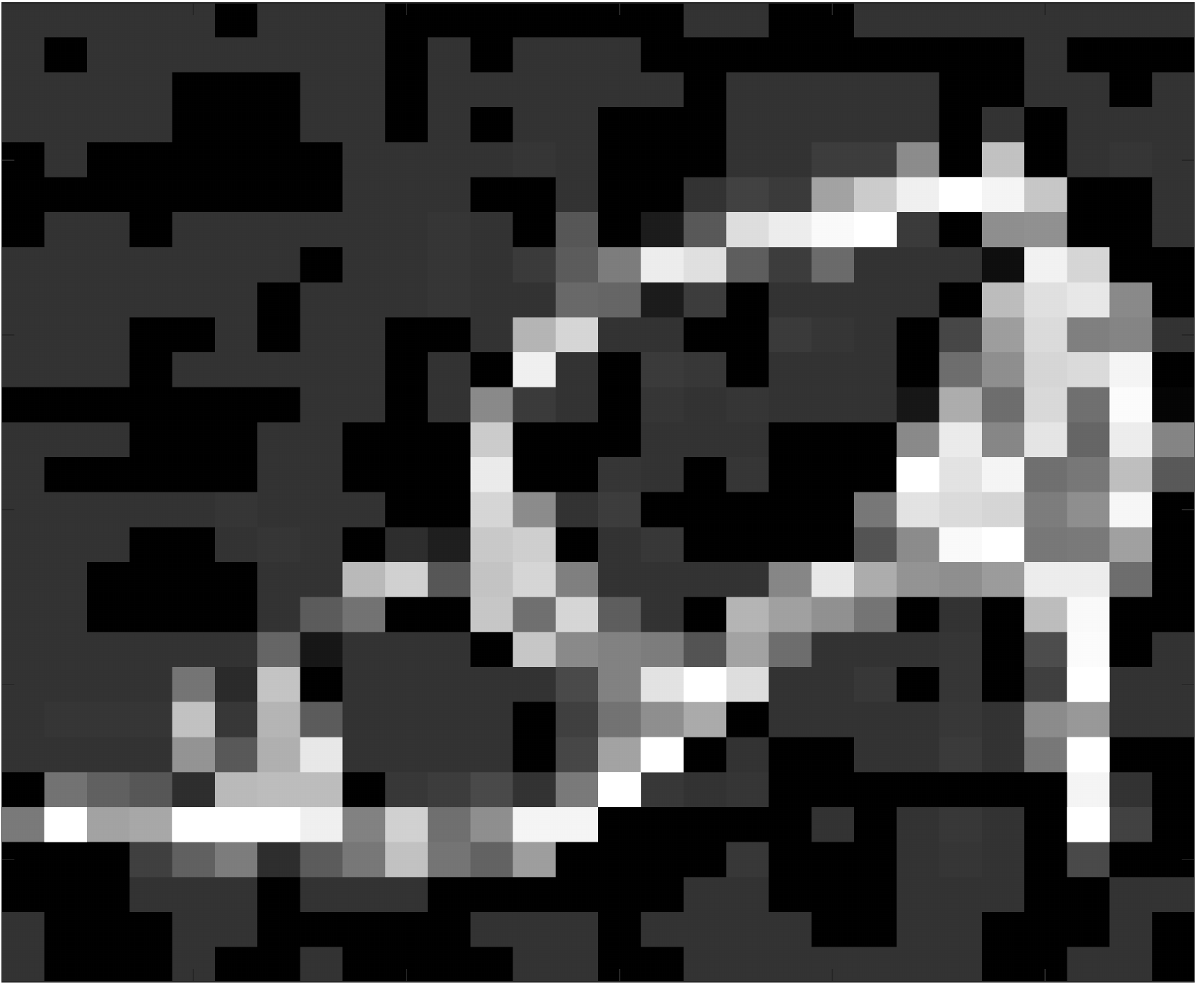}} 
\hfill
\end{subfigure}
\begin{subfigure}[C\&W]{\includegraphics[width=2.3cm]{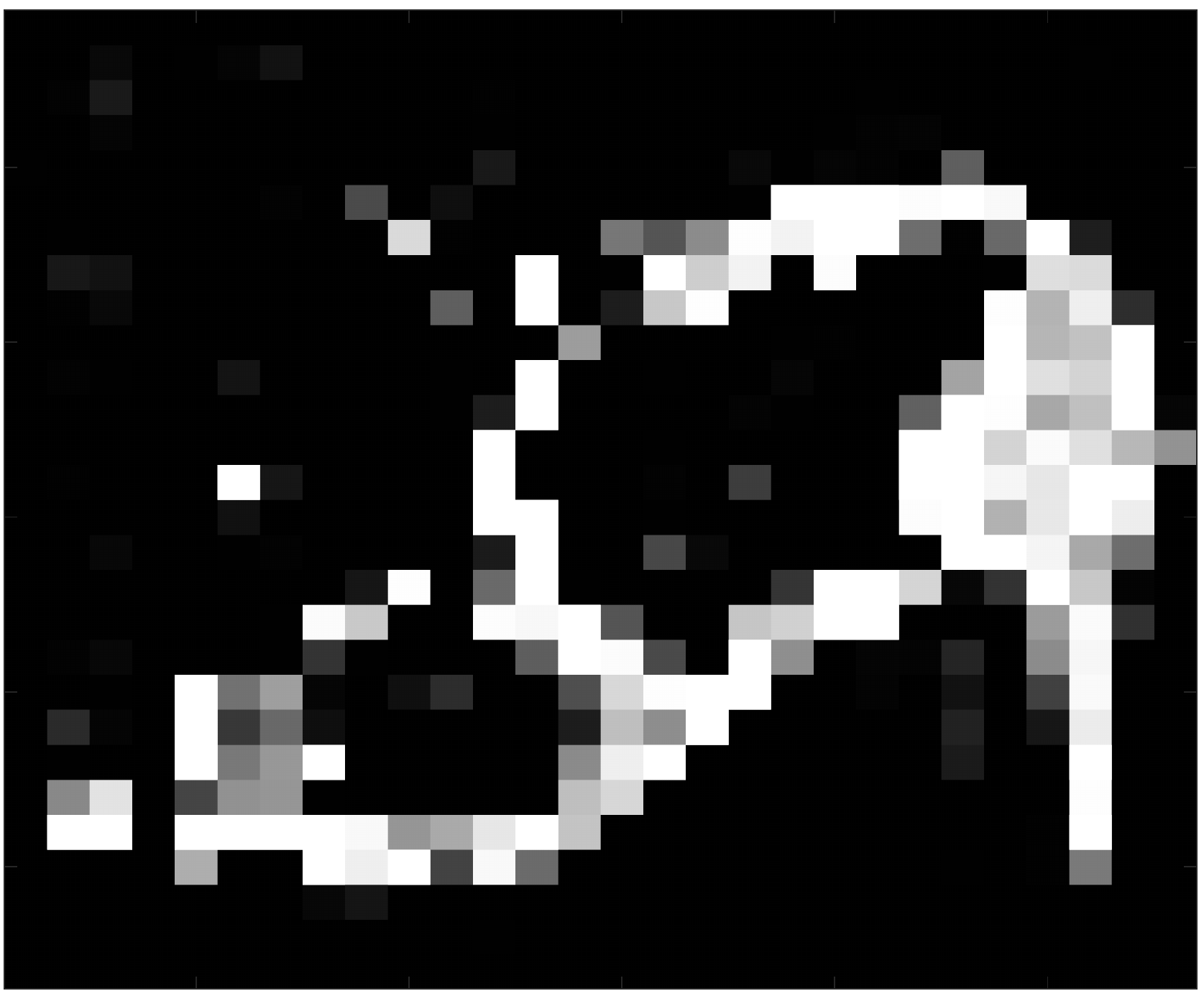}} 
\hfill
\end{subfigure}
\begin{subfigure}[PGD]{\includegraphics[width=2.3cm]{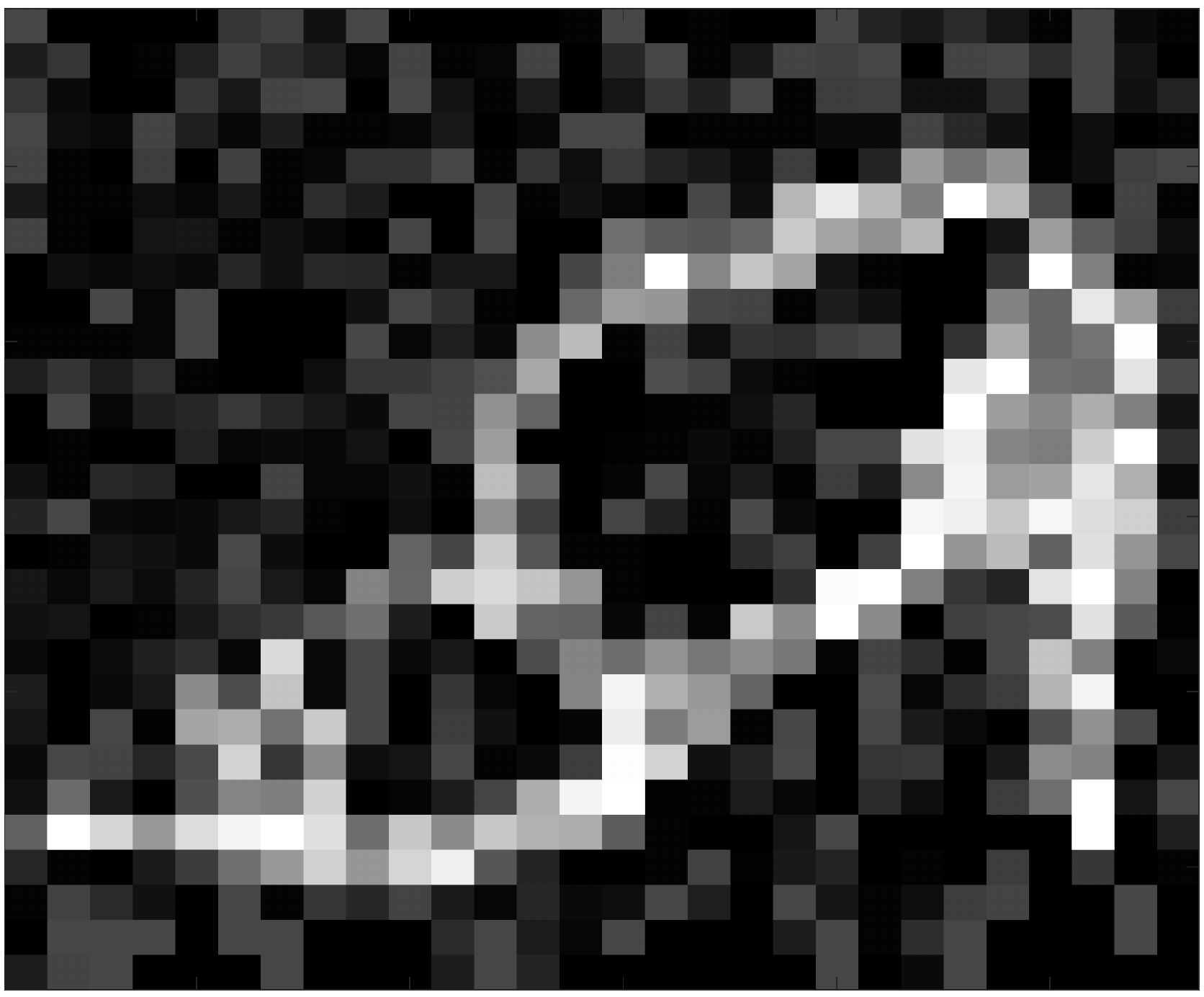}}
\hfill
\end{subfigure}
\begin{subfigure}[MIM]{\includegraphics[width=2.3cm]{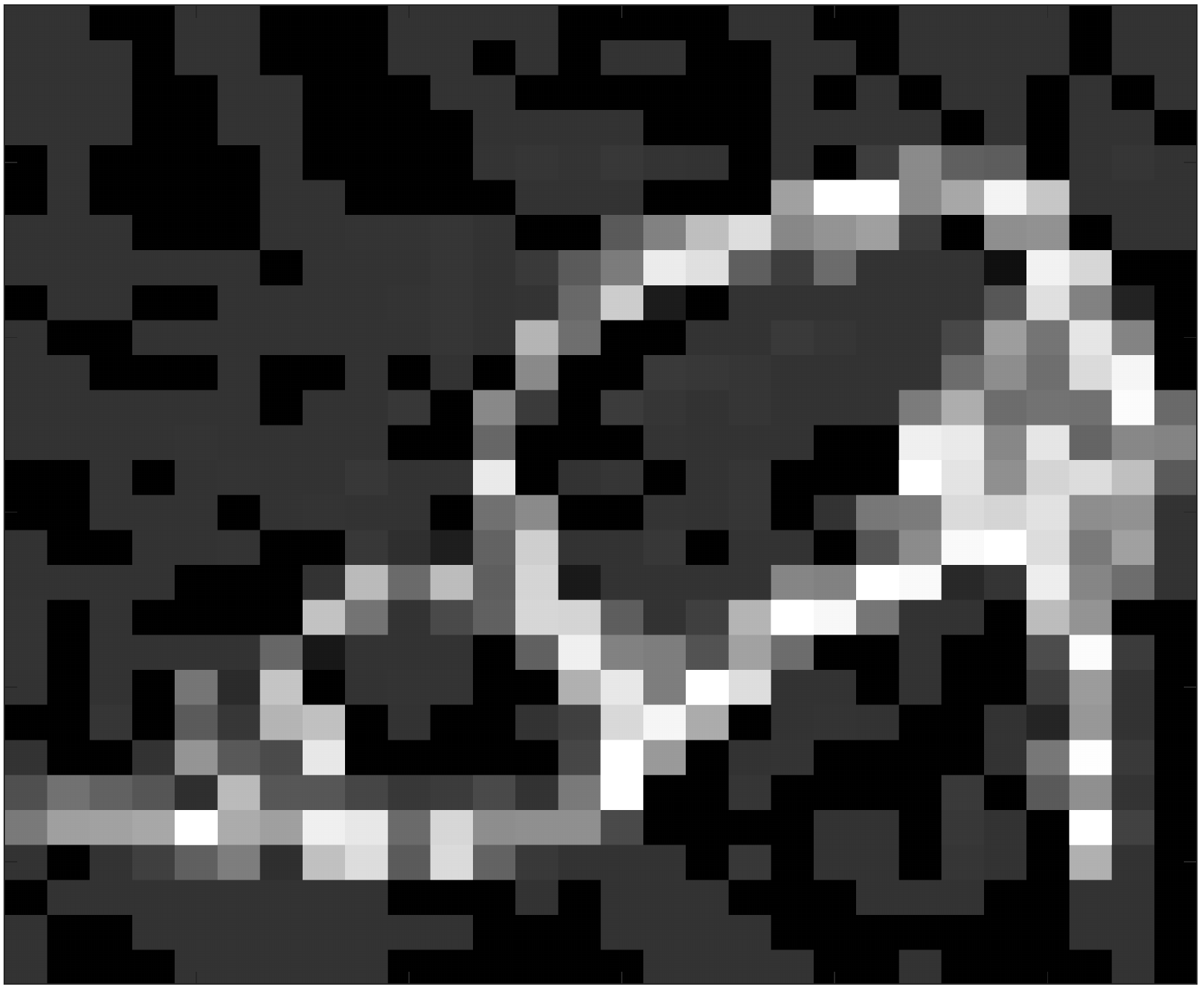}} 
\hfill
\end{subfigure}
\begin{subfigure}[JSMA]{\includegraphics[width=2.3cm]{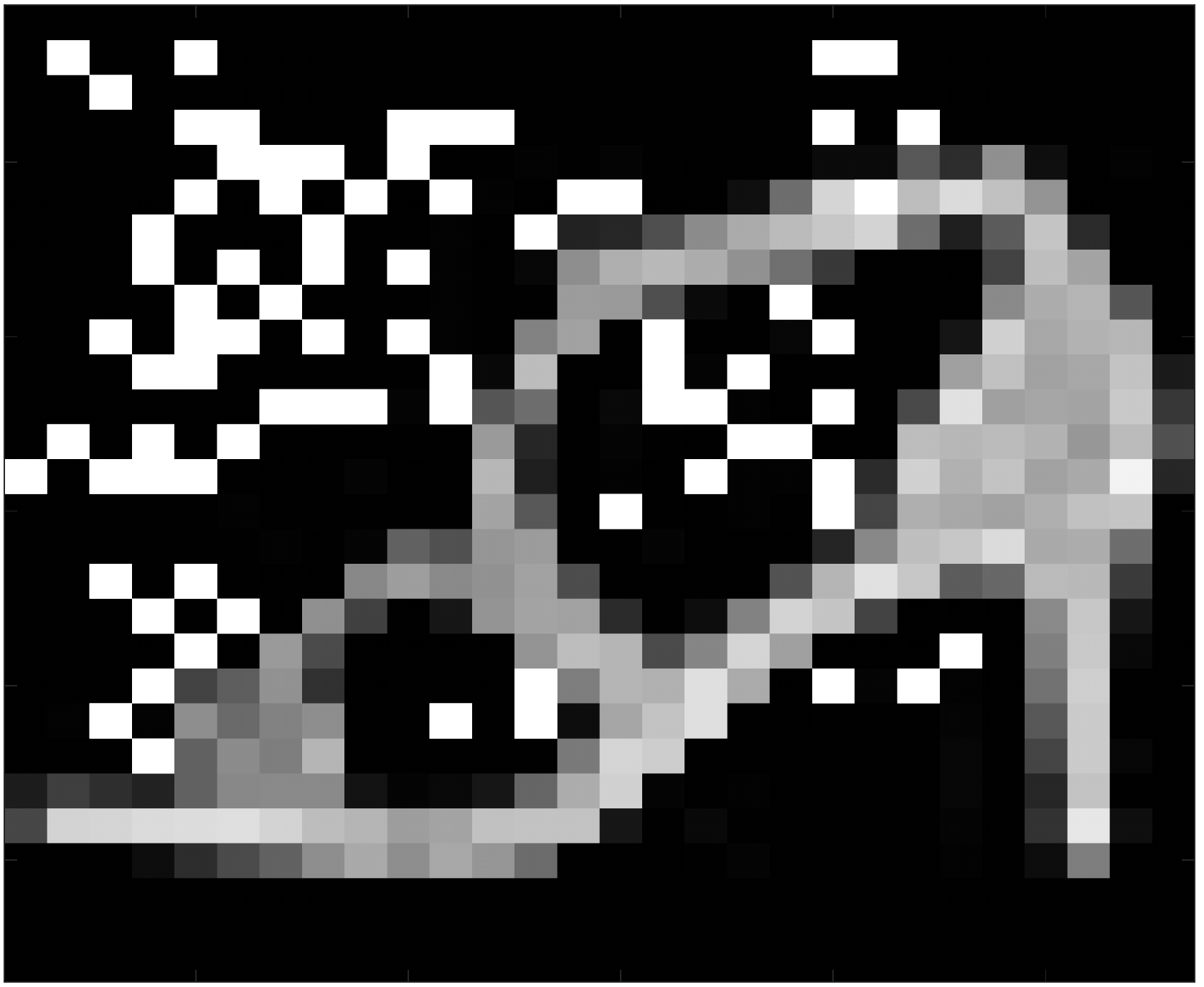}} 
\hfill
\end{subfigure}
\begin{subfigure}[\ours]{\includegraphics[width=2.3cm]{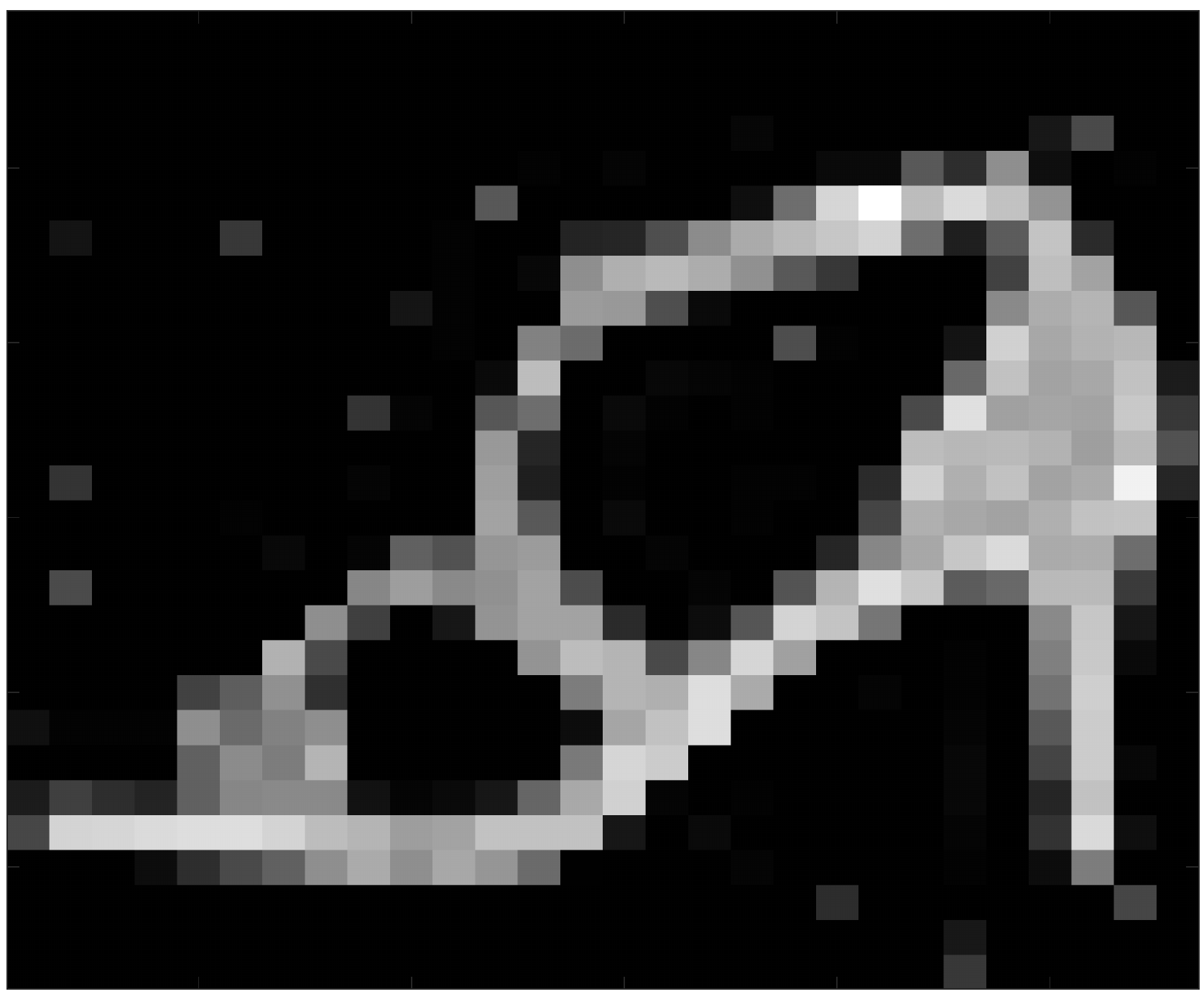}} 
\hfill
\end{subfigure}\vspace{-3mm}
\caption{Sample AEs generated for fashion MNIST dataset. Our approach produces a misclassification rate of more than 97.40 by perturbing $3.82\%$ of the pixels in a given greyscale image on average while showing high IQA metrics.}
\label{fig:Org_Adv_samples_Fashion}

\end{minipage}

\begin{minipage}{0.99\textwidth}
\begin{subfigure}[Original]{\includegraphics[width=2.3cm]{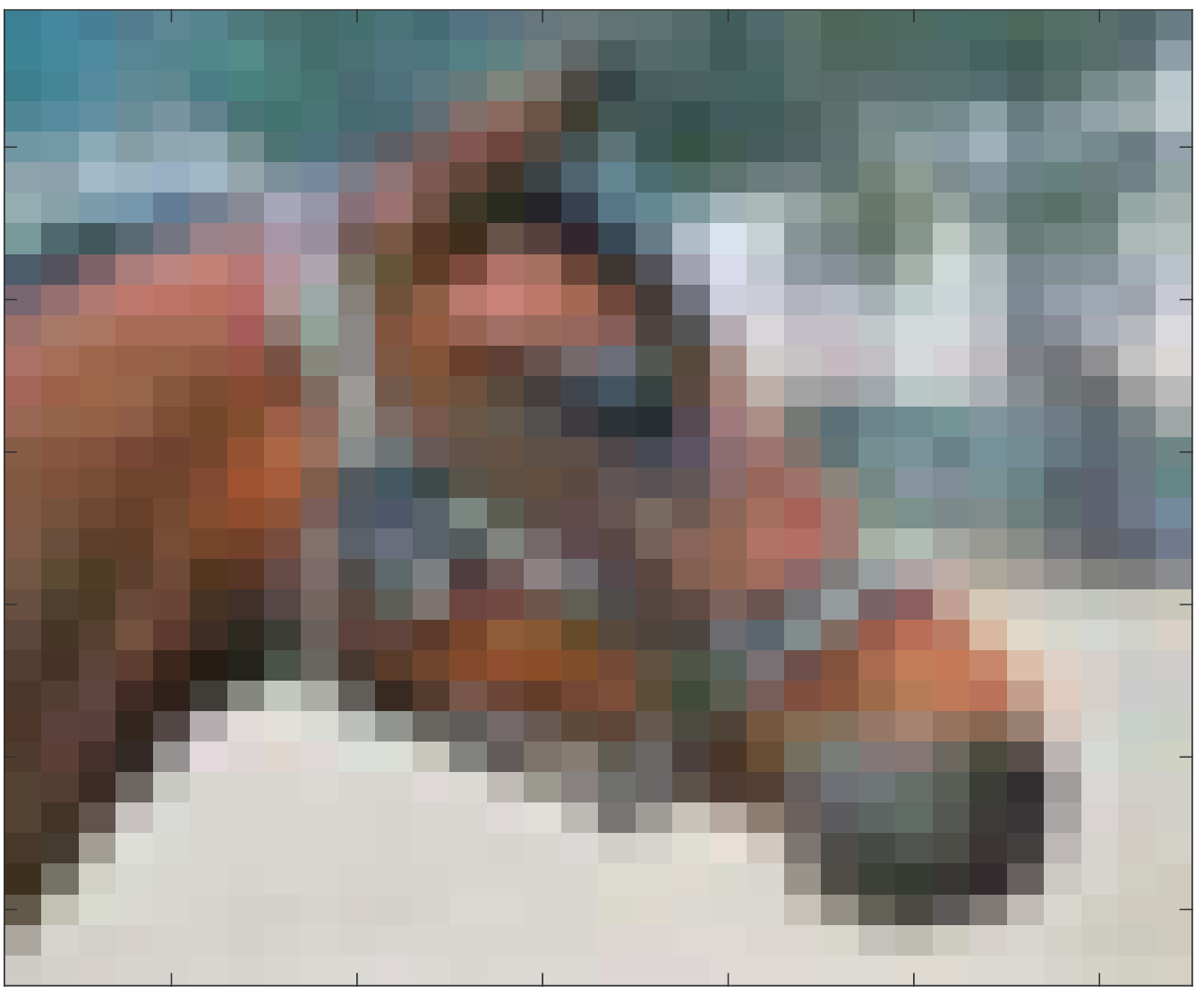}} 
\hfill
\end{subfigure}
\begin{subfigure}[FGSM]{\includegraphics[width=2.3cm]{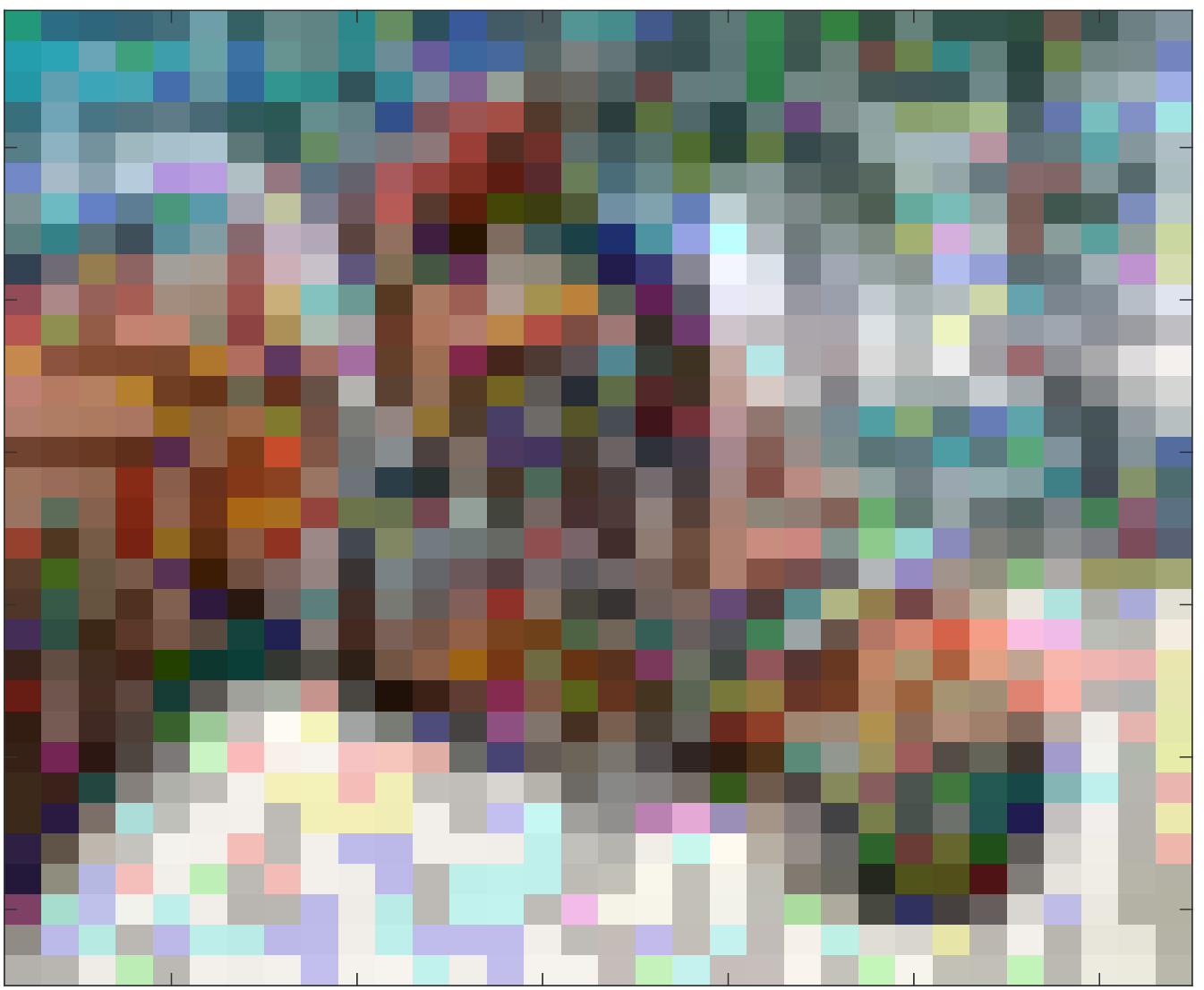}} 
\hfill
\end{subfigure}
\begin{subfigure}[C\&W]{\includegraphics[width=2.3cm]{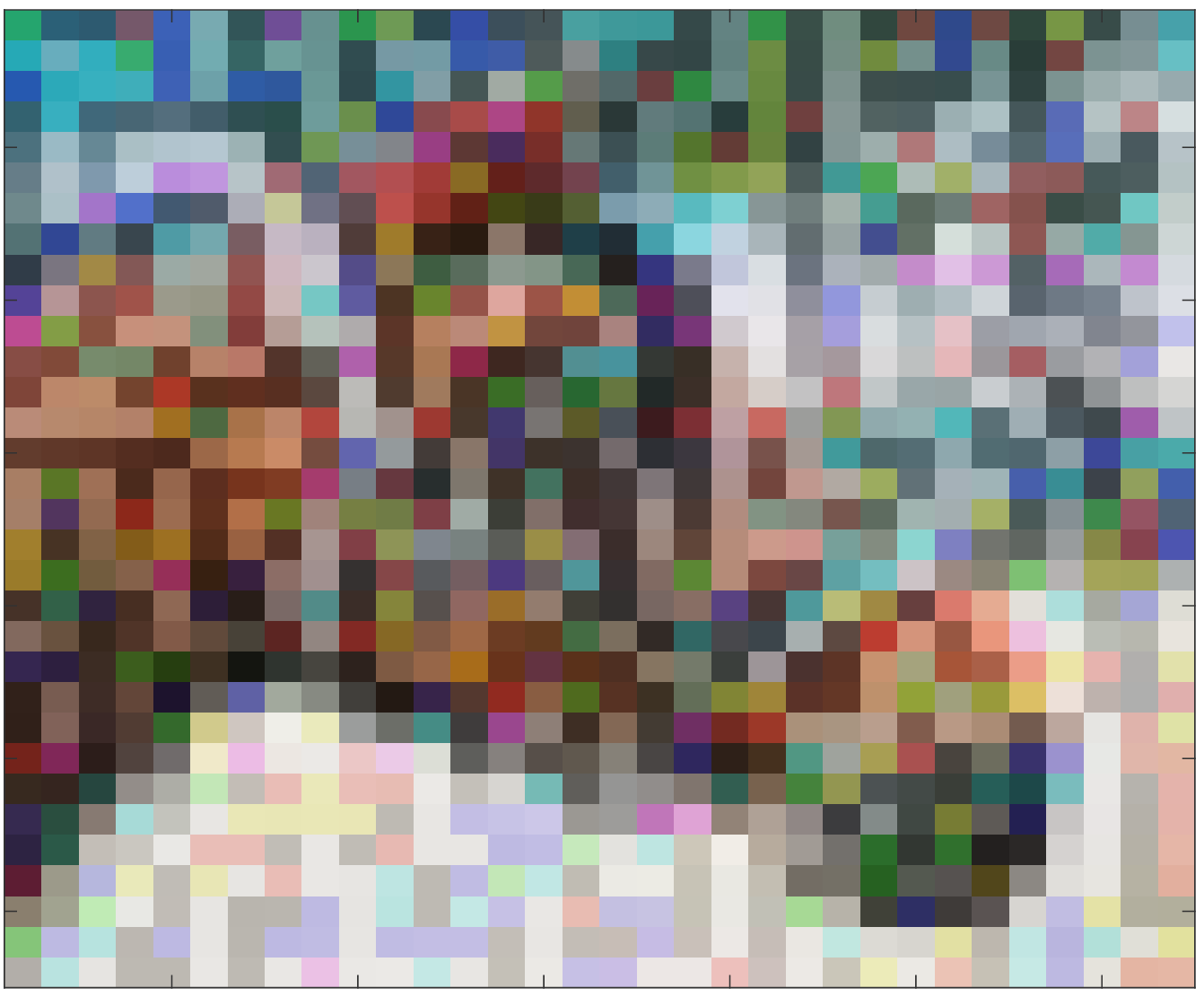}} 
\hfill
\end{subfigure}
\begin{subfigure}[PGD]{\includegraphics[width=2.3cm]{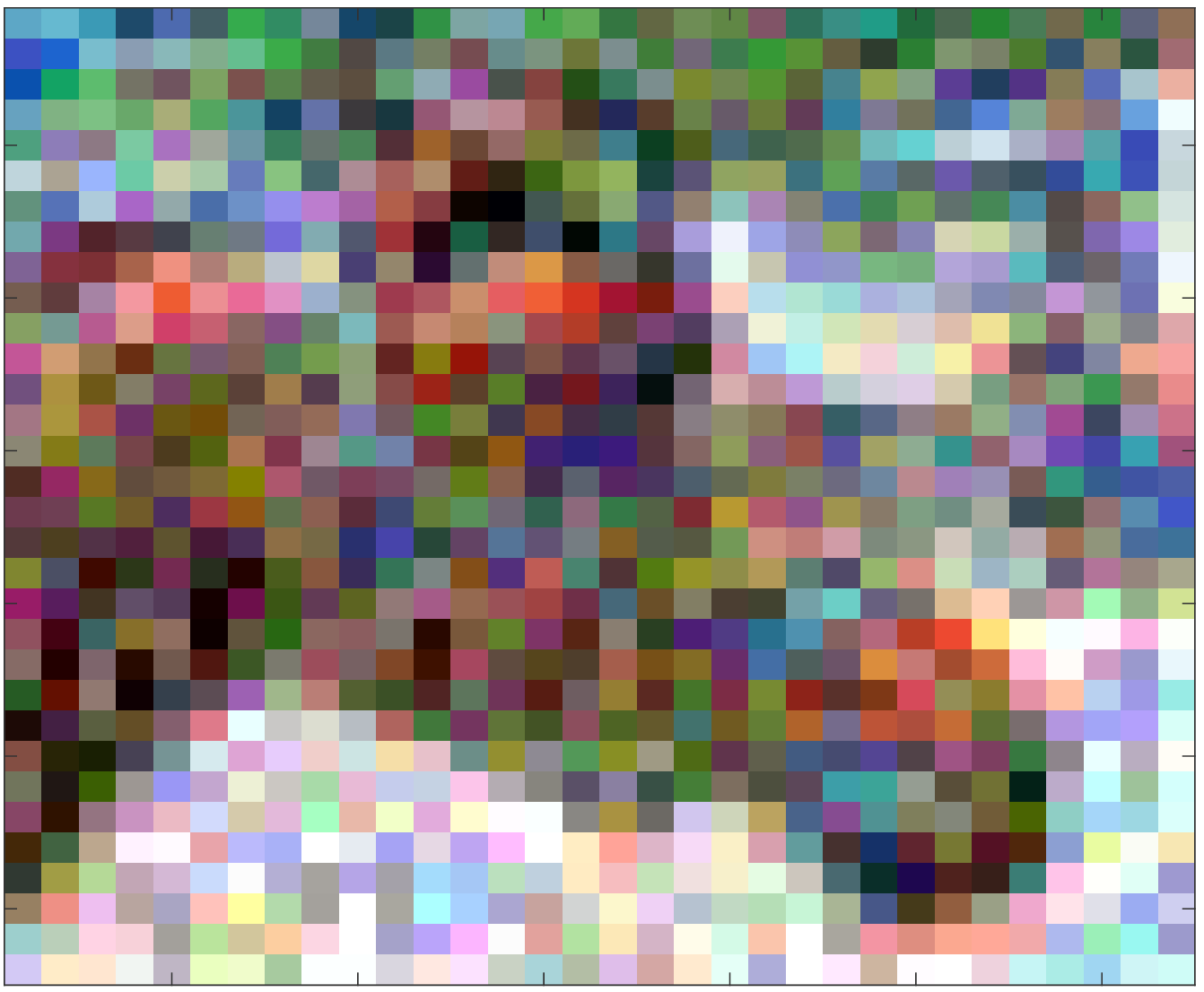}}
\hfill
\end{subfigure}
\begin{subfigure}[MIM]{\includegraphics[width=2.3cm]{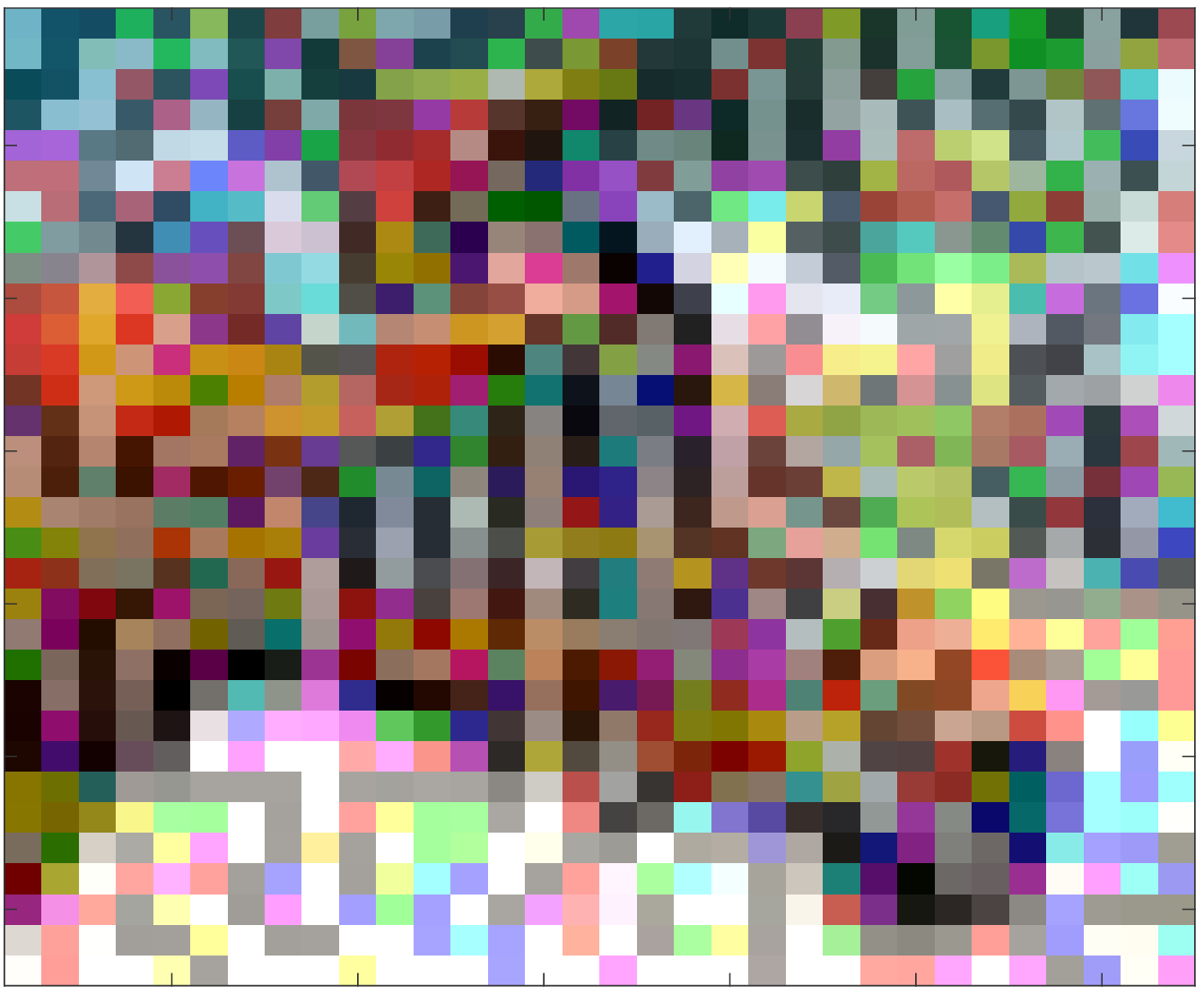}} 
\hfill
\end{subfigure}
\begin{subfigure}[JSMA]{\includegraphics[width=2.3cm]{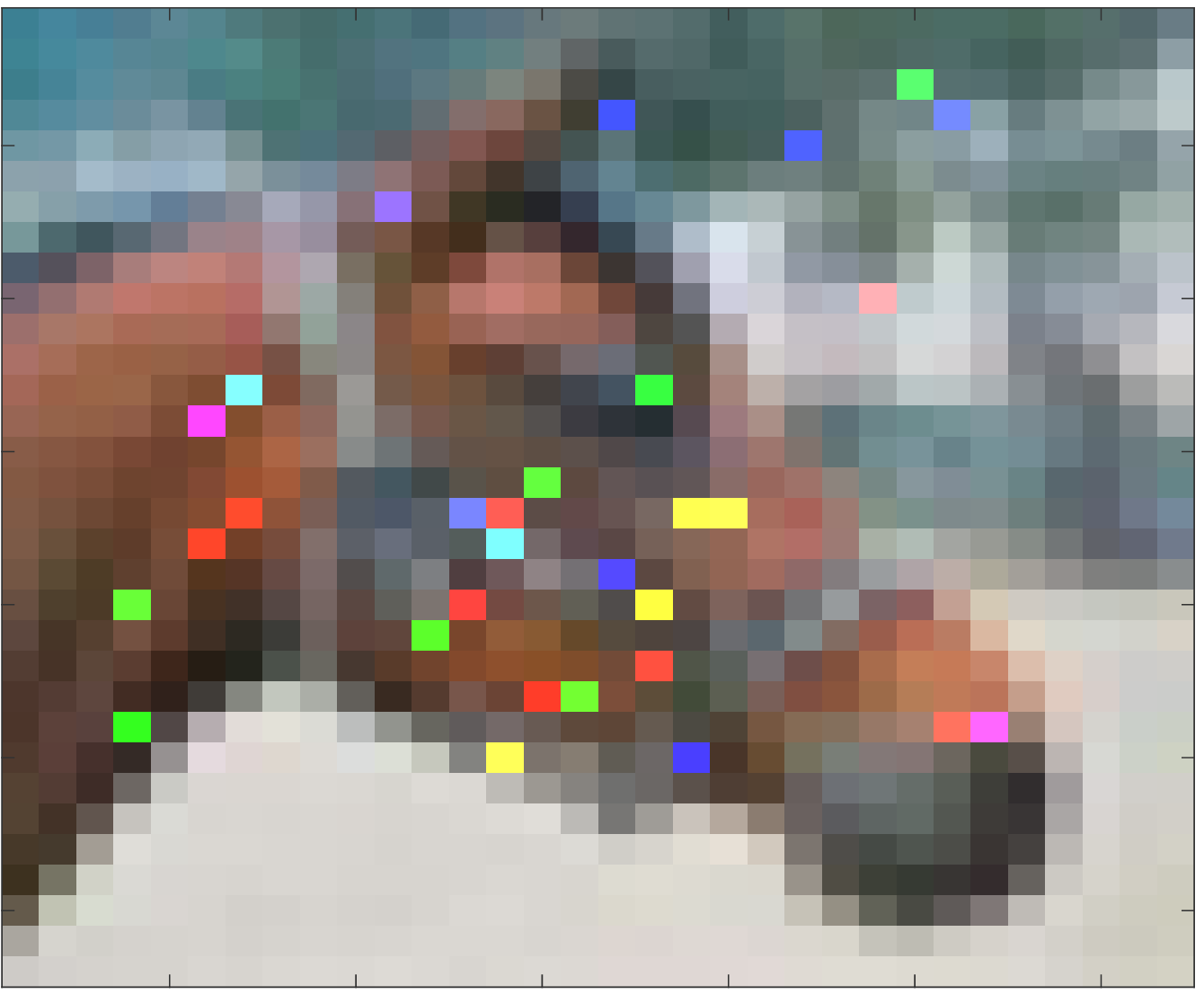}} 
\hfill
\end{subfigure}
\begin{subfigure}[\ours]{\includegraphics[width=2.3cm]{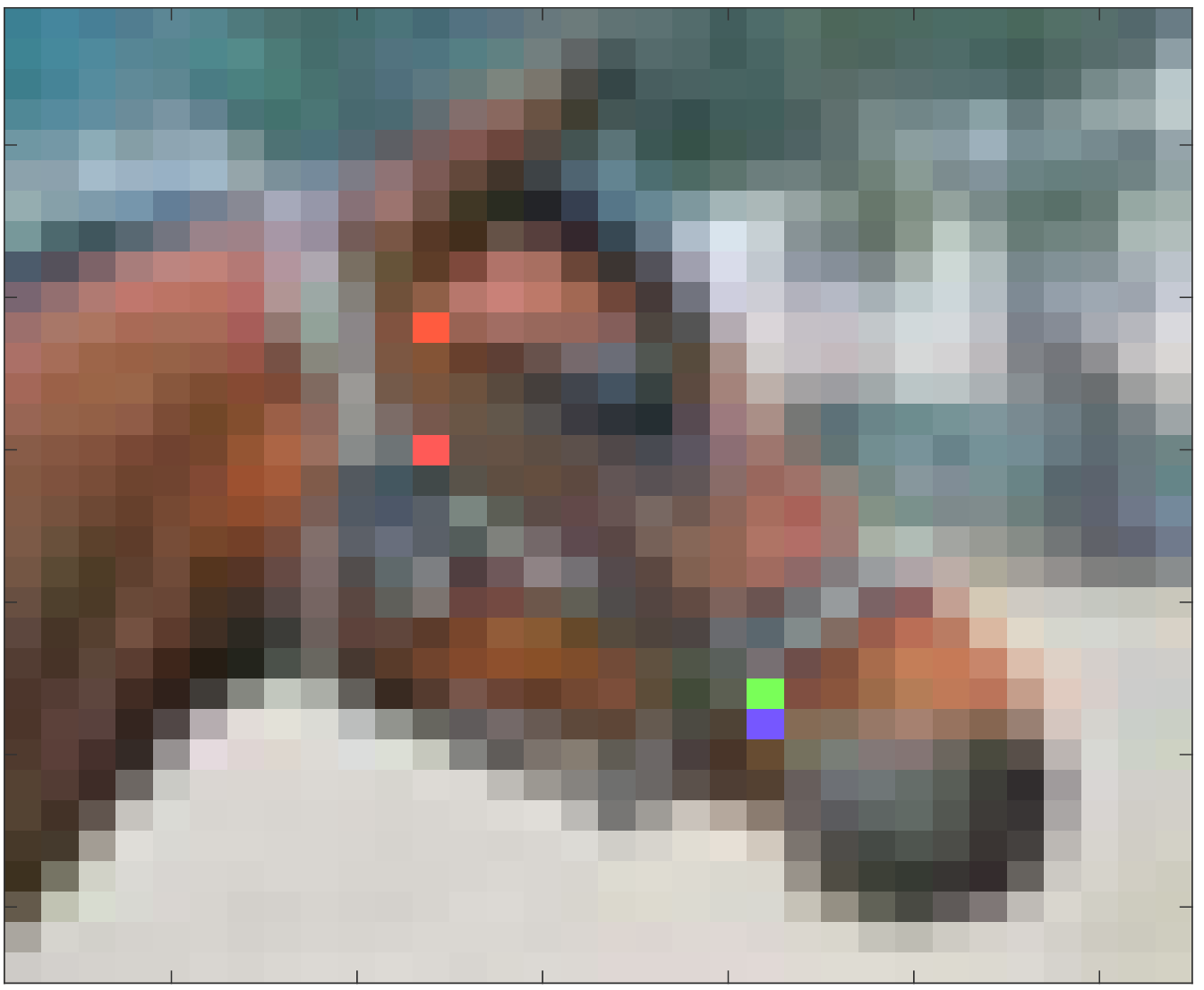}} 
\hfill
\end{subfigure}\vspace{-3mm}

\caption{AEs generated for CIFAR-10 using several adversarial attack methods. The adversary can force the DL model to misclassification only by perturbing small proportion of the image, while maintaining very close to the original image in terms of IQA metrics.}
\label{fig:Org_Adv_samples_CIFAR}

\begin{subfigure}[Original]{\includegraphics[width=2.3cm]{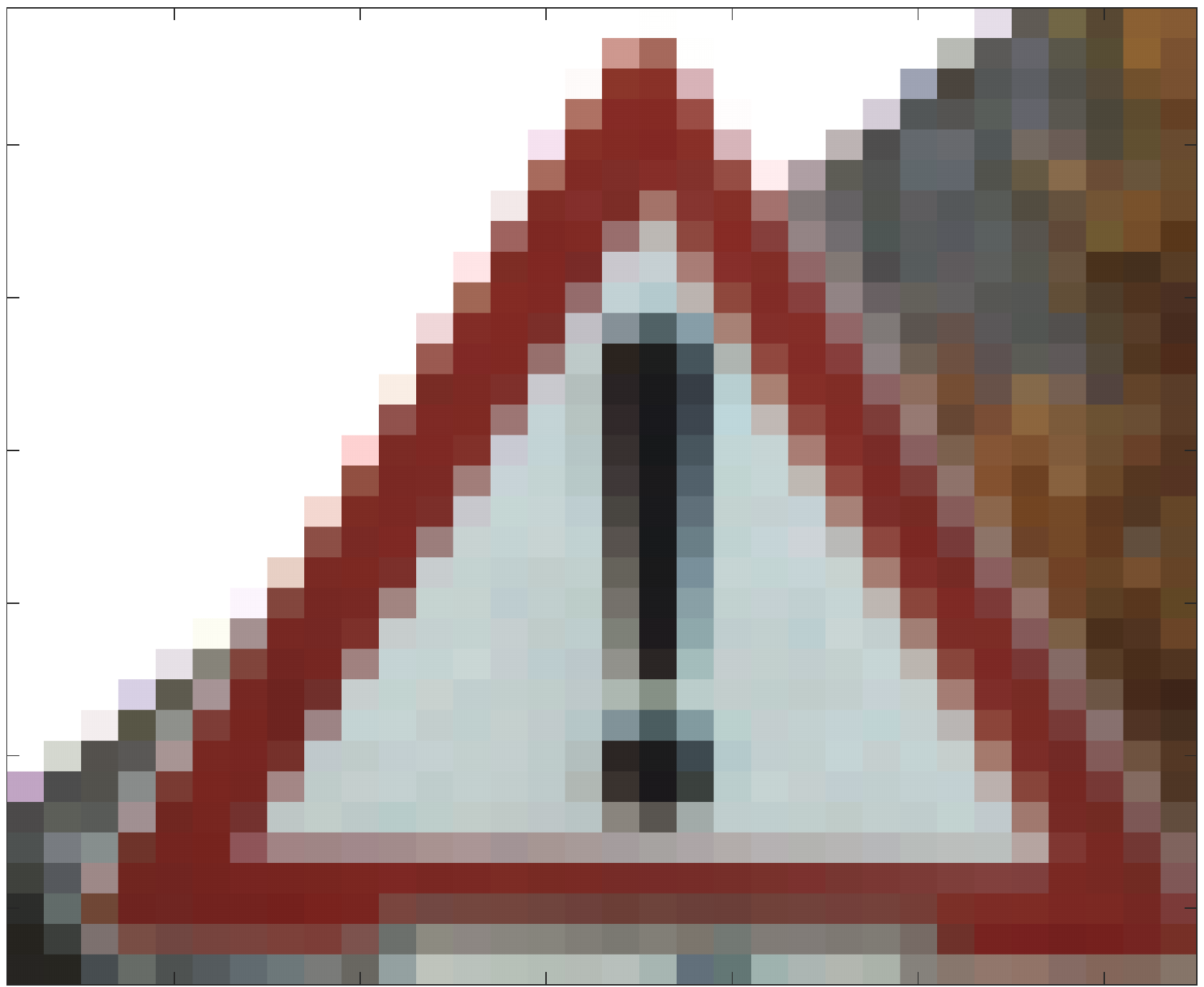}} 
\hfill
\end{subfigure}
\begin{subfigure}[FGSM]{\includegraphics[width=2.3cm]{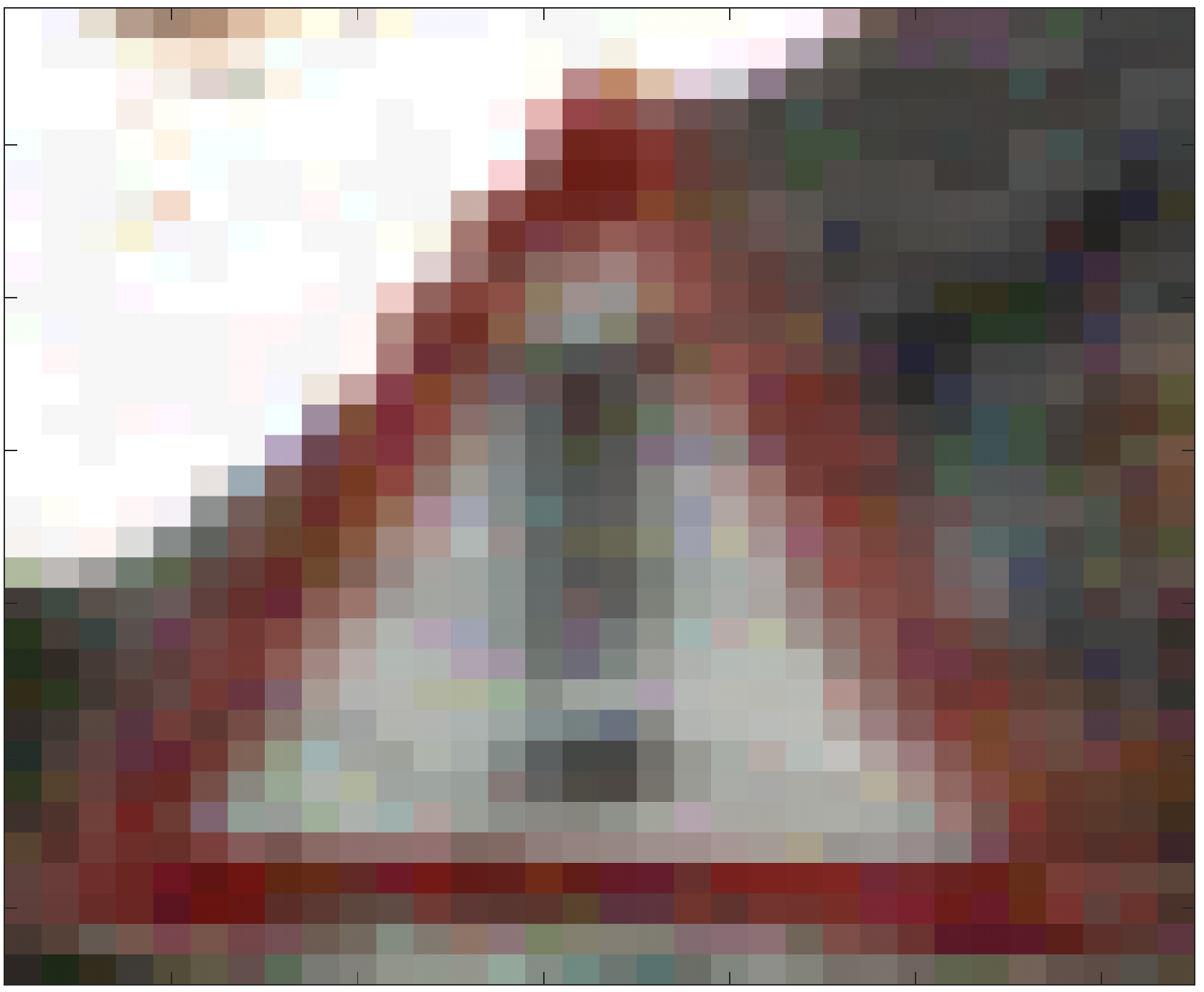}} 
\hfill
\end{subfigure}
\begin{subfigure}[C\&W]{\includegraphics[width=2.3cm]{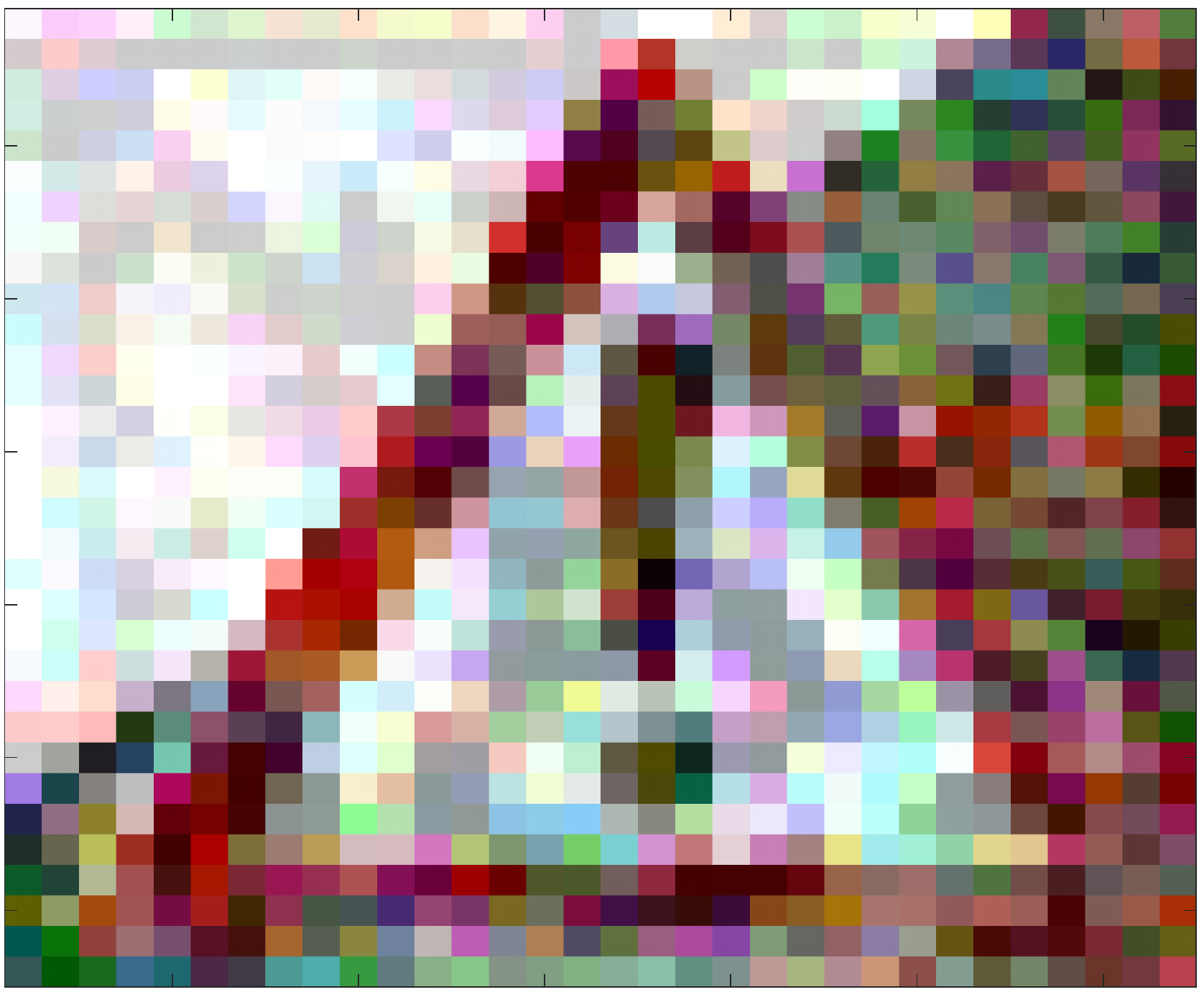}} 
\hfill
\end{subfigure}
\begin{subfigure}[PGD]{\includegraphics[width=2.3cm]{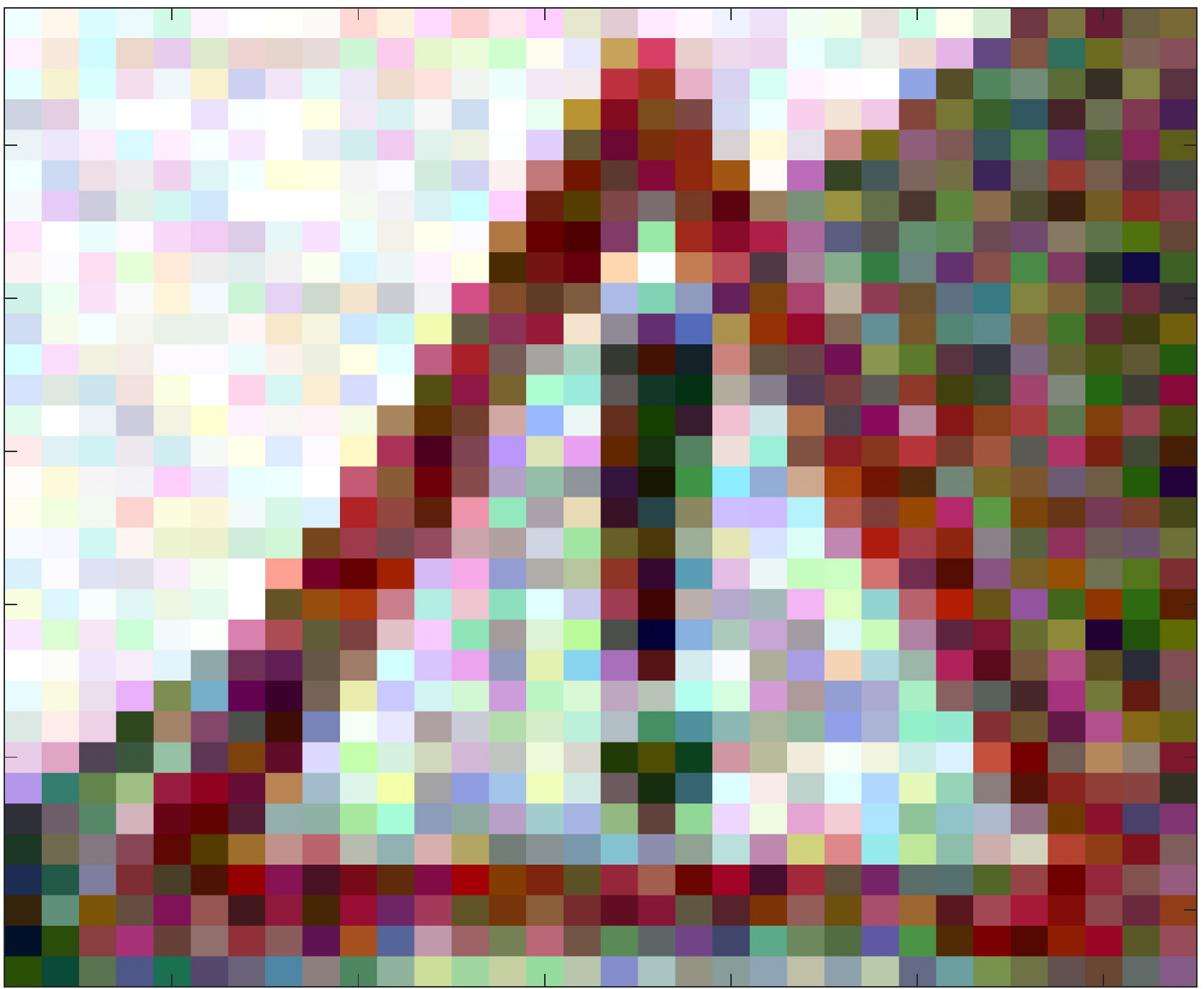}}
\hfill
\end{subfigure}
\begin{subfigure}[MIM]{\includegraphics[width=2.3cm]{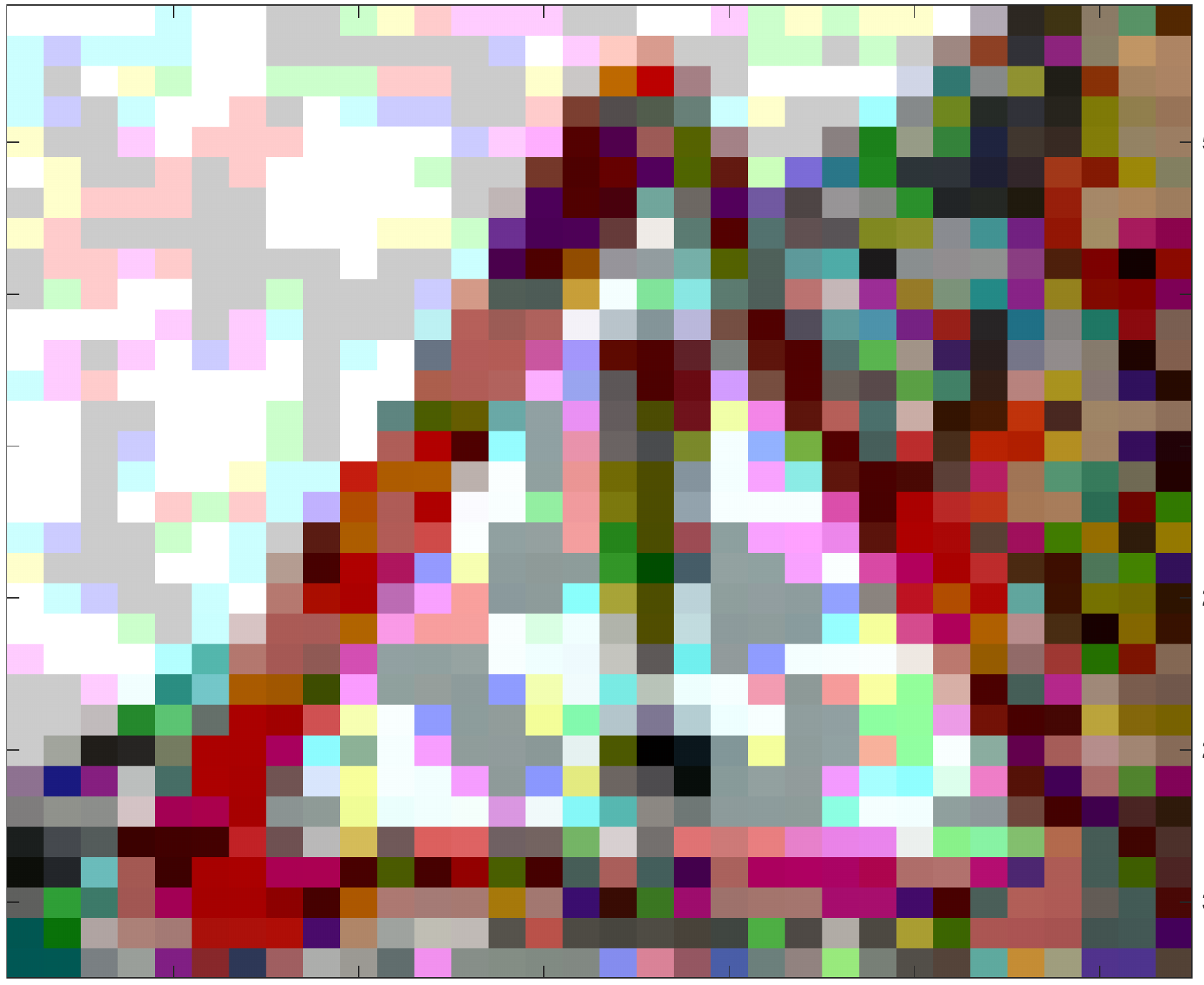}}
\hfill
\end{subfigure}
\begin{subfigure}[JSMA]{\includegraphics[width=2.3cm]{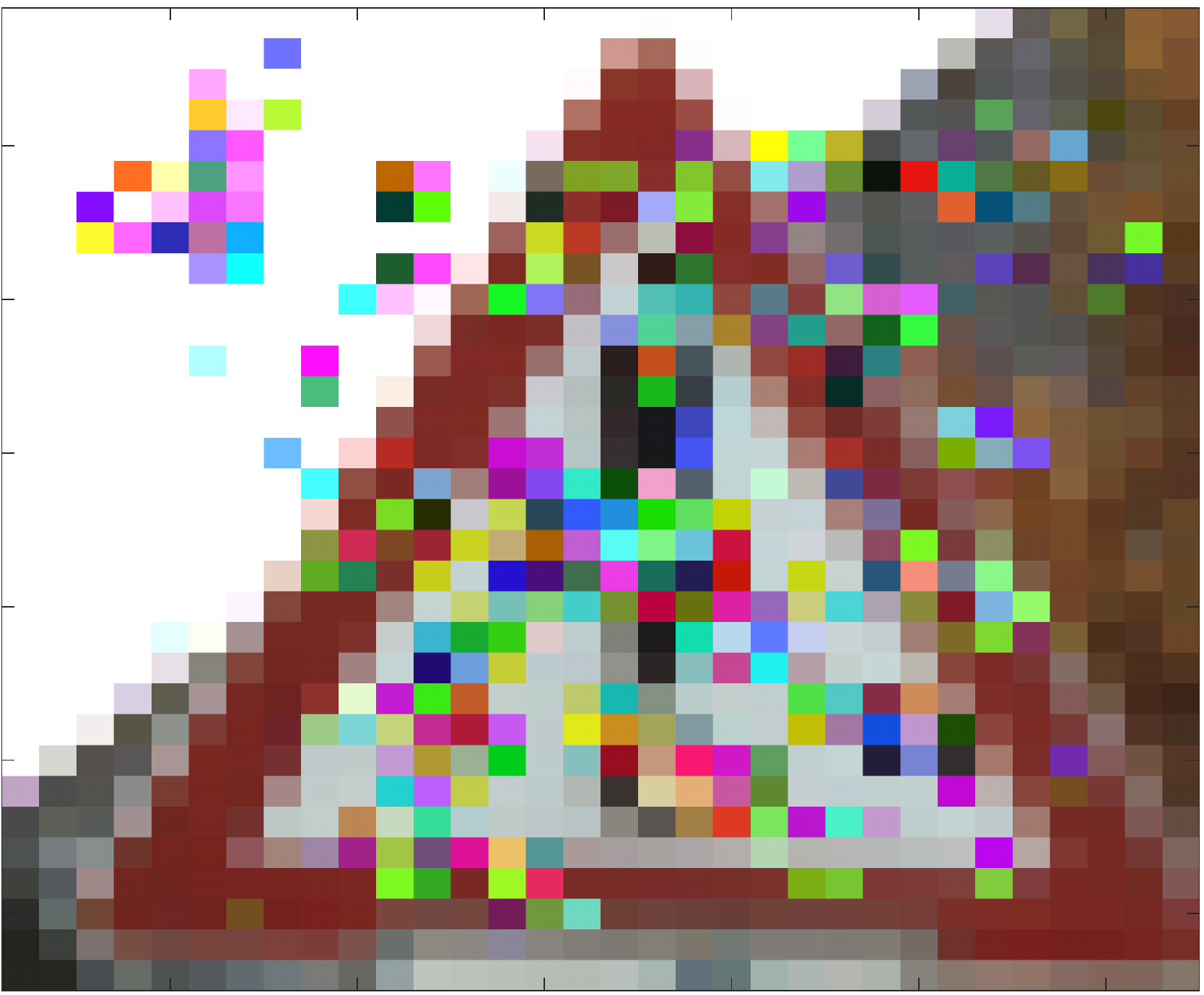}}
\hfill
\end{subfigure}
\begin{subfigure}[\ours]{\includegraphics[width=2.3cm]{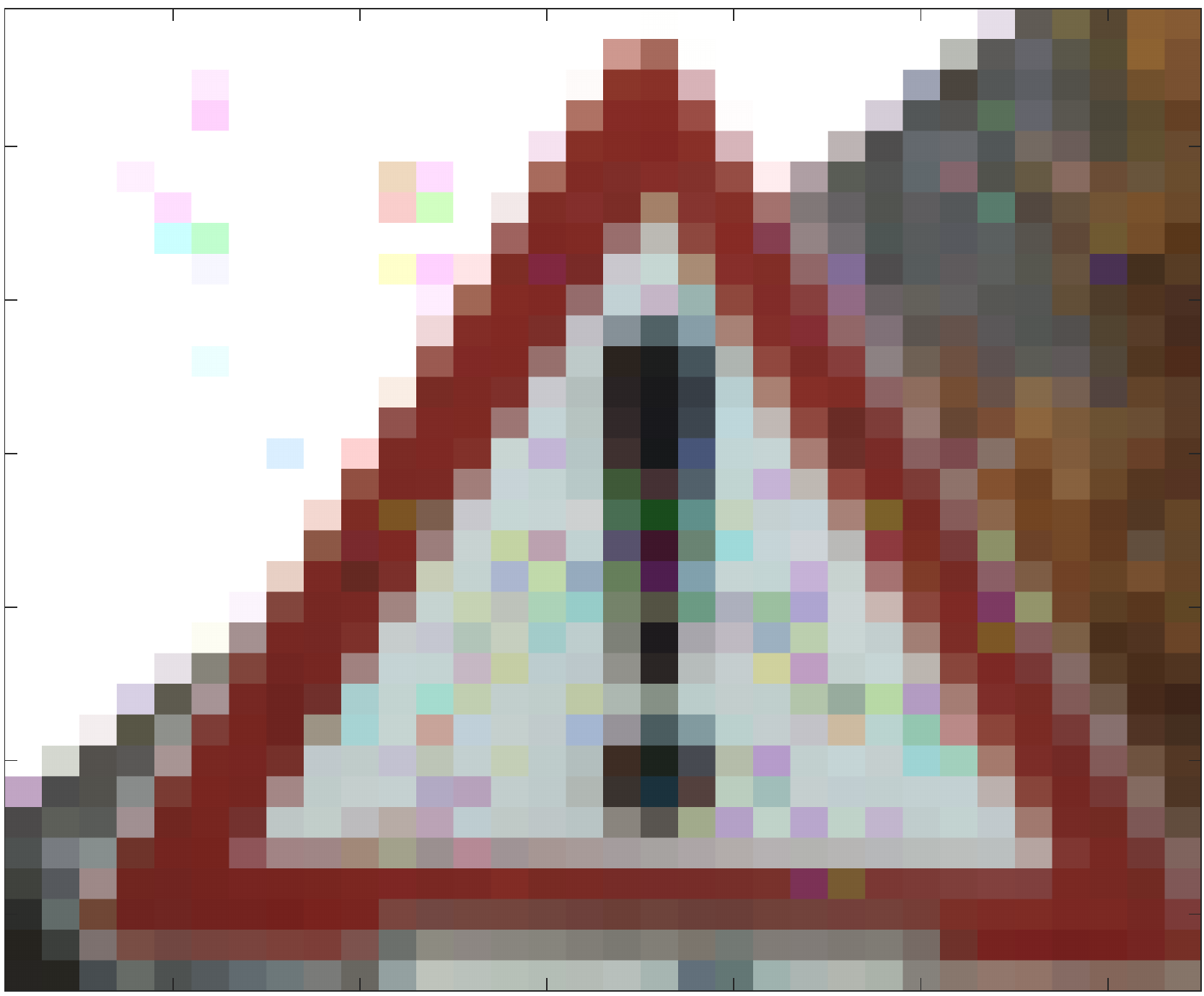}} 
\hfill
\end{subfigure}\vspace{-3mm}

\caption{Sample AEs generated for GTSRB benchmark, where the FAQ outperforms existing methods in terms of IQA metrics while perturbing only $8.72\%$ of pixels in a given image on average.}
\label{fig:Org_Adv_samples_GTSRB}

\begin{subfigure}[Original]{\includegraphics[width=2.3cm]{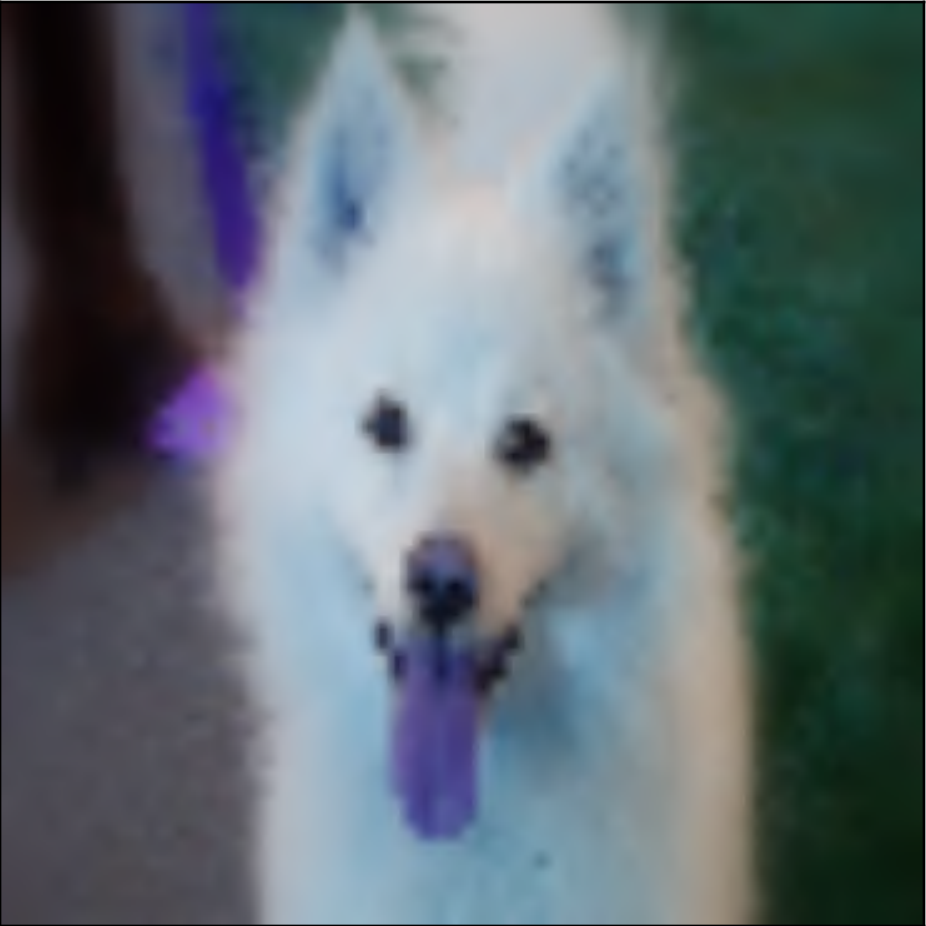}}
\hfill
\end{subfigure}
\begin{subfigure}[FGSM]{\includegraphics[width=2.3cm]{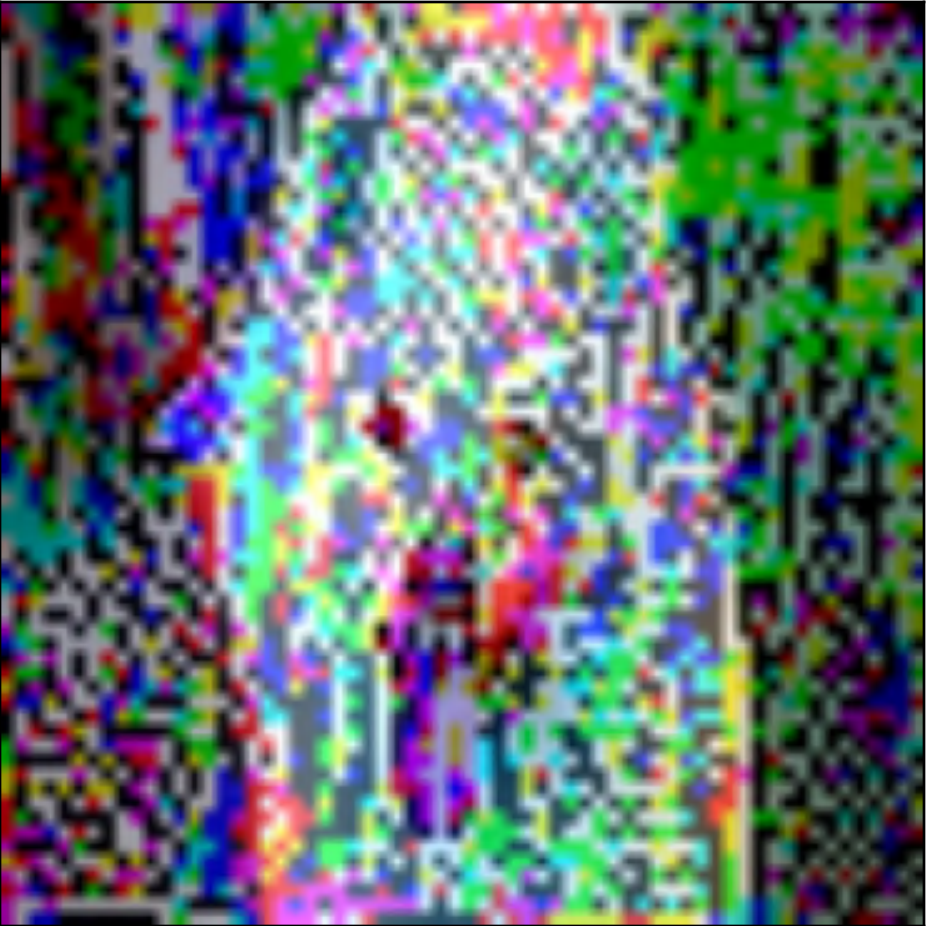}} 
\hfill
\end{subfigure}
\begin{subfigure}[C\&W]{\includegraphics[width=2.3cm]{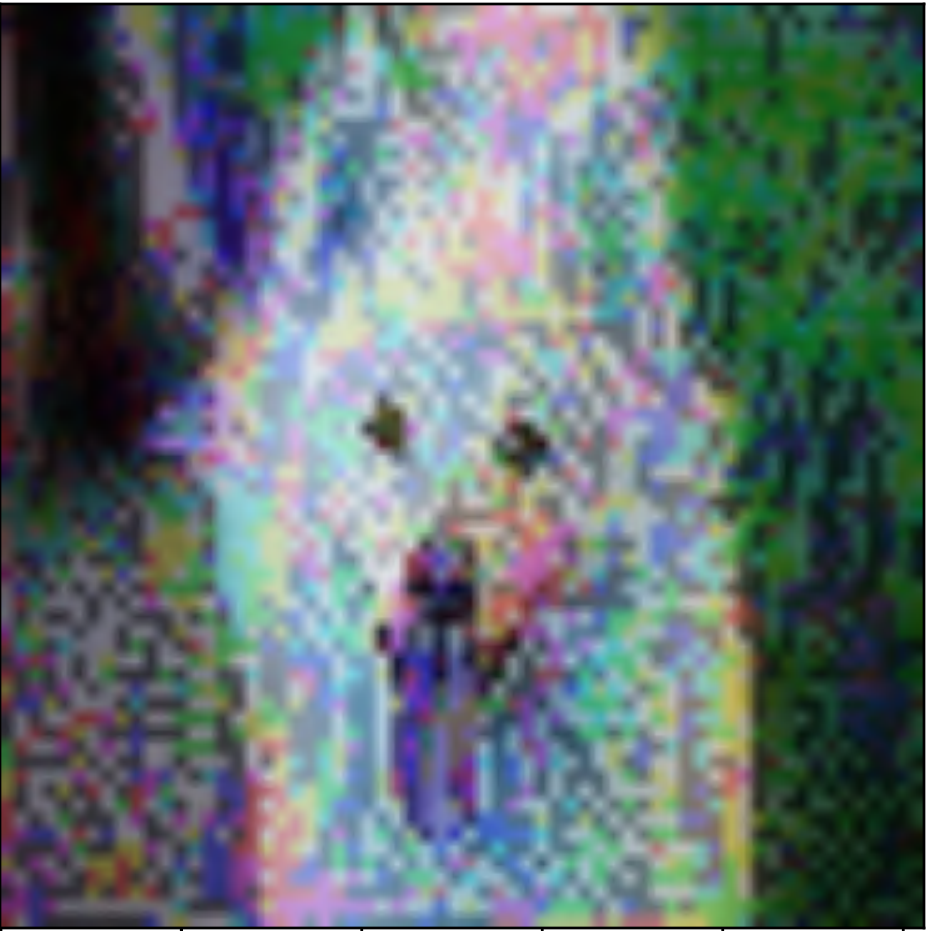}} 
\hfill
\end{subfigure}
\begin{subfigure}[PGD]{\includegraphics[width=2.3cm]{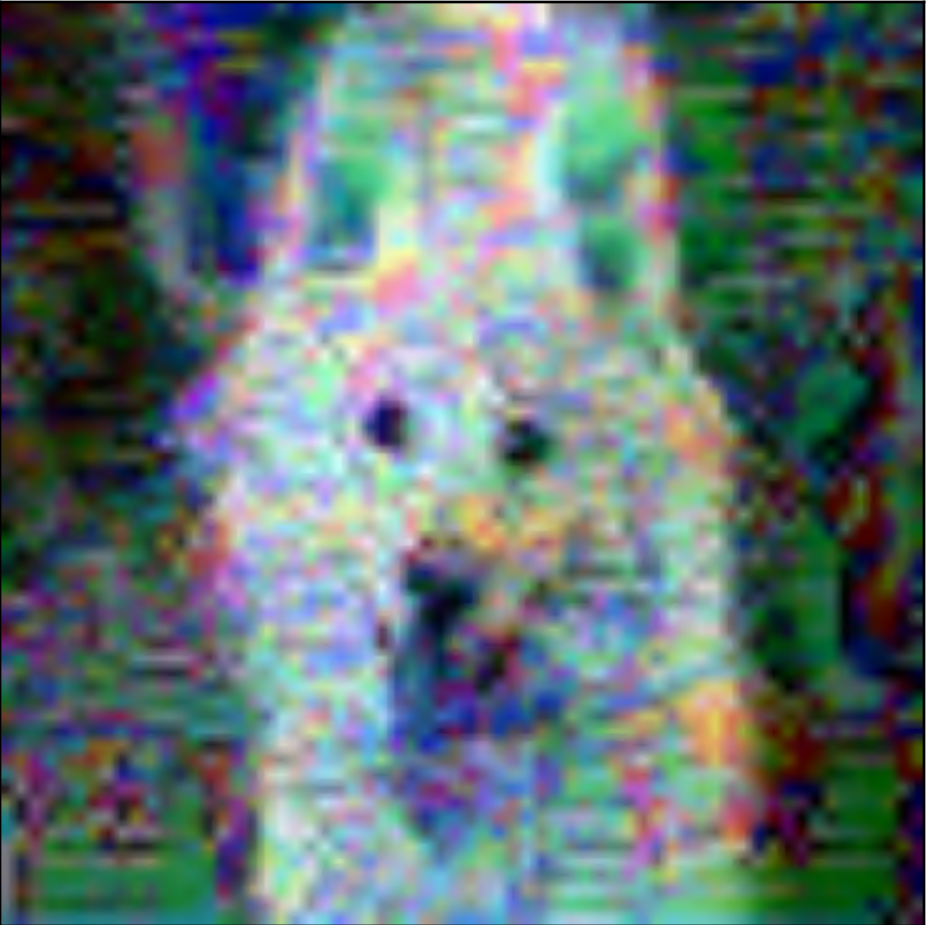}}
\hfill
\end{subfigure}
\begin{subfigure}[MIM]{\includegraphics[width=2.3cm]{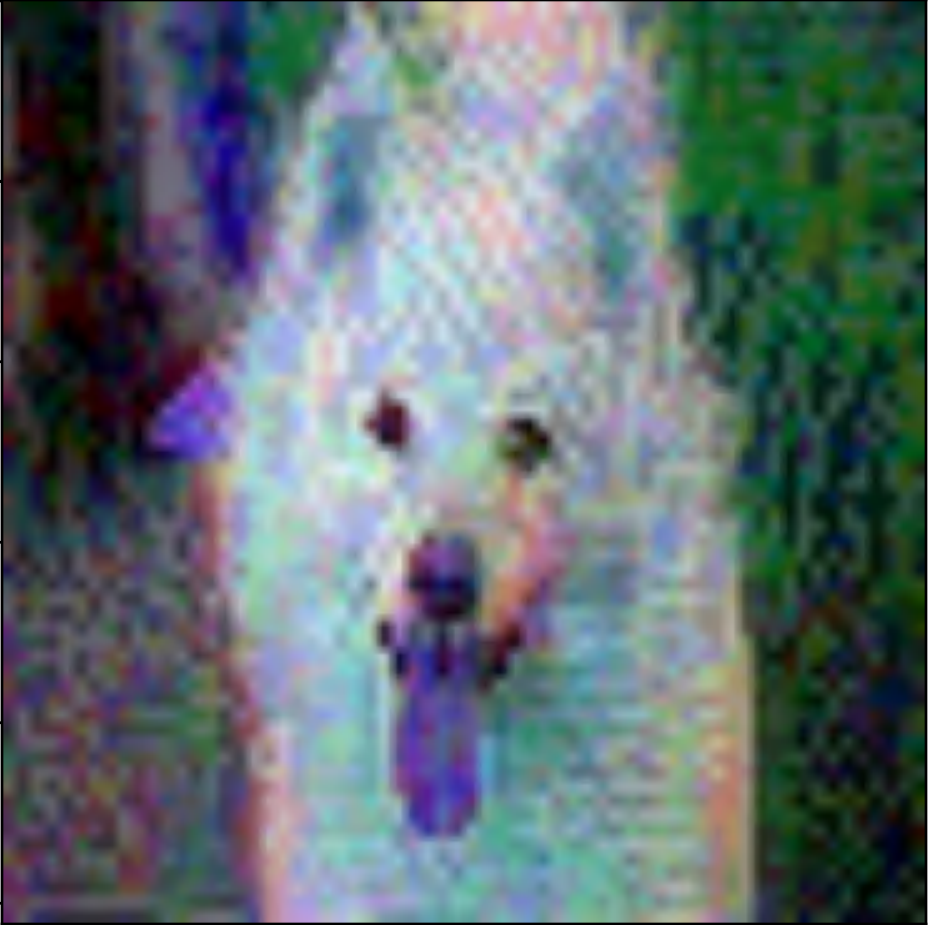}} 
\hfill
\end{subfigure}
\begin{subfigure}[JSMA]{\includegraphics[width=2.3cm]{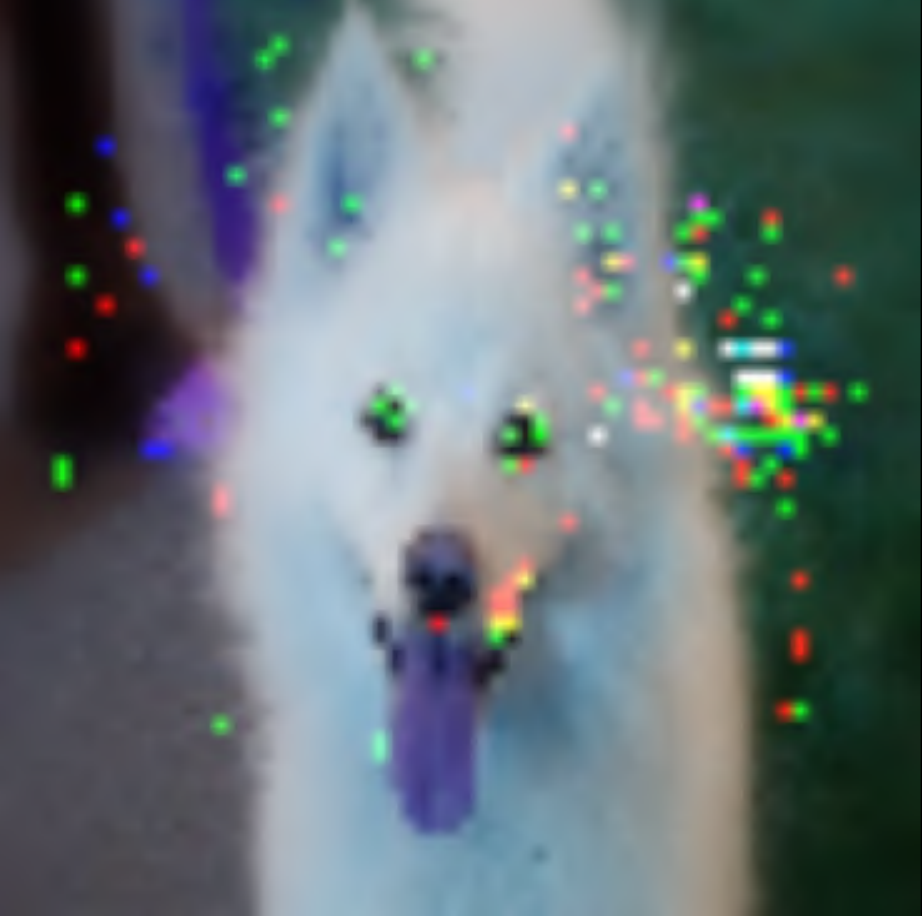}} 
\hfill
\end{subfigure}
\begin{subfigure}[\ours]{\includegraphics[width=2.3cm]{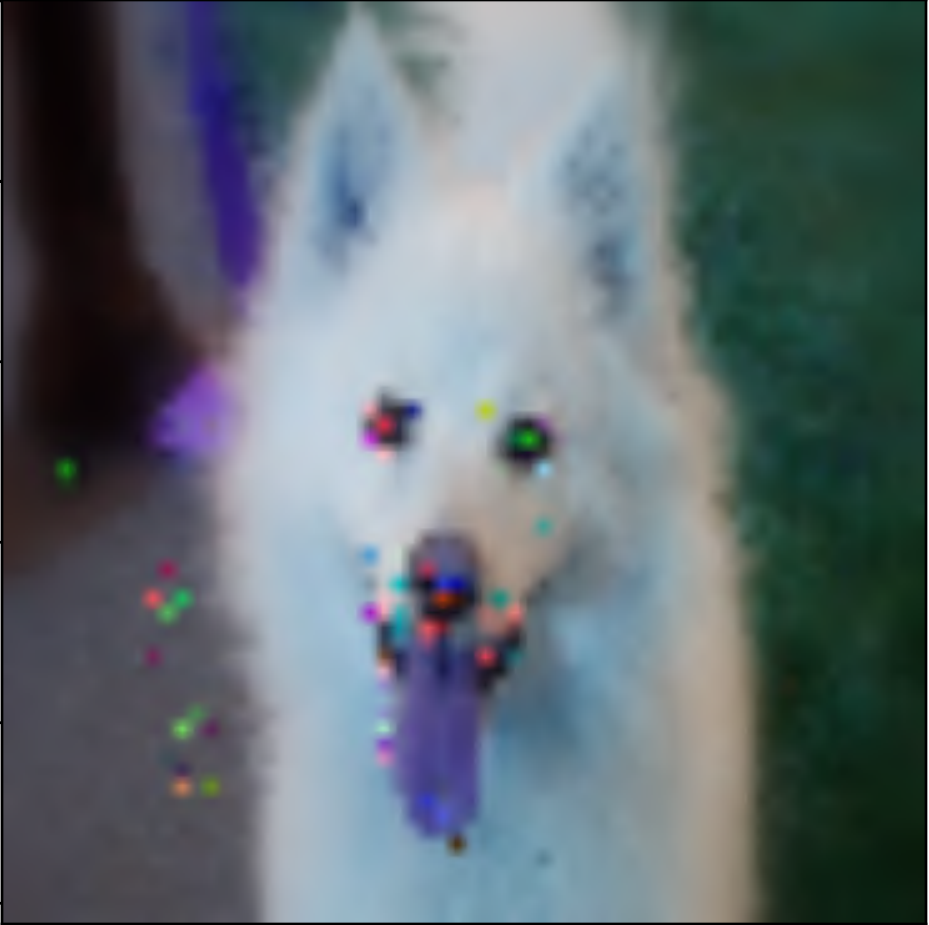}} 
\hfill
\end{subfigure}\vspace{-3mm}

\caption{Sample AEs generated for high resolution images from the Open Image Dataset V5~\cite{OpenImage} dataset using various adversarial attack methods. Our approach is able to generate AEs with better IQA metrics in comparison with other approaches.}
\label{fig:Org_Adv_samples_hires}

\end{minipage}

\end{figure*}

\cref{fig:Org_Adv_samples_Digits}--\cref{fig:Org_Adv_samples_hires}  show sample AEs generated for the five different benchmarks---MNIST hand-written digits, Fashion MNIST, CIFAR-10, GTSRB, and Open Image---using various attack methods.

\subsubsection{SOO Configuration} \label{SOO_Conff}

\begin{figure*}[t]
\begin{minipage}{.24\textwidth}
\centering
\includegraphics[width=0.99\textwidth, height=1.125in]{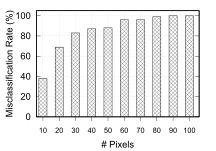}\vspace{-3mm}
\caption{No. of perturbed pixels effect on AE's misclassification rate using Fashion MNIST dataset.}
\label{fig:SuccessRate}\vspace{-4mm}
\end{minipage}\hspace{1mm}
\begin{minipage}{.24\textwidth}
\centering
\includegraphics[width=0.99\textwidth, height=1.125in]{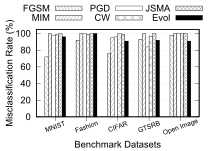}\vspace{-3mm}
\caption{Misclassification rate of various adversarial attack methods on different benchmarks.}
\label{fig:misclassification}\vspace{-4mm}
\end{minipage}\hspace{1mm}
\begin{minipage}{.24\textwidth}
\centering
\includegraphics[width=0.99\textwidth, height=1.125in]{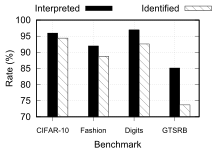}\vspace{-3mm}
\caption{Human perception. GTSRB's results are due to users' lack of  knowledge on road signs.}
\label{fig:HumanPerception}\vspace{-4mm}
\end{minipage}\hspace{1mm}
\begin{minipage}{.24\textwidth}
\centering
\includegraphics[width=0.99\textwidth, height=1.125in]{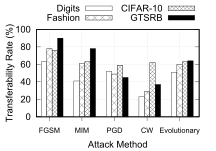}\vspace{-2mm}
\caption{Transferability using various adversarial attack methods and benchmarks.}
\label{fig:Transferability_Various}\vspace{-4mm}
\end{minipage}
\end{figure*}

In order to find the minimal value of pixels to be perturbed, we conducted several experiments using different values of $n$, and selected the one that leads to higher misclassification rate. For example, as it is shown in~\cref{fig:SuccessRate}, the distortion of 90 pixels on Fashion MNIST dataset can definitely lead to $100\%$ misclassification rate, without further need to perturb more pixels. Note that the selected pixels are not identical across images, as their saliency score pattern might be different depending on the nature of the image. For instance, on MNIST datasets the saliency score at the center of the image is higher, as the main object (digits, fashion items) are located at the center of the image. However, the saliency score of images from  CIFAR-10 and Open Image datasets are more distributed across the image.  
Several experiments are conducted on other benchmarks and we found that \ours is able to achieve reasonable misclassification rates by perturbing only 5.22\%, 7.21, 7.28\%, and 9.11\% of pixels, on average, in MNIST hand-written digits, CIFAR-10, GTSRB, and Open Image datasets, respectively. Note that this is promising, as an adversary does not need to distort a large number of pixels to achieve his/her goal. At the same time, this implies that the decision of DL model is mostly contingent upon those salient pixels, while other pixels have smaller impact. 
It should be noted that in the SOO configuration of \ours we do not consider any of the IQA metrics, and the only objective is achieving a higher misclassification rate. In other words, the generated AEs do not necessarily offer good quality metrics. 

In order to provide better insight and comparison between our SOO approach and other existing AE generation methods, where all of them have only one goal; that is achieving misclassification, the obtained results are presented categorically and based on benchmarks, as follows:  MNIST digits, Fashion MNIST, CIFAR-10, GTSRB, and Open Image all in~\cref{tab:res_single_GTSRB}. As it can be seen from these tables, the proposed SOO approach not only is able achieve a reasonable misclassification rate through distortion of only a small proportion of pixels but also produces AEs with better IQA metrics, in comparison to the existing adversarial attack methods. In the case of Fashion MNIST dataset, for example, we are able to achieve a misclassification rate of 100\% while perturbing only 36.9 pixels (4.71\% of all pixels), on average. Moreover, all other metrics, such as IQA (HOG metric=0.66, FFT metric=82.5, and Edge diff=1.49) and L2 distance between the AE and original image (2.89) is better than that of other adversarial attack methods.

\begin{table}[htb]
\centering
\caption{Results for various adversarial attack methods with the SOO configurations and four benchmark datasets. In those experiments, we use W1=1, and W2=W3=W4=0. The results clearly demonstrate that traditional methods offer poor IQA features.} \label{tab:res_single_GTSRB}\vspace{-3mm}
\scalebox{0.9}
{
\begin{tabular}{|c|c|c|c|c|c|c|c|}
\Xhline{2\arrayrulewidth}
\multicolumn{8}{|c|}{\bf MNIST hand-written digits dataset}\\
\Xhline{2\arrayrulewidth}
 \multirow{2}{*}{App.} & \multirow{2}{*}{Pixels} & \multirow{2}{*}{Pert.} & \multicolumn{3}{c|}{Edge}                & \multirow{2}{*}{FFT} & \multirow{2}{*}{HOG} \\
\cline{4-6}
                       &           &                        & I\_Org.                & I\_Adv. & Diff. &                      &                      \\ \hline
FGSM                  & 59.2       & 34.6                   & \multirow{6}{*}{77.34} & 77.10   & 6.80  & 175.84               & 1.54                 \\ \cline{1-1} \cline{2-3} \cline{5-8} 
MIM                  & 54.4       & 20.2                   &                        & 76.67   & 3.09  & 111.66               & 1.18                 \\ \cline{1-1} \cline{2-3} \cline{5-8} 
PGD                    & 77.2         & 23.3                  &                        & 77.27   & 2.76  & 121.51               & 1.24                 \\ \cline{1-1} \cline{2-3} \cline{5-8} 
C\&W                  & 82.1         & 6.3                   &                        & 76.91   & 2.88  & 107.11               & 1.19                 \\ \cline{1-1} \cline{2-3} \cline{5-8} 
JSMA                 & 4.1        & 4.50                   &                        & 79.33   & 3.83  & 110.39               & 1.02                 \\ \cline{1-1} \cline{2-3} \cline{5-8} 
Evol.                 & 5.45        & 3.76                   &                        & 78.91   & 2.13  & 102.23               & 0.72                 \\ 
                       \Xhline{2\arrayrulewidth}
\multicolumn{8}{|c|}{\bf Fashion MNIST dataset}\\
\Xhline{2\arrayrulewidth}
 \multirow{2}{*}{App.} & \multirow{2}{*}{Pixels} & \multirow{2}{*}{Pert.} & \multicolumn{3}{c|}{Edge}               & \multirow{2}{*}{FFT} & \multirow{2}{*}{HOG} \\
\cline{4-6}
                     &           &                        & I\_Org.               & I\_Adv. & Diff. &                      &                      \\ \hline
 FGSM                & 76.7       & 28.8                   & \multirow{6}{*}{65.4} & 71.4    & 7.88  & 192.93               & 1.55                 \\ \cline{1-1} \cline{2-3} \cline{5-8}  
 MIM                 & 72.8       & 17.1                  &                       & 68.68   & 4.71  & 127.68               & 1.39                 \\ \cline{1-1} \cline{2-3} \cline{5-8}  
 PGD                 & 81.9       & 13.3                  &                       & 67.53   & 4.48  & 102.51               & 1.28                 \\ \cline{1-1} \cline{2-3} \cline{5-8}  
 C\&W                & 88.9       & 11.23                  &                       & 67.37   & 4.26  & 129.11               & 1.26                 \\ \cline{1-1} \cline{2-3} \cline{5-8}  
 JSMA                & 4.3        & 4.95                   &                       & 67.12   & 2.47  & 107.35               & 0.97                 \\ \cline{1-1} \cline{2-3} \cline{5-8}  
 Evol.               & 4.7        & 2.89                   &                       & 66.71   & 1.49  & 82.5                 & 0.66                 \\ 
\Xhline{2\arrayrulewidth}
\multicolumn{8}{|c|}{\bf CIFAR-10 dataset}\\
\Xhline{2\arrayrulewidth}
 \multirow{2}{*}{App.} & \multirow{2}{*}{Pixels} & \multirow{2}{*}{Pert.} & \multicolumn{3}{c|}{Edge}               & \multirow{2}{*}{FFT} & \multirow{2}{*}{HOG} \\
\cline{4-6}
                        &          &                        & I\_Org.               & I\_Adv. & Diff. &                      &                      \\ \hline
 FGSM                & 98.3       & 11.4                   & \multirow{6}{*}{89.8} & 100.1   & 15.06 & 230.35               & 2.58                 \\ \cline{1-1} \cline{2-3} \cline{5-8}  
 MIM                 & 80.6       & 3.1                   &                       & 93.57   & 5.49  & 81.5                 & 1.84                 \\ \cline{1-1} \cline{2-3} \cline{5-8}  
 PGD                 & 96.4       & 4.6                   &                       & 94.71   & 7.45  & 84.88                & 1.95                 \\ \cline{1-1} \cline{2-3} \cline{5-8}  
 C\&W                & 97.9       & 8.9                   &                       & 96.02   & 9.48  & 127.9                & 2.04                 \\ \cline{1-1} \cline{2-3} \cline{5-8}  
 JSMA                & 6.6        & 2.9                   &                       & 94.43   & 7.65  & 79.79                & 1.78                 \\ \cline{1-1} \cline{2-3} \cline{5-8}  
 Evol.               & 7.21        & 3.1                   &                       & 92.35   & 5.06  & 56.24                & 1.53                 \\ 
                       \Xhline{2\arrayrulewidth}
\multicolumn{8}{|c|}{\bf GTSRB dataset}\\
\Xhline{2\arrayrulewidth}
 \multirow{2}{*}{App.} & \multirow{2}{*}{Pixels} & \multirow{2}{*}{Pert.} & \multicolumn{3}{c|}{Edge}                & \multirow{2}{*}{FFT} & \multirow{2}{*}{HOG} \\
\cline{4-6}
                     &           &                        & I\_Org.                & I\_Adv. & Diff. &                      &                      \\ \hline
 FGSM                & 96.8       & 19.5                   & \multirow{6}{*}{95.23} & 141.01  & 48.88 & 230.36               & 2.87                 \\ \cline{1-1} \cline{2-3} \cline{5-8}  
 MIM                 & 88.8        & 11.4                   &                        & 119.19  & 28.35 & 150.16               & 2.66                 \\ \cline{1-1} \cline{2-3} \cline{5-8}  
 PGD                 & 95.4       & 5.8                   &                        & 105.08  & 14.89 & 85.37                & 2.32                 \\ \cline{1-1} \cline{2-3} \cline{5-8}  
 C\&W                & 96.8       & 6.77                   &                        & 104.13  & 13.08 & 82.41                & 2.04                 \\ \cline{1-1} \cline{2-3} \cline{5-8}  
 JSMA                & 7.6        & 3.67                   &                        & 102.35  & 9.20  & 100.24               & 2.39                 \\ \cline{1-1} \cline{2-3} \cline{5-8}  
 Evol.               & 7.28        & 3.34                   &                        & 97.56   & 3.45  & 42.36                & 1.48                 \\ 
                       \Xhline{2\arrayrulewidth}
                       
\multicolumn{8}{|c|}{\bf Open Image dataset}\\
\Xhline{2\arrayrulewidth}
 \multirow{2}{*}{App.} & \multirow{2}{*}{Pixels} & \multirow{2}{*}{Pert.} & \multicolumn{3}{c|}{Edge}                & \multirow{2}{*}{FFT} & \multirow{2}{*}{HOG} \\
\cline{4-6}
                     &           &                        & I\_Org.                & I\_Adv. & Diff. &                      &                      \\ \hline
 FGSM                & 98.8       & 20.69                   & \multirow{6}{*}{347.9} & 403.01  & 60.18 & 801.69               & 5.18                 \\ \cline{1-1} \cline{2-3} \cline{5-8}  
 MIM                 & 97.4        & 4.18                   &                        & 350.33  & 13.94 & 246.43               & 3.43                 \\ \cline{1-1} \cline{2-3} \cline{5-8}  
 PGD                 & 98.2       & 9.05                   &                        & 355.90  & 23.53 & 364.46                & 4.20                 \\ \cline{1-1} \cline{2-3} \cline{5-8}  
 C\&W                & 98.3       & 8.01                   &                        & 354.01  & 21.05 & 320.07                & 3.72                 \\ \cline{1-1} \cline{2-3} \cline{5-8}  
 JSMA                & 2.59        & 12.10                   &                        & 354.26  & 13.31  & 157.94               & 1.94                \\ \cline{1-1} \cline{2-3} \cline{5-8}  
 Evol.               & 2.75        & 3.76                   &                        & 350.10   & 3.24  & 112.36                & 1.86                 \\ \Xhline{2\arrayrulewidth}

\end{tabular}}
\end{table}

\subsubsection{MOO Configuration}
Unlike SOO, where the goal was to achieve misclassification, the main goal of MOO is to improve all three aforementioned IQA metrics of the AEs at the same time, while perturbing only small proportion of pixels. 
The results show that our MOO approach is able to improve all IQA  metrics simultaneously, while maintaining the misclassification rate relatively high.
It should be noted that in this configuration we consider a weighted linear combination of four different objectives as a final objective function that the optimizer, the PSO algorithm, attempts to optimize. The results for different datasets are shown in~\cref{tab:res_MOO}, where it can be observed that while the misclassification rate remains high, all of the IQA metrics are improved in comparison with traditional adversarial attack methods as well as SOO approach (see~\cref{tab:res_single_GTSRB} for all benchmarks). This means that the MOO configuration is able to successfully fulfill all of the aforementioned objectives, with smaller L2 distance, and improving several IQA metrics. It should be noted that improving one quality  metric in MOO might affect other metrics, as the final objective function is a combination of multiple unique objectives, thus the final result is a compromise among all objectives. 

Our results show that \ours is able to offer a significant improvement in the IQA metrics of the AEs as well as L2 distance, regardless of the images' resolution, and with only a small misclassification rate degradation with high resolution images, while perturbing limited number of pixels. The misclassification rate degradation is reasonable as we have selected only a small number of pixels with high saliency score. However, per~\cref{SOO_Conff} and \cref{fig:SalielncyScore}, the saliency score pattern of the images differs based on their content. For instance, for the Open Image dataset the saliency score pattern is distributed and its value for pixels remain close to each other, which means that more pixels have a significant effect on the decision of the model. Thus, it is required to perturb larger number of pixels to achieve misclassification.

\begin{table}[t]
\centering
\caption{Results when the MOO configuration on various benchmarks with W1=0.4 and W2=W3=W4=0.2. We observe that all IQA metrics are improved, compared to the SOO configuration and other adversarial attack approaches.}\label{tab:res_MOO}\vspace{-3mm}
\scalebox{0.9}
{
\begin{tabular}{|c|c|c|c|c|c|c|c|}
\Xhline{2\arrayrulewidth}
 \multirow{2}{*}{App.} & \multirow{2}{*}{Pixels} & \multirow{2}{*}{Pert.} & \multicolumn{3}{c|}{Edge} & \multirow{2}{*}{FFT} & \multirow{2}{*}{HOG} \\
\cline{4-6}
                                              & \%          &                        & I\_Org. & I\_Adv. & Diff. &                      &                      \\ \Xhline{2\arrayrulewidth}
                        Digits                & 5.4         & 3.5                   & 77.3    & 78.7    & 1.4   & 99.8                 & 0.7                  \\ \cline{1-8}  
                        Fashion               & 4.7         & 2.9                   & 65.4    & 66.4    & 1.4   & 82.1                 & 0.6                  \\ \cline{1-8}  
                        CIFAR-10              & 7.2         & 3.2                   & 89.8    & 91.8    & 3.3   & 36.2                 & 1.1                  \\ \cline{1-8} 
                        GTSRB                 & 8.0         & 3.7                    & 95.2    & 96.1    & 1.3   & 42.3                 & 1.4                  \\ \cline{1-8} 
                        Open Image            & 6.34         & 3.90                    & 347.9    & 349.1    & 1.6   & 107.3                 & 1.5                  \\ 
                  \Xhline{2\arrayrulewidth}
\end{tabular}}\vspace{-3mm}
\end{table}


\subsection{Human Perception Study}\label{sec:human}

Adversarial examples need to be both misclassified by the deep learning model and classified correctly by humans. To test the latter requirement, we conducted an IRB-approved human perception study on the generated AEs employing 70 participants. To do so, we presented 10 samples from each of our datasets and asked if they can interpret the image, if so, we further asked them to label the image by selecting a potential label from provided bag of labels.

\BfPara{Participants demographic information} 57.14\% and 40\% of the participants identified as male and female, respectively, while the rest wished not to identify as either. The participants' age range was between 15 and 35, with 80\% of them between 20 and 30 years old. While a majority of the participants had a high level of education, only a small number of them were familiar with the concept of adversarial machine learning (23.80\% in total). Due to the lack of space, detailed demographic information, such as gender, age, and education level are delegated to \autoref{app:1},~\cref{fig:demographic}. 

\BfPara{Results} To understand the effect of AEs on human perception we selected samples from four benchmarks, 10 AEs from each, covering a wide range of subjects, \eg digits, fashion, road signs, \etc As it is demonstrated in~\cref{fig:HumanPerception}, participants were able to interpret samples at a rate of 95.9\%, 91.9\%, 96.0\%, and 85.1\% for CIFAR-10, Fashion MNIST, MNIST-Digits, and GTSRB datasets, respectively. Moreover, the participants were able to identify the correct label of the AEs with a success rate of 94.4\%, 88.8\%, 92.5\%, and 73.6\% for CIFAR-10, Fashion MNIST, MNIST-Digits, and GTSRB datasets, respectively. We note that the lower rate of the GTSRB dataset is due to the participants' lack of technical knowledge of traffic road signs. Moreover, we found that 16 individuals, out of 70, did not finish the GTSRB section of the experiment, whose feedback was  ``{\em this section is confusing and I don't have technical knowledge about those signs}'' and ``{\em the number of labels is very large (43)}''. Nevertheless, these results demonstrate that the generated AEs maintain human recognizablity, while being misclassified by the DL model.

\subsection{AE Transferability}\label{sec:transfer}

In order to investigate the transferability of the generated AEs we conducted several experiments on several benchmarks covering a wide range of domains, including digits, fashion, road signs, \etc. The findings of our experiments were in line with that of the theoretical work concerning the transferability of features and crafted AEs across different structures of deep networks~\cite{szegedy2014intriguing, papernot2017practical, yosinski2014transferable}. In particular, we found that \ours offers better transferability properties in comparison with the PGD and C\&W methods when using all of the benchmarks, as shown in~\cref{fig:Transferability_Various}. It should be noted that although the transferability rate of the FGSM and MIM approaches are higher than \ours{'}s, their IQA metrics are far below that of the \ours{'}s as shown in \autoref{fig:Transferability_Various}. 

\begin{figure*}[ht]
\centering
        \subfigure[MNIST-Digits\label{fig:mnist_transfer}] {\includegraphics[width=0.245\textwidth]{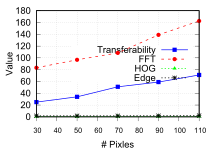}}
		\subfigure[Fashion MNIST\label{fig:fashion_transfer}] {\includegraphics[width=0.245\textwidth]{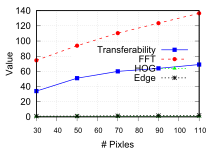}}
		\subfigure[CIFAR-10 \label{fig:cifar_transfer}] {\includegraphics[width=0.245\textwidth]{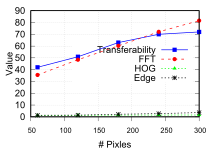}}
		\subfigure[GTSRB \label{fig:gtsrb_transfer}] {\includegraphics[width=0.245\textwidth]{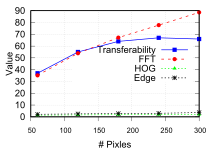}}
\vspace{-2mm}
\caption{The impact of the number of perturbed pixels on various characteristics of the AEs transferability and IQA metrics. While there is a positive correlation between transferability and number of perturbed pixels, the correlation is negative between IQA metrics.}\label{fig:pixel_trans}\vspace{-3mm}
\end{figure*}

We also investigated the impact of the number of perturbed pixels on various characteristics of the AEs, including transferability and IQA metrics. Our analyses demonstrate that there is a positive correlation between the number of perturbed pixels and transferability. For example, for CIFAR-10 dataset the transferability rate increases from 41\% to 72\% while perturbing 60 and 300 pixels, respectively~\cref{fig:cifar_transfer}, and the same pattern is shown with other benchmarks as well. Moreover, we observed that the IQA metrics of the AEs have a negative correlation with all of the IQA metrics, which is undesired. Note that this negative correlation is also anticipated, as the more distortion by modifying more pixels the less quality we get. In other words, transferability of AEs can be boosted at the cost of AEs with smaller IQA metrics~\cref{fig:pixel_trans}.

\subsection{Defense: Detectability of \ours}\label{sec:detection}
A final concern is how \ours affects the detectability of AEs. We note the number of modified pixels, distance (between an original and AE), and human perception are possible detection features. Moreover, as shown in this work, we note that the number of modified pixels in our work is smaller than in the literature, leading to a smaller distance between an original image and the corresponding AE than state-of-the-art. As a result, we conclude that our approach would bypass even a smaller threshold of detection when using those features, as compared to the literature. 

A unique feature of \ours is explicitly optimizing for IQA metrics. While the prior work tries to obtain good IQA scores, that is only implicitly, by calculating the IQA of the AEs without any control over the quality in the generation process. As a result, and as demonstrated in this work, \ours would bypass a detection method that utilizes IQA metrics even with a smaller threshold, where other competing approaches would fail. However, for both categories of features, a small threshold should be utilized to detect our AEs.

\section{Related Work}\label{sec:related}
Security assessment of machine learning networks~\cite{barreno2010security} is an evolving research area in both the security and machine learning communities with a range of attack categories, adversarial capabilities, and  defense considerations~\cite{huang2011adversarial,barreno2006can}. Biggio \etal studied the security of pattern classifiers at the design phase and under attack~\cite{biggio2014security}. They proposed an empirical framework to evaluate the security of machine learning models. However, their work considers simple binary classification models, such as LR and SVM, and not deep learning models, such as DNN or CNN. 

\BfPara{Categorization} Adversarial machine learning can be categorized based on the attack type and model type. Attacks against machine learning models can be divided into three main categories:  training phase attack~\cite{biggio2011support, Goodfellow2015Explaining}, where the training set is poisoned, test phase attack~\cite{biggio2013evasion}, where AEs are generated to fool the model, and model theft~\cite{papernot2017practical}, where model parameters are extracted. Adversarial machine learning models are also two main categories: white box models~\cite{Jang0J17,PapernotMJFCS16,biggio2013evasion}, where the adversary has a complete access to the machine learning model and associated parameters, and the black box models~\cite{papernot2017practical}, where the adversary has oracle access and no prior information about the model itself.

\BfPara{Gradient-based approach} AEs generation was developed using simple methods based on the back-propagation approach used for network parameter training~\cite{Goodfellow2015Explaining,nguyen2015deep,szegedy2014intriguing}. In this approach, the AEs are generated by determining an optimization problem using the deep learning model's cost function. Specifically, gradients are calculated to update the input, rather than the deep learning network parameters. These inputs are then used to fool the deep learning model. For example, Goodfellow \etal presented a fast gradient sign method to generate AEs~\cite{Goodfellow2015Explaining}. Moosavi-Dezfooli \etal introduced DeepFool to generate universal perturbations while being able to find small directions and being fast~\cite{Moosavi_Dezfooli16}. Similarly, Jang \etal presented a simple gradient-descent based algorithm, called Newtonfool, to find AEs with high performance~\cite{Jang0J17}.

\BfPara{Transferability} Szegedy \etal showed that AEs generated to fool a given neural network model are likely to be misclassified using other  models~\cite{szegedy2014intriguing}. This property, called transferability, enables adversaries to generate AEs and conduct a misclassification attack on a machine learning system even when he has no access to the underlying model~\cite{papernot2017practical}. Furthermore, Yosinki \etal studied the transferability of features among different deep neural networks~\cite{yosinski2014transferable}.

\BfPara{Quality Metrics} In most research works the misclassification rate is considered as the evaluation metric~\cite{Goodfellow2015Explaining} along with human perceptibility~\cite{PapernotMJFCS16}. Jang \etal leveraged a few image quality metrics from the computer-vision community to assess the quality of crafted samples~\cite{Jang0J17}. However, they did not incorporate those quality metrics in the process of generating AEs, thus had no control over the quality of the crafted AEs. This is, the approach does not necessarily guarantee high IQA metrics, and rather assess those metrics once the samples are generated. However, we consider the IQA metrics within along with misclassification rate for optimization.

\section{Conclusion}\label{sec:conclusion}

We proposed \ours, an approach to generate AEs with high misclassification rate while maintaining high-quality in terms of IQA metrics. Those metrics provide significant information, such as brightness properties, about the image. We incorporated edge analysis, Fourier analysis, and feature descriptor into the process of generating AEs to control the quality of the crafted examples within the generation process. In order to obtain high IQA metrics and misclassification rate, we devised a multi-objective particle swarm optimization algorithm.  To evaluate the performance of \ours we have conducted several experiments through different configurations of quality metrics on several benchmarks. A comparison of the results obtained from \ours with several well-known  methods shows that \ours is able to improve the quality of the AEs in terms of multiple IQA metrics. For instance, we are able to achieve misclassification rate of 100\% when distorting only 36.9 pixels, on average, which leads to L2 distance of 2.94, edge difference of 1.37, FFT of 82.12, and HOG feature descriptor metric of 0.63 in the Fashion MNIST benchmark. Our human perception study shows that human subjects were able to easily detect the AEs, with high identification and interpretation rates (as labeled), while transferrability of the AEs was competitive, in comparison with the prior work, and even better than some competing approaches. We found that we could also increase the transferrability rate by increasing the number of perturbed pixels, although at the cost of IQA metrics.

\BfPara{Acknowledgement}
This work is supported in part by National Research Foundation of South Korea under grant NRF-2016K1A1A2912757.

\bibliographystyle{IEEEtran}
\bibliography{ref,conf}

\appendix
\section{Appendix}\label{app:1}


\subsection{Human Perception Study: Demographics}
\begin{figure*}[t]
\centering
        \subfigure[Gender\label{fig:gender}] {\includegraphics[width=0.22\textwidth]{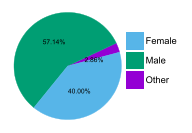}}
		\subfigure[Age Group\label{fig:age}] {\includegraphics[width=0.22\textwidth]{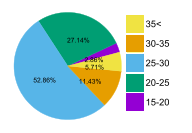}}
		\subfigure[Education Level \label{fig:education}] {\includegraphics[width=0.22\textwidth]{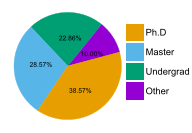}}
		\subfigure[Familiarity with AML \label{fig:familiarity}] {\includegraphics[width=0.22\textwidth]{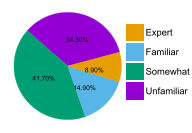}}

\caption{ General demographic information of the participants. In total, we asked 70 individuals, with wide range of background in terms of age, education level, and familiarity with the concept of adversarial machine learning. The majority of the participants were in the age range of 20 to 30 years old~\cref{fig:age}, while with higher level education~\cref{fig:education}, were not familiar or had some familiarity with the concept of AML~\cref{fig:familiarity}.}\label{fig:demographic}
\end{figure*}

Among our participants, 57.14\% and 40\% identified as male and female, respectively, while the rest wished not to identify with either category. The participants' age range was between 15 and 35, with 80\% of them between 20 30 years old. While a majority of the participants had high level of education, only a small number of them were familiar with the concept of adversarial machine learning (23.80\% in total)~\cref{fig:demographic}.

\subsection{Canny Edge Detection}\label{app:CED}
This section describes the five steps that CED takes to detect edges, including noise reduction, computing the intensity gradient, non-maximum suppression, and hysteresis thresholding.

\BfPara{Preprocessing} Images may contain noise that can easily affect the performance of the edge detector, requiring some noise reduction techniques to improve the stability of the edge detector and prevent false detection. Typically, a Gaussian filter is applied on a given image for noise reduction. A Gaussian filter with kernel size of $(2k+1)\times(2k+1)$ is defined as:
\[G(x,y)=\frac{1}{2\pi\sigma^2}\exp^{\left(-{\left(x-\left(k+1\right)\right)^2+\left(y-\left(k+1\right)\right)^2}/{2\sigma^2}\right)},\]  
where $1\leq x,y\leq(2k+1)$ and $\sigma$ represents the standard deviation of the Gaussian distribution. Note that the performance of the CED is dependent on the proper selection of $k$ and $\sigma$. For instance, a larger kernel size reduces the sensitivity of CED into noise, while increasing the localization error of the CED.

\BfPara{Computing the intensity gradient}
Once the given image is smoothed using a Gaussian filter, CED convolves the blurred image with two $3\times3$ Sobel kernels to find the gradient of the image in horizontal $G_{x}$ and vertical $G_{y}$ directions as in~\autoref{eq:X}. Note that $*$ is a convolution operator, which adds each pixel of an image to its neighbor pixels, weighted by the kernel. For instance, for a given image $I$ the convolution of a kernel $f$ with size  $m \times n$ and data value $d$ of a pixel at locations corresponding to the kernel are defined as:
\begin{equation}\label{eq:convolution}
  c =  \left | \left(\sum_{i=1}^{m}\sum_{j=1}^{n} f_{i,j}d_{i,j} \right)/{s}   \right |. 
\end{equation}
Here $c$ represents the output pixel value, while $s$ represents the sum of the coefficients of the kernel.

For a given image $I$ and its gradient images, $G_{x}$ and $G_{y}$, which contain the horizontal and vertical derivative approximations, defined as:
\begin{align}\label{eq:X}
  G_{x} =  \left [\begin{matrix} 1 & 0 & -1 \\ 2 & 0 & -2 \\  1 & 0 & -1 \end{matrix}\right ] * I,\;\;\;  G_{y}  =  \left [\begin{matrix} 1 & 2 & 1 \\ 0 & 0 & 0 \\  -1 & -2 & -1 \end{matrix}\right ] * I, 
\end{align}
 we compute the magnitude $G$ and direction $\theta$ of the gradients of each pixel as: 
\begin{align}
 \label{eq:gtheta}G  =  \sqrt{G^2_{x}+G^2_{y}}, \;\;\; \theta =  \atan\left ( \frac{G_{y}}{G_{x}} \right), 
\end{align}
where $*$ denotes the 2D convolution operation. The found edge directions $\theta$ are rounded to 0\textdegree, 90\textdegree, 45\textdegree, or 135\textdegree, corresponding to horizontal, vertical, and diagonal directions, respectively.

\BfPara{Non-maximum suppression}
It is desired to detect thin and clear edges with no duplicated detection of a given edge. Thus, CED takes an edge thinning step, called the Non-Maximum Suppression (NMS) technique, to provide an accurate response to the edge. NMS sets all the gradient values to zero except the locations with local maxima values. To this end, CED moves a $3\times3$ window over the magnitude and direction of the gradients of the image and sets the magnitude of the central pixel to zero if its gradient value is smaller than the gradient value of two neighbors in the rounded gradient direction $\theta$. In particular, CED selects the central pixel of the window as an edge if its gradient value is as follows.
\begin{enumerate*}
    \item If $\theta =0$\textdegree, the gradient value of the pixels in west and east directions.
    \item If $\theta =$ 45\textdegree, the gradient value of the pixels in south west and north east directions.
    \item If $\theta = $ 90\textdegree, the gradient value of the pixels in south and north directions.
    \item If $\theta = $ 135\textdegree, the gradient value of the pixels in south east and north west directions.
\end{enumerate*}

\BfPara{Hysteresis thresholding}
The output of the NMS step would discard less accurate edges, while preserving edges close to real edges in a given image. However, factors such as color variation result in spurious edge responses, and weak gradient magnitude. These pixels should be filtered out, while pixels with strong gradient magnitude should be preserved. CED performs this task by considering two threshold values, higher threshold $t_{high}$ and lower threshold $t_{low}$. This approach defines strong edges as edges with gradient magnitude larger than $t_{high}$, while edges with magnitude smaller than $t_{high}$ and larger than $t_{low}$ are considered weak edges. Finally, the values smaller than $t_{low}$ are considered non-edge and are set to zero. 

In order to evaluate the validity of weak edges, CED tracks edge connections through hysteresis analysis. A weak edge is considered a valid edge only if it is connected to one strong edge, while invalid weak edges will be suppressed. The intuition behind this technique is that edges produced by noise or color variations are not connected, whereas a weak edge caused by a true edge would be connected into a strong edge. The performance of CED is highly dependent on the value of higher threshold $t_{high}$ and lower threshold $t_{low}$. Moreover, the values of these two thresholds are dependent on the content of the input image, thus their value should be empirically defined.

\begin{figure}[t]
\centering
       {\includegraphics[width=0.45\textwidth]{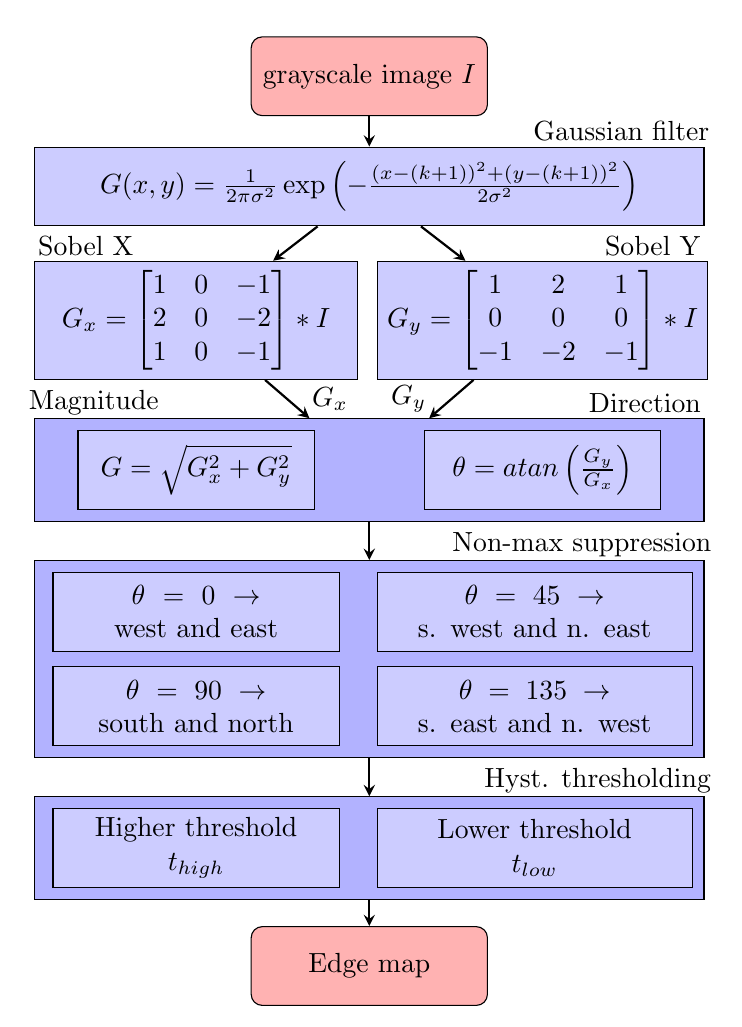}}
  \caption{High-level flowchart of the Canny Edge Detection (CED) algorithm. The performance of the CED depends on the kernel size of the Gaussian filter and the value of upper $t_{high}$ and lower $t_{low}$  thresholds.}
\label{fig:Canny_flowchart}
\end{figure}

\subsection{Fast Fourier Transform}\label{app:FFT}

In this section we briefly describe two dimensional Fourier transform as it is an operation on spatial domain where an image lies. 

Considering a two-dimensional integrable function $f$ of $x$ and $y$ variables, the Fourier transform $F$ on $u$ and $v$ spatial frequencies in $x$ and $y$ directions can be written as:
\begin{equation}\label{eq:FFT_Continuos}
  F(u,v) = \int_{-\infty}^{\infty} \int_{-\infty}^{\infty} f(x,y)\exp^{-2\pi i \left ( ux+vy \right )} dxdy.
\end{equation}

Note that~\autoref{eq:FFT_Continuos} is defined for continuous spatial domain, whereas images are discrete values. Thus, a discrete form of~\autoref{eq:FFT_Continuos} on discrete two dimensional (2D) domain is defined in~\autoref{eq:FFT_discrete}, known as the discrete Fourier transform, and has been used in many applications~\cite{brigham1988fast}.   Namely, the discrete Fourier transform $F$ of a given image $I$ can be achieved by considering $f$ as a function of intensity on $(x,y)$ spatial domains:
\begin{equation}\label{eq:FFT_discrete}
  F(k,l) = \sum_{x=0}^{m-1} \sum_{y=0}^{n-1} f(x,y)\exp^{-2\pi i\left({kx}/{m}+{ly}/{n}\right)}, 
\end{equation}
where $m$ and $n$ represent image dimensions, while $k$ and $l$ are spatial frequencies in $x$ and $y$ directions. Note that the size of both intensity function on spatial domain $f(x,y)$ and frequency domain $F(k,l)$ are the same, matrices of $m \times n$. The discrete Fourier transform can be computed using Fast Fourier Transform (FFT), an efficient algorithms with time complexity of $O(n$\ log $n)$. More details about 2D FFT are in~\cite{brigham1988fast,alleyne1991two}.

\subsection{Histogram of Oriented Gradient}

\BfPara{Computing  the  gradient} The first step in HOG is to compute the magnitude and direction of the gradient for each pixel of a given image $I$. This can be achieved using a one dimensional point derivative mask on the horizontal $G_{x}$ and vertical $G_{y}$ directions as follows:
\begin{align}\label{eq:HOG_Gxy}
  G_{x} =  \left [\begin{matrix} -1 & 0 & 1 \end{matrix}\right ] * I, \;\;\;
G_{y} =  \left [\begin{matrix} -1 \\ 0 \\ 1 \end{matrix}\right ] * I. 
\end{align}
Notice that $*$ represents the convolution operator as defined in~\autoref{eq:convolution}. Then, the magnitude $G$ and direction $\theta$ of the gradients of each pixel can be calculated using~\autoref{eq:gtheta}, used for CED in~\cref{sec:CED}. We note that no preprocessing steps, e.g., image smoothing, are required for HOG descriptor as it performs better without smoothing~\cite{DalalT05}. In addition, the HOG descriptor, unlike CED, does not round the direction $\theta$ to preserve more information for next steps.

\BfPara{Orientation binning} The next step is to divide image into small sub-images, called cells, consisting of pixels, \eg $8 \times 8$. Every pixel inside the cell makes a weighted vote, based on the gradient values computed on previous step, for an orientation-based histogram channel. Once the histogram of each cell is computed, the histograms are assigned to certain histogram channel. Each histogram has multiple channels equally spread over 0\textdegree to 180\textdegree, e.g, nine channels corresponding to 0\textdegree, 20\textdegree,$\dots$, 160\textdegree. Contribution of each pixel would be calculated based on the magnitude of the gradient. 

\BfPara{Descriptor blocks} In order to prevent the effect of local illumination and contrast on the magnitude of the gradients, every histogram needs to be normalized locally. Thus, the HOG descriptor groups multiple cells, e.g., $2 \times 2$, in larger blocks, where the histogram normalization takes place. 

\BfPara{Normalization} As mentioned in the previous section, the HOG descriptor applies normalization to blocks consisting of several histograms to eliminate the impact of local illumination and contrasts on the performance of the descriptor. There are several different techniques for block normalization, including L1 and L2 normalization, among others. However, except the L1 normalization method, the choice of normalization  rarely impacts the overall performance of the HOG descriptor~\cite{DalalT05}. For example, L2 normalization of non-normalized histograms in a block can be defined as:
\[v = \frac{f}{\sqrt{\left|| f \right||_{2}^{2}+e^{2}}},\]
where $f$ represents a non-normalized feature vector of all histograms in a block, $\left|| f \right||_{2}^{2}$ shows L2-norm, $e$ is a small constant, and $v$ is the normalized feature vector.

\begin{figure}[t]
\centering
       {\includegraphics[width=0.45\textwidth]{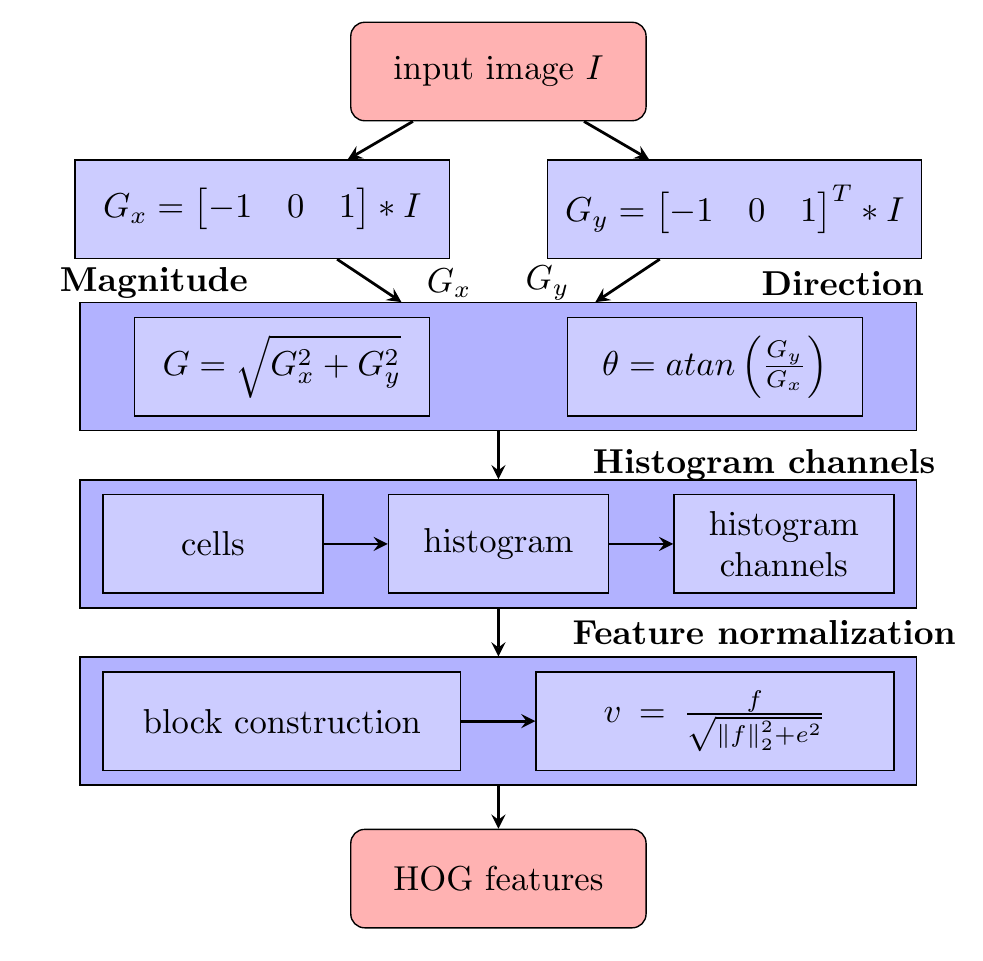}}
\caption{High-level flowchart of the Histogram of Oriented Gradient (HOG) feature descriptor algorithm.}
\label{fig:HOG_flowchart}
\end{figure}

\subsection{Adversarial Attacks on Deep Learning}

A brief description of these adversarial attack categories assumed in our threat model is provided below. 

\BfPara{Targeted attacks} 
The focus of this attack is to generate AE $x'$ that forces the classifier $f$ to misclassify into a specific target class $t$. For instance, the adversary generates a set of malicious IoT software samples, which are classified as benign. That is:
   $ x': [f\left ( x' \right) = t] \wedge [ \Delta \left ( x , x' \right) \leq \epsilon ]$,
where $f(.)$  represents the classifier’s output, $\Delta \left ( x , x' \right)$ denotes the difference between $x$ and the crafted AE $x'$, whereas $\epsilon$ is a distortion threshold.

\BfPara{Untargeted attacks} 
The focus of untargeted attack is to generate an AE that forces the classifier $f$ to misclassify to any class other than the original class $f(x)$, where $x$ is the original input. That is:
 $x': [f\left ( x' \right) \neq f\left ( x \right)] \wedge [\Delta \left ( x , x' \right) \leq \epsilon]$, 
where $f(.)$ shows the classifier's output, $\Delta \left ( x , x' \right)$ represents the difference between $x$ and $x'$, and  $\epsilon$ is the distortion threshold.

\BfPara{White-box attacks} 
In this attacks, the adversary has full access to the model and complete knowledge of its network topology and all links weights. Accordingly, the adversary can carry out an unlimited number of queries to the model until a successful adversarial example is ultimately generated~\cite{Carlini017, PapernotMJFCS16, sharifBBR16,Moosavi-Dezfooli16}. With the prior knowledge of the model internals, the adversary has a high chance in generating AE with small perturbation, which increases both the success rate and the effort required to detect the crafted AEs~\cite{wangYVZZ18}.

\BfPara{Black-box attacks} 
In this attack, the adversary has no prior knowledge of the model internals. Here, the adversary has oracle access to the model, enabling him to conduct queries to the model~\cite{sharifBBR16,PapernotMGJCS17}. Two main approaches are followed in the black-box attacks: reverse-engineering the decision boundaries of the network~\cite{PapernotMGJCS17} or querying intermediate AEs and improving the generated AE iteratively~\cite{sharifBBR16}. Therefore, this type of attack requires enormous number of quires to the model and achieves a relatively lower misclassification rate~\cite{sharifBBR16}.

\subsection{Adversarial Attack Methods}

In order to demonstrate the performance of the proposed method, the obtained results are compared with those of popular existing adversarial attack methods. In the following a brief description of these attacks are provided.

\subsubsection{Fast Gradient Sign Method} \label{sec:attacks_FGSM}

FGSM is designed to generate AEs in a fast matter, while not guaranteeing similarity of the generated examples compared to the original samples\cite{GoodfellowSS14}. It performs based on one-step gradient update, which can be expressed using~\autoref{eq:FGSM_Main} as following:
\begin{equation}\label{eq:FGSM_Main}
    \delta= \epsilon \cdot sign \left(\nabla_{x} J_{\theta} \left( x, l \right) \right)
\end{equation}
where, $\delta$ is the perturbation parameter, $\epsilon$ is a small scalar value that limits the distortion of the image and sets the magnitude of the perturbation. In addition, $sign(.)$ denotes the sign function, $J(.)$ is the cost function, where $x$ is the input image and $l$ is the label associated with it, and $\nabla$ computes the gradient of the cost function $J$ around current value of $x$. Finally, the output image is can be represented using~\autoref{eq:FGSM_OutImage}.

\begin{equation}\label{eq:FGSM_OutImage}
    \ x' =x-\delta
\end{equation}
Here, $x'$ is the adversarial example corresponding to image $x$. Controlling the distortion parameter $\epsilon$ will affect the $L_2$ norm distance between the original and the AEs. The higher $\epsilon$ value is, the larger $L_2$ distance. For visualization, the $x'$ values is clapped to suits 0-255 one channel scale~\cite{SuVPSFS18}.

\subsubsection{Carlini \& Wagner Method} \label{sec:attacks_CW}

Carlini and Wagner \cite{Carlini017} introduced three gradient based attacks by optimizing the penalty and distance metrics on  $L_\infty$, $L_2$, and $L_0$ norms as shown in the equation~\autoref{eq:CW_Main} below:
\begin{equation}\label{eq:CW_Main}
    min \, ||\delta||^2_p \quad s.t. \  g\left( x+\delta \right) = y^* \ \& \  x+\delta \in X 
\end{equation}
where, $\delta$ is the perturbation parameter, $g(.)$ is the objective function based on hinge loss, $y^*$ is the targeted class and $x$ is the input image. This equation ensures that added perturbation will be small and only as required to misclassify to class $y^*$.\\
A small modification can be made to launch non-targeted attacks as shown in equation~\autoref{eq:CW_NonTargetted} below:
\begin{equation}\label{eq:CW_NonTargetted}
    min \, ||\delta||^2_p \quad s.t. \  g\left( x+\delta \right) \neq y \ \& \  x+\delta \in X 
\end{equation}
In this paper, we will focus on $L_2$ based C\&W attack to generate AEs. $L_2$ distance represents the required amount of changes into the application binaries to generate the adversary example, lower $L_2$ distance indicates lower altering in application binaries. The perturbation $\delta$ is defined as the following in equation~\autoref{eq:CW_Perturbation}.
\begin{equation}\label{eq:CW_Perturbation}
    \delta = \frac{1}{2}\left( tanh\left(w\right)+1\right)-x.
\end{equation}
where $tanh(.)$ is hyperbolic tangent function, $w$ is an auxiliary variable optimized in equation~\autoref{eq:CW_Wopt}.
\begin{equation}\label{eq:CW_Wopt}
    \min\limits_{w} || \frac{1}{2}\left( tanh\left(w\right)+1\right)||_2 + c\cdot g\left(\frac{1}{2}\left(tanh\left(w\right)+1\right)\right)
\end{equation}
where $c$ is a constant. C\&W method minimizes the $L_p$ norm distance between the generated adversary example and original image to increase the similarity and harden the detection process.

\subsubsection{Momentum Iterative Method} \label{sec:attacks_MIM}
Momentum iterative method (MIM) is based on applying momentum gradient over basic FGSM to generate AEs \cite{dong2018boosting}, the main goal is to maintain efficiency against black box models. The main object for this method is to find $x^*$ that misclassify the model that satisfy the equation~\autoref{eq:MIM_Object} where $J$ is the loss function, and $\epsilon$ is small scalar that control the maximum distortion allowed.
\begin{equation}\label{eq:MIM_Object}
    arg \max\limits_{x^*} \  J\left(x^*,y\right), \quad s.t. \  ||x^*-x||_\infty \leq \epsilon
\end{equation}
The momentum gradient then calculated using~\autoref{eq:MIM_Momentum}.
\begin{equation}\label{eq:MIM_Momentum}
    g_{t+1} = \mu g_t + \frac{\nabla_x J_{\theta}\left(x'_t , l\right)}{||\nabla_x J_{\theta}\left(x'_t , l\right)||}
\end{equation}
where $\mu$ is the decay factor, $x'_0$ initially is the original input, and $g_0$ initially 0. Each iteration, $x'$ will be updated according to this equation: 
\begin{equation}\label{eq:MIM_Adversary}
    x^*_{t+1} = x^*_t + \epsilon \cdot sign\left(g_t+1\right)
\end{equation}
after $n$ number of iterations, the $x'_{t+1}$ will be returned as the adversarial example for input $x$.

\subsubsection{PGD Method} \label{sec:attacks_Madry}
Madry \etal introduced iterative projected gradient descent (PGD) attack \cite{madry2018towards}. The aim is to generate adversarial sample under a minimized empirical risk with a trade of high performance cost. The original empirical risk minimization ($ERM$) for the model is referred in as the following notation: 
\begin{equation}\label{eq:Madry_RiskNotation}
    \mathbb{E}_{(x,y)\sim D}[L\left(x,y,\theta\right)]
\end{equation}
where $L$ is the loss function. By modifying the definition of $ERM$ and allowing the adversary to perturb the input, s described in equation~\autoref{eq:Madry_ERM}, our goal is to minimize the risk while adding perturbation.
\begin{equation}\label{eq:Madry_ERM}
    \min\limits_\theta \rho\left(\theta\right), \quad where \ \  \rho\left(\theta\right) = \mathbb{E}_{(x,y)\sim D}[\max\limits_{\delta\in S}  L\left(x+\delta,y,\theta\right)]
\end{equation}
 
where $\delta$ is the perturbation, and $\rho\left(\theta\right)$ is the objective function to minimize. As an iterative method, $x'_{t+1}$ is updated each iteration according to the previous $x'_{t}$ and the generated perturbation, after $n$ iterations, $x'_{t+1}$ will be returned as the adversarial sample of $x$.

\end{document}